\pdfoutput=1

\documentclass[11pt]{article}

\usepackage{amsmath,amsfonts,bm}

\def\eqref#1{equation~\ref{#1}}

\def\1{\bm{1}}

\def\vc{{\bm{c}}}

\def\vr{{\bm{r}}}
\def\vs{{\bm{s}}}

\def\mL{{\bm{L}}}

\def\mS{{\bm{S}}}

\def\mU{{\bm{U}}}

\def\mW{{\bm{W}}}
\def\mX{{\bm{X}}}

\DeclareMathAlphabet{\mathsfit}{\encodingdefault}{\sfdefault}{m}{sl}
\SetMathAlphabet{\mathsfit}{bold}{\encodingdefault}{\sfdefault}{bx}{n}

\DeclareMathOperator*{\argmin}{arg\,min}

\usepackage[final]{acl}

\usepackage{times}
\usepackage{latexsym}

\usepackage[T1]{fontenc}

\usepackage[utf8]{inputenc}

\usepackage{microtype}

\usepackage{inconsolata}

\usepackage{graphicx}

\usepackage{amsmath}
\usepackage{amssymb}
\usepackage{mathtools}
\usepackage{amsthm}

\usepackage{subfigure}
\usepackage{booktabs}
\usepackage{makecell}
\usepackage{arydshln}
\usepackage{wrapfig}

\title{LRQ: Optimizing Post-Training Quantization for Large Language Models by Learning Low-Rank Weight-Scaling Matrices}

\author{Jung Hyun Lee\thanks{Equal contribution} \\
    NAVER Cloud \\
    \texttt{onliwad101@gmail.com} \\\And
    Jeonghoon Kim\footnotemark[1] \\
    NAVER Cloud \\
    \texttt{jeonghoon.samuel@gmail.com} \\\And
    June Yong Yang \\
    KAIST AI \\
    \texttt{laoconeth@kaist.ac.kr} \\\And
    Se Jung Kwon \\
    NAVER Cloud \\
    \texttt{sejung.kwon@navercorp.com} \\\AND
    Eunho Yang \\
    KAIST AI, AITRICS \\
    \texttt{eunhoy@kaist.ac.kr} \\\And
    Kang Min Yoo \\
    NAVER Cloud, SNU AI Center \\
    \texttt{kangmin.yoo@gmail.com} \\\And
    Dongsoo Lee \\
    NAVER Cloud \\
    \texttt{dongsoo.lee@navercorp.com} \\
  }

\author{
 \textbf{Jung Hyun Lee\thanks{Equal contribution}\textsuperscript{1$\dagger$}},
 \textbf{Jeonghoon Kim\footnotemark[1]\textsuperscript{1$\dagger$}},
 \textbf{June Yong Yang\textsuperscript{2}},
 \textbf{Se Jung Kwon\textsuperscript{1}},
\\
 \textbf{Eunho Yang\textsuperscript{2,3}},
 \textbf{Kang Min Yoo\textsuperscript{1,4}},
 \textbf{Dongsoo Lee\textsuperscript{1}}
\\
\\
 \textsuperscript{1}NAVER Cloud,
 \textsuperscript{2}KAIST AI,
 \textsuperscript{3}AITRICS,
 \textsuperscript{4}SNU AI Center
\\
 \small{
   \textbf{Correspondence\textsuperscript{$\dagger$}:} \href{mailto:onliwad101@gmail.com}{onliwad101@gmail.com}, \href{mailto:jeonghoon.samuel@gmail.com}{jeonghoon.samuel@gmail.com}
 }
}

\begin{document}
\maketitle
\begin{abstract}
With the commercialization of large language models (LLMs), weight-activation quantization has emerged to compress and accelerate LLMs, achieving high throughput while reducing inference costs. However, existing post-training quantization (PTQ) techniques for quantizing weights and activations of LLMs still suffer from non-negligible accuracy drops, especially on massive multitask language understanding. To address this issue, we propose Low-Rank Quantization (LRQ) $-$ a simple yet effective post-training weight quantization method for LLMs that reconstructs the outputs of an intermediate Transformer block by leveraging low-rank weight-scaling matrices, replacing the conventional full weight-scaling matrices that entail as many learnable scales as their associated weights. Thanks to parameter sharing via low-rank structure, LRQ only needs to learn significantly fewer parameters while enabling the individual scaling of weights, thus boosting the generalization capability of quantized LLMs. We show the superiority of LRQ over prior LLM PTQ works under (i) $8$-bit weight and per-tensor activation quantization, (ii) $4$-bit weight and $8$-bit per-token activation quantization, and (iii) low-bit weight-only quantization schemes. Our code is available at Software.
\end{abstract}


\section{Introduction}\label{sec:intro}

As ChatGPT and GPT-4 \citep{openai2023gpt4} have showcased unprecedented capabilities across various domains such as common sense reasoning, mathematical problem-solving, and coding proficiency, there has been an exponential surge in interest surrounding the development of Large Language Models (LLMs). This surge in interest has culminated in the recent release of cutting-edge LLMs like Llama \citep{touvron2023llama}, PaLM $2$ \citep{google2023palm}, and Llama $2$ \citep{touvron2023llama2}. Accordingly, serving LLMs has rapidly emerged as a significant concern in both academia and industry. This stems from the substantial memory footprint and considerable computational cost incurred when operating these language models with tens or hundreds of millions of parameters in FP16 format. Therefore, extensive efforts \citep{frantar2023optq, liu2023deja} such as quantization or pruning are underway to compress LLMs and provide efficient deployment. In particular, quantization has garnered considerable interest among LLM engineers and researchers because quantization aids in not just model compression but also inference acceleration.

LLM quantization techniques fall into two primary categories: weight-only quantization and weight-activation quantization. Weight-only quantization concentrates on enhancing memory-bound operations like matrix-vector multiplication by quantizing weights of LLMs into low-bit integers (e.g., $2$-$4$ bits). With activations kept in FP16, weight-only quantization exhibits marginal accuracy degradation but is only effective in accelerating text generation inference for small batch sizes (e.g., a single batch). In contrast, weight-activation quantization aims to expedite computationally intensive operations, such as matrix-matrix multiplication, typically by quantizing both weights and activations of LLMs into 8-bit integers and employing INT8 GEMM kernels. This comprehensive quantization approach enables LLM serving for large batch sizes, thus enhancing LLM throughput and expediting LLM inference through integer matrix multiplication. Yet, it comes with the trade-off of potential non-negligible accuracy drop. While each approach boasts its own set of strengths and weaknesses, we first focus on weight-activation quantization on the grounds that achieving high-throughput LLM inference is important to handle a substantial volume of user requests in real time.

\begin{figure*}
    \vskip -0.39in
    \centering
    \subfigure[Common Sense Reasoning tasks]
    {
    \includegraphics[width=0.463\linewidth]{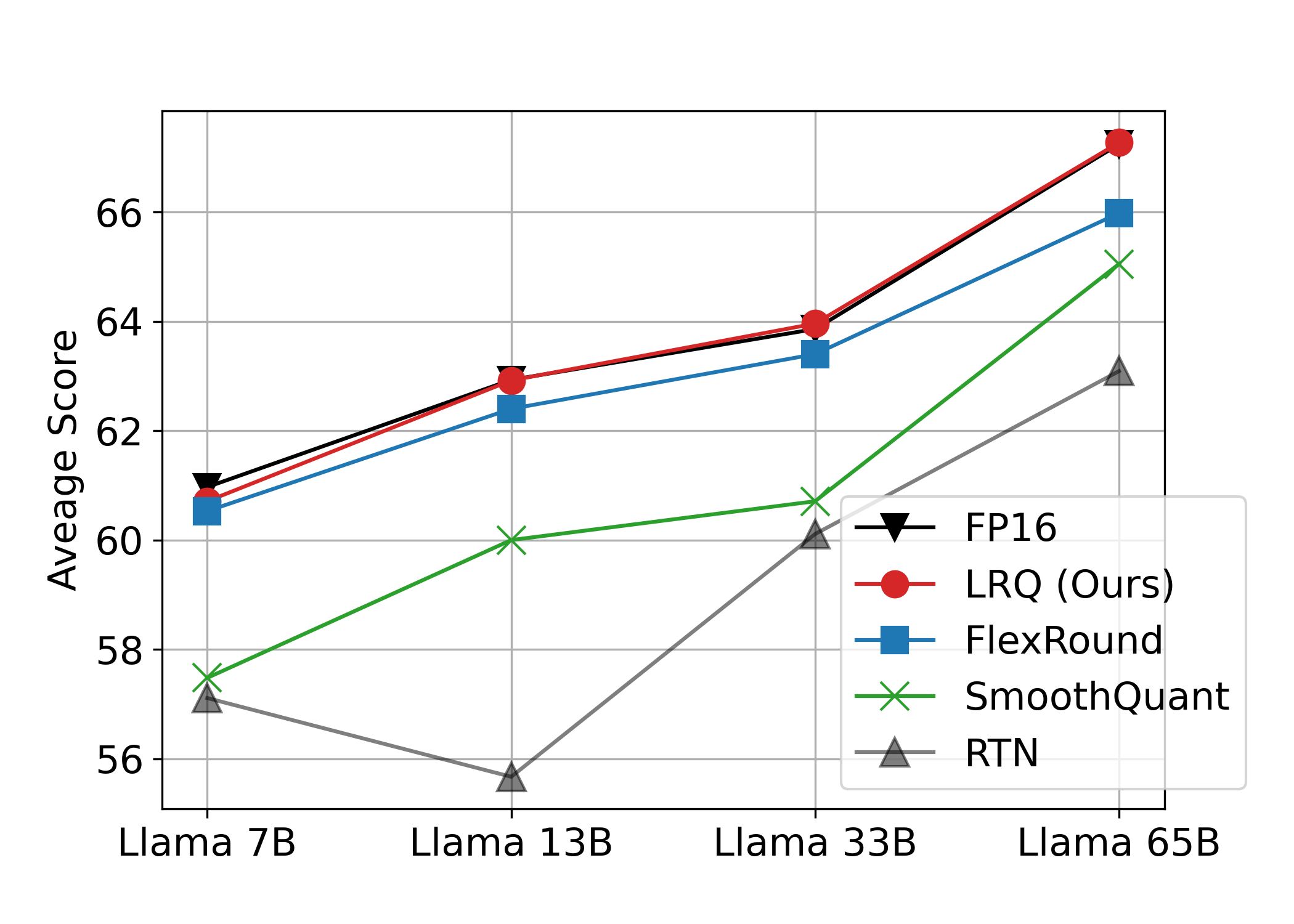}
    \label{fig1:a}
    }
    \subfigure[Massive Multitask Language Understanding] 
    {
    \includegraphics[width=0.463\linewidth]{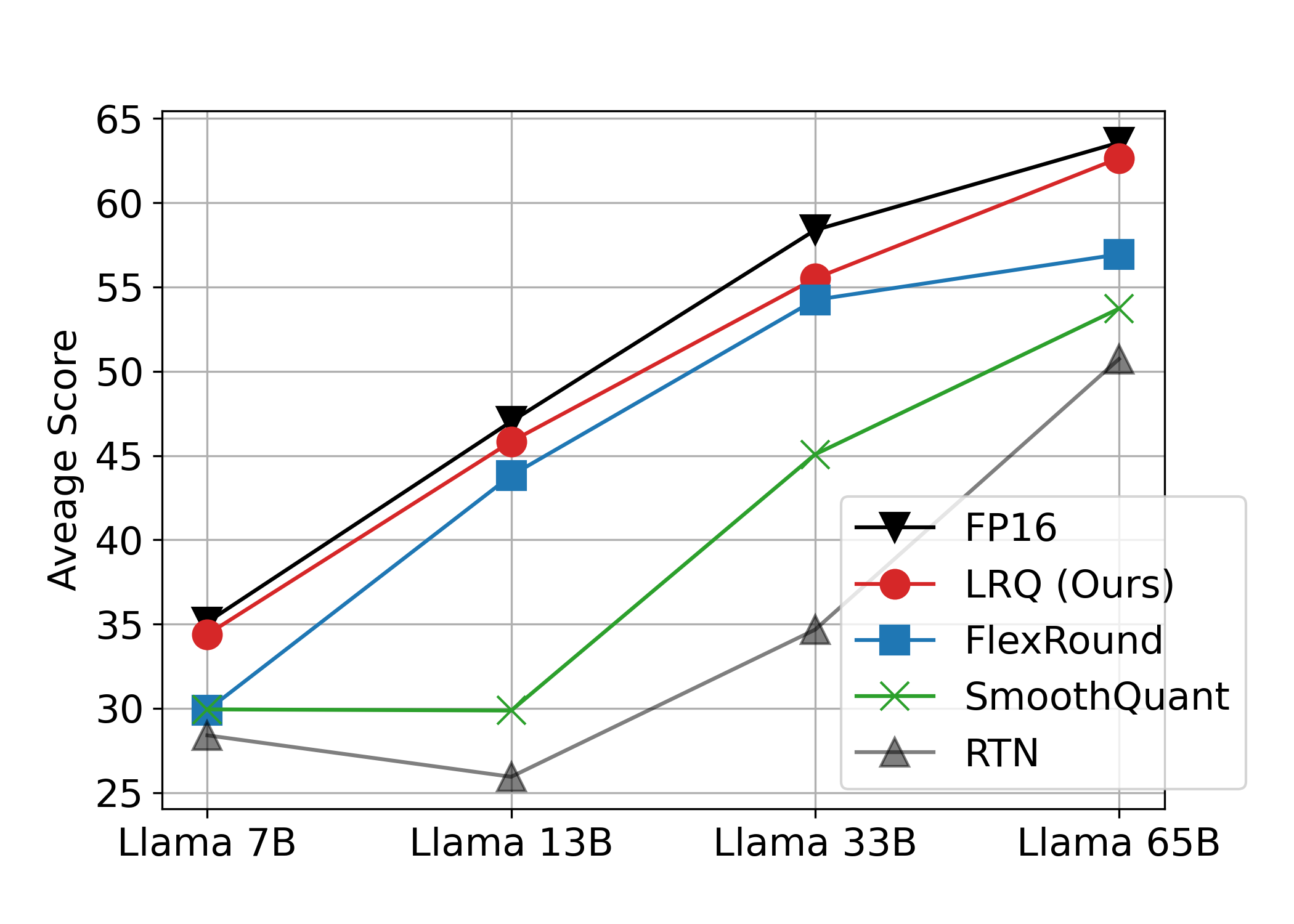}
        \label{fig1:b}
    }
    \vskip -0.15in
    \caption{(a) Zero-shot performance and (b) five-shot accuracy of Llama with $8$-bit per-channel asymmetric weight quantization and $8$-bit per-tensor asymmetric static activation quantization, while keeping the KV cache in FP16.} 
    \label{fig:smoothquant_sniper}
    \vskip -0.2in
\end{figure*}

Recent studies \citep{dettmers2022llm, yao2022zeroquant, xiao2022smoothquant, lee2023flexround, liu2023llmqat} have attempted to quantize both weights and activations of LLMs. Among these works, only SmoothQuant \citep{xiao2022smoothquant} and FlexRound \citep{lee2023flexround} demonstrated the potential for a hardware-efficient per-tensor static activation quantization scheme that can reduce the inference latency and memory usage by up to two-thirds and half respectively compared to FP16 baselines as thoroughly elucidated in \citet{xiao2022smoothquant}. Given the compelling advantages of this scheme, we also stick mainly to per-tensor static activation quantization, with a primary focus on preventing non-negligible performance degradation, one of its key drawbacks, from occurring.

Despite promising results that SmoothQuant and FlexRound yielded, they still possess inherent limitations on enhancing model accuracy when using per-tensor static activation quantization.
Although SmoothQuant is a potent technique for alleviating the difficulty of quantizing activation outliers, it uniformly divides activations in each channel and multiplies the weights in the corresponding input channel by some constant. Since such an uniform per-channel smoothing transformation can only scale the weights collectively per channel, not individually,
SmoothQuant may lead to non-negligible accuracy loss after quantization for certain models as in Figure \ref{fig:smoothquant_sniper}. On the other hand, as FlexRound learns a separate scale for each weight and thus enables flexible weight quantization based on individual characteristics of each weight,
FlexRound can show marginal zero-shot accuracy drop on common sense reasoning tasks in Figure \ref{fig1:a}. However, as depicted in Figure \ref{fig1:b}, FlexRound falls short in performing well on massive multitask language understanding (MMLU), which necessitates problem-solving skills, specialized knowledge, as well as basic knowledge across diverse subjects. We empirically confirm that this phenomenon is because  FlexRound has to learn too many scales relative to limited calibration samples due to the assignment of an independent scale to every weight.

To improve generalization performance on such a challenging benchmark, we propose a new post-training weight quantization approach, “Low-Rank Quantization (LRQ)”, as a middle ground between SmoothQuant and FlexRound. LRQ is designed to minimize the mean squared error between the outputs of an intermediate FP16 Transformer block and those of its quantized counterpart with respect to low-rank weight-scaling matrices instead of full weight-scaling matrices that involve as many scales as their associated weights.
Using such low-rank matrices, we can reduce the number of learnable parameters effectively while maintaining the concept of scaling weights individually by sharing learnable parameters via low-rank structure. As a result, LRQ can attain comparable accuracy to FP16 baselines on both common sense reasoning tasks and MMLU for every Llama model as seen in Figure \ref{fig:smoothquant_sniper}.

Our main contribution is threefold:
\begin{itemize}
    \item We propose a new post-training weight quantization method coined LRQ that leverages low-rank weight-scaling matrices 
    for intermediate Transformer block output reconstruction, which improves the generalization performance of quantized LLMs as in Figure \ref{fig:smoothquant_sniper}.
    \item We provide empirical insights into the significance of reducing the number of learnable parameters 
    and how the utilization of low-rank matrices to effectively decrease learnable parameters impacts the generalization ability of quantized LLMs.
    \item We validate the effectiveness of LRQ 
    under a wide variety of quantization schemes ($8$-bit weight and per-tensor activation quantization, $4$-bit weight and $8$-bit per-token activation quantization, and low-bit weight-only quantization) with marginal accuracy loss. 
\end{itemize}

\section{Method}\label{sec:method}

In this section, we outline the post-training quantization (PTQ) background that our method, LRQ is based on, figure out the problem arising when quantizing LLMs, and formulate LRQ. Finally, we deepen an empirical understanding of how LRQ can improve generalization in quantized LLMs.

\subsection{Background}\label{subsec:background}

\paragraph{Block-wise Reconstruction} First of all, our method is based on block-wise reconstruction, which originates from BRECQ \citep{li2021brecq} for the purpose of taking into account the intra-block dependency and has been widely used in QDrop \citep{wei2022qdrop}, FlexRound \citep{lee2023flexround}, and AQuant \citep{li2023efficient} due to its efficacy to yield less generalization error than layer-wise reconstruction. As we concentrate on weight-activation quantization of LLMs that are generally Transformer-based models, the block-wise reconstruction process is applied to every Transformer block in the order of arrangement. To be more concrete, with a small set of calibration data, the objective of block-wise reconstruction is to find quantized weights $\widehat{\mW}$ by minimizing the block reconstruction error $\|\mW\mX - \widehat{\mW}\widetilde{\mX}\|^2_2$ where $\mW$ and  $\mX$ are the weights and inputs of a FP16 Transformer block while $\widetilde{\mX}$ is the inputs of its quantized counterpart (i.e., the outputs of its immediately preceding Transformer block with all its previous Transformer blocks quantized).

\paragraph{FlexRound} Among PTQ studies that take advantage of block-wise reconstruction, FlexRound shows the state-of-the-art performance for a wide variety of models ranging from computer vision models to large language models including Llama. In FlexRound, the formulation of $\widehat{\mW}$ is written as
\begin{equation}
    \widehat{\mW} = \vs_1 \Big\lfloor \frac{\mW}{{\vs_1 \odot \text{exp}(\mS_2)}} \Big\rceil\label{eq:flexround},
\end{equation}
where $\vs_1$ is a quantization step size, $\mS_2$ is a weight-scaling matrix whose shape is exactly the same as that of $\mW$, $\lfloor \cdot \rceil$ and $\text{exp}(\cdot)$ indicate the rounding and exponential function, and $\odot$ and $\mathbin{/}$ represent element-wise multiplication and division. Depending on the type of $\mW$, some supplementary vectors are added to $\mS_2$, but we exclude these additional vectors to keep the expression uncluttered. At the beginning of learning, $\mS_2$ is set to a zero matrix and $\vs_1$ is initialized to $\argmin_{\vs_1} \|\mW - \widehat{\mW}\|^2_2$ to start learning from rounding-to-nearest (RTN). Then, both $\vs_1$ and $\mS_2$ are learned to minimize the block reconstruction error $\|\mW\mX - \widehat{\mW}\widetilde{\mX}\|^2_2$ with a small amount of calibration data as explained above. As FlexRound learns a separate scale for each weight, FlexRound can quantize each weight to one of not just the two nearest quantization grids but also more distant ones, based on individual characteristics of each weight. Nonetheless, as shown in Figure \ref{fig:smoothquant_sniper}, quantized LLMs via FlexRound might exhibit reduced scores on challenging tasks like MMLU.  

\subsection{Motivation}\label{subsec:motivation}

We hypothesize that the failure to generalize well on challenging benchmarks like MMLU arises from the necessity of learning an individual scale for every weight with limited calibration samples. Now that $\mS_2$ has as many learnable parameters as the size of $\mW$ in Eq. \ref{eq:flexround}, FlexRound's objective to achieve flexible weight quantization through the assignment of an independent scale to each weight may be deemed excessive when applied to LLM. 
\begin{figure}
    \vskip -0.05in
    \subfigure{\includegraphics[scale=0.51]{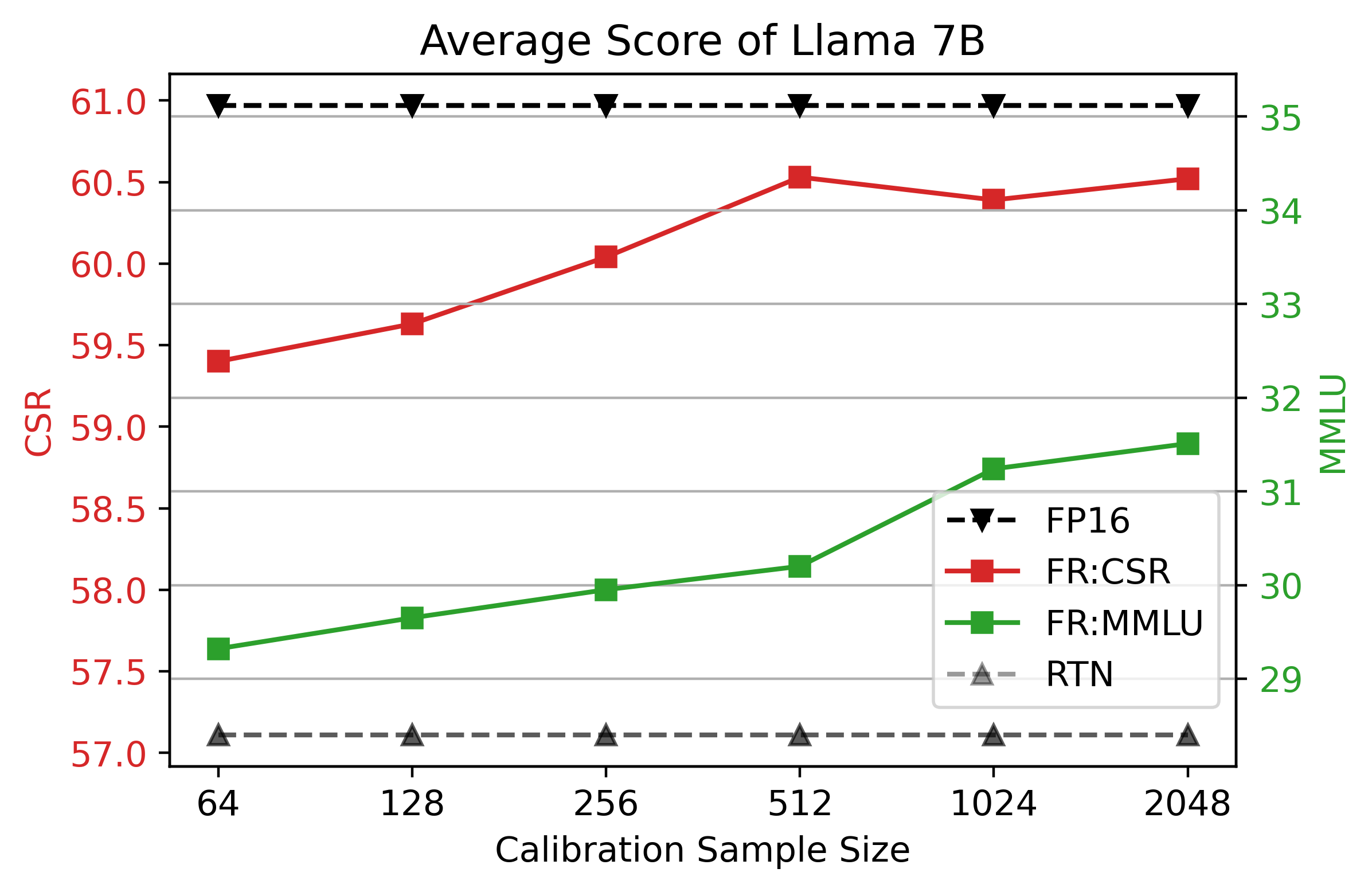}}
    \vskip -0.1in
    \caption{Zero-shot performance and five-shot accuracy of Llama $7$B for FlexRound (FR) on common sense reasoning (CSR) tasks and MMLU according to the calibration sample size, with $8$-bit per-channel asymmetric weight and $8$-bit per-tensor asymmetric static activation quantization, while keeping the KV cache in FP16.}
    \vskip -0.2in
    \label{fig:flexround_sample_size}
\end{figure}

\begin{figure*}
    \vskip -0.3in
    \centering
    \subfigure[Calibration sample] 
    {
    \includegraphics[width=0.325\linewidth]{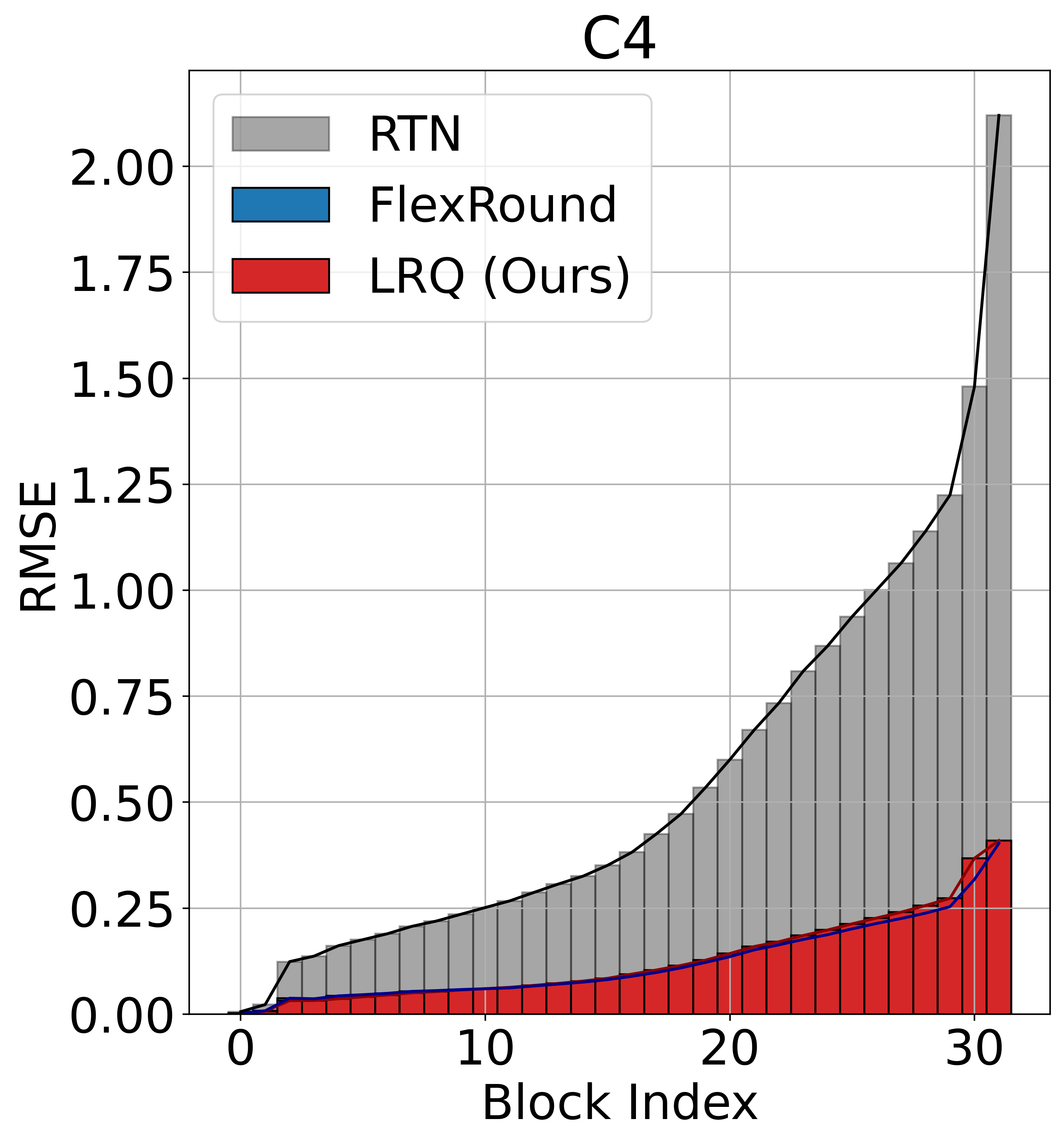}
        \label{fig2:a}
    }
    \subfigure[Unseen sample]
    {
    \includegraphics[width=0.615\linewidth]{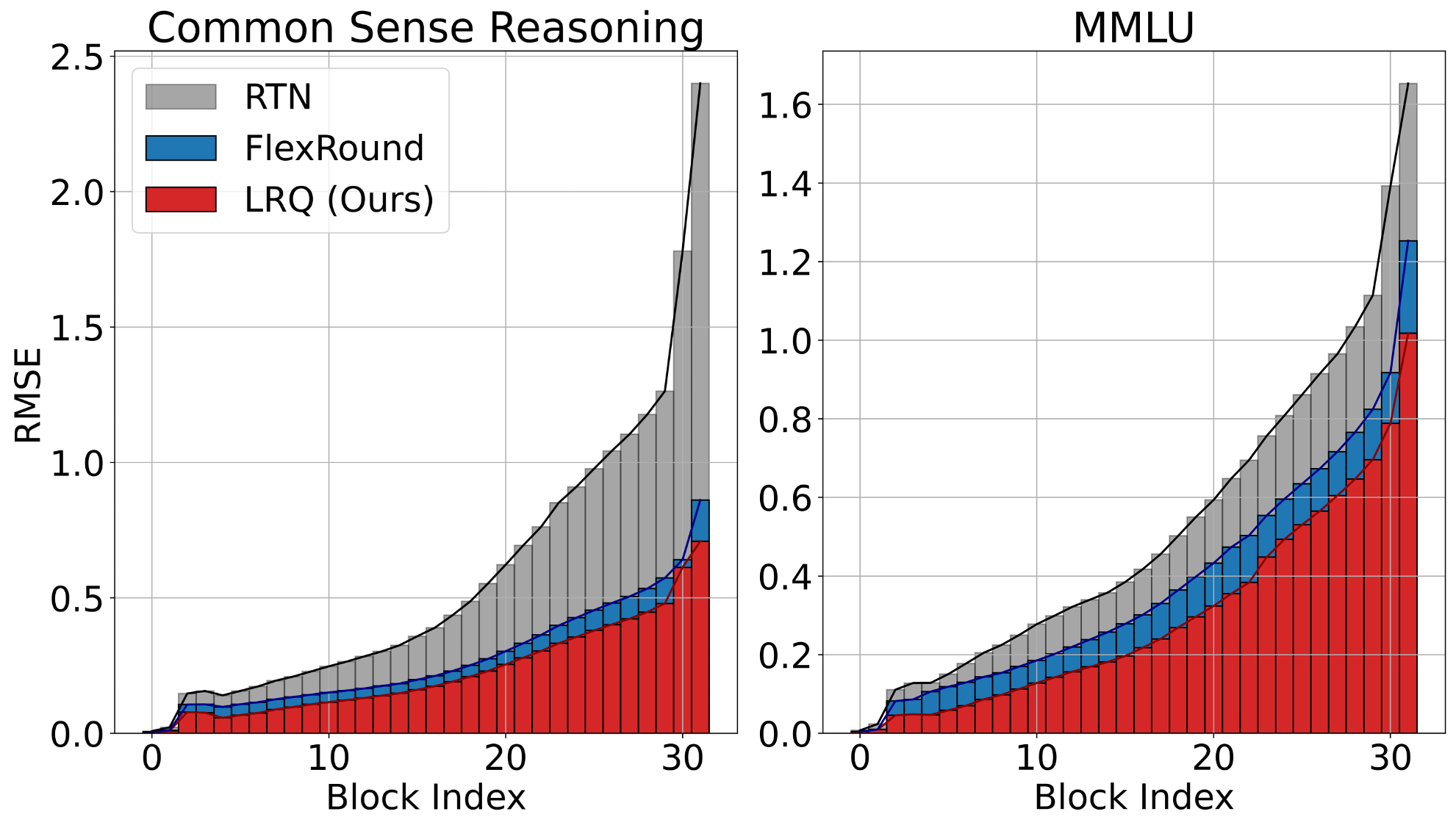}
    \label{fig2:b}
    }
    \vskip -0.15in
    \caption{Accumulated root mean square error (RMSE) between $\mW\mX$ and $\widehat{\mW}\widetilde{\mX}$ for RTN, FlexRound, and LRQ on (a) a calibration sample from the C4 dataset and (b) an unseen sample from common sense reasoning and MMLU benchmarks, ranging from the first Transformer block to the last Transformer block of Llama $7$B. Here, weights and activations are quantized to $8$-bit with per-channel asymmetric quantization and per-tensor asymmetric static quantization, while the KV cache remains in FP16. Note that RMSE tends to rise in line with the block index due to the presence of $\widetilde{\mX}$ that accumulates quantization error resulting from previous quantized Transformer blocks.}
    \label{fig:recon}
    \vskip -0.2in
\end{figure*}

For instance, for Llama $7$B, the smallest model in Llama, FlexRound has to learn more than $200$ million scales with only just a few hundred or thousand calibration samples. FlexRound may be therefore prone to overfitting when quantizing LLMs. To resolve this issue, there might be two solutions: (i) increasing calibration samples, and (ii) decreasing learnable parameters. In the former case, as shown in Figure \ref{fig:flexround_sample_size}, the accuracy of FlexRound on MMLU increases as the calibration sample size grows larger. Yet, FlexRound still falls behind the FP16 baseline on MMLU by more than $3.5$ percent, even when utilizing $2048$ calibration samples, the maximum number we can use on a single NVIDIA A100-80GB GPU during the block-wise reconstruction process. Thus, we turn our focus toward reducing the number of learnable parameters.

\subsection{Low-Rank Quantization}\label{subsec:LRQ}
To reduce the number of learnable parameters, we decompose a weight-scaling matrix, $\mS_2$, into a low-rank matrix before performing the reconstruction process. To be more specific, for $\mW \in \mathbb{R}^{C_{out} \times C_{in}}$, $\mS_2 \in \mathbb{R}^{C_{out} \times C_{in}}$ is factorized into $\mL_2 \mU_2$ where $\mL_2 \in \mathbb{R}^{C_{out} \times r}$ and $\mU_2 \in \mathbb{R}^{r \times C_{in}}$ for $r < \min(C_{out}, C_{in})$. Additionally, we supplement $\mL_2 \mU_2$ with a row vector, $\vr_2 \in \mathbb{R}^{C_{out} \times 1}$ and a column vector, $\vc_2 \in \mathbb{R}^{1 \times C_{in}}$, which is inspired by the addition of a row or column vector (or both) to a low-rank matrix in recommendation systems, one of the most popular applications of low-rank structure, for better prediction of ratings by considering a bias for each user or each item \citep{jahrer12bcollaborative, goodfellow2016deep, koren2021advances}.
As a result, we formulate $\widehat{\mW}$ as
\begin{equation}
    \widehat{\mW} = 
    \vs_1 \Big\lfloor \frac{\mW}{{\vs_1 \odot \text{exp}(\mL_2 \mU_2 + \vr_2 + \vc_2)}} \Big\rceil\label{eq:LRQ},
\end{equation}
which we refer to as `Low-Rank Quantization (LRQ)'. At first, $\mL_2$ and $\mU_2$ are initialized to zeros and random values from a normal distribution respectively, and $\vr_2$ and $\vc_2$ are set to zero vectors so that $\mL_2 \mU_2 + \vr_2 + \vc_2$ starts from a zero matrix like $\mS_2$ in Eq. \ref{eq:flexround}. Then, $\vs_1$, $\mL_2$, $\mU_2$, $\vr_2$, and $\vc_2$ are learned to minimize $\|\mW\mX - \widehat{\mW}\widetilde{\mX}\|^2_2$ in a block-by-block manner. 
The ablation study on the effect of $\vr_2$ and $\vc_2$ in LRQ is presented in Appendix \ref{appendix:r2_and_c2}.

\begin{table*}[t]
\vskip -0.2in
\caption{Zero-shot performance of Llama on common sense reasoning tasks (BoolQ, PIQA, HellaSwag, WinoGrande, ARC easy and challenge, and OpenBookQA) with per-channel asymmetric weight quantization, per-tensor asymmetric static activation quantization, and per-token asymmetric KV cache quantization. The accuracy ($\%$) is reported for all tasks. The number of bits used for weights, activations, and KV cache is $8$-bit.} 
\vskip -0.15in
\label{tab:llama_csr_per_tensor}
\begin{center}
\small
\resizebox{\linewidth}{!}{
\begin{tabular}{lccccccccc}
\toprule
Method & \makecell{\# Bits (W/A/KV)} & BoolQ & PIQA & HellaSwag &  WinoGrande & ARC-e & ARC-c & OBQA & Average\\
\midrule
Llama $7$B & $16/16/16$ & $73.15$ & $77.31$ & $72.96$ & $67.09$ & $52.48$ & $41.38$ & $42.40$ & $60.97$ \\
\midrule
SmoothQuant & $8/8/8$ & $69.42$ & $72.63$ & $69.07$ & $64.72$ & $48.61$ & $37.12$ & $39.20$ & $57.25$ \\
FlexRound & $8/8/8$ & $72.54$ & $76.50$ & $71.88$ & $66.77$ & $53.03$ & $39.76$ & $42.00$ & $60.35$ \\
LRQ (Ours) & $8/8/8$ & $72.84$ & $77.37$ & $72.04$ & $67.01$ & $53.03$ & $40.53$ & $41.60$ & $\mathbf{60.63}$ \\
\midrule
Llama $13$B & $16/16/16$ & $68.53$ & $79.11$ & $76.23$ & $70.01$ & $59.89$ & $44.54$ & $42.20$ & $62.93$ \\
\midrule
SmoothQuant & $8/8/8$ & $67.34$ & $75.19$ & $71.78$ & $69.06$ & $54.92$ & $40.44$ & $38.80$ & $59.65$ \\
FlexRound & $8/8/8$ & $68.78$ & $78.51$ & $75.23$ & $70.56$ & $58.46$ & $44.03$ & $41.00$ & $62.37$ \\
LRQ (Ours) & $8/8/8$ & $68.84$ & $78.78$ & $75.56$ & $70.80$ & $59.13$ & $44.62$ & $41.60$ & $\mathbf{62.76}$ \\
\midrule
Llama $33$B & $16/16/16$ & $68.38$ & $80.09$ & $79.21$ & $72.93$ & $58.92$ & $45.48$ & $42.00$ & $63.86$ \\
\midrule
SmoothQuant & $8/8/8$ & $71.31$ & $75.30$ & $71.29$ & $68.98$ & $53.66$ & $43.26$ & $41.00$ & $60.69$ \\
FlexRound & $8/8/8$ & $69.05$ & $79.49$ & $77.49$ & $70.88$ & $56.86$ & $43.60$ & $42.00$ & $62.77$ \\
LRQ (Ours) & $8/8/8$ & $68.84$ & $79.98$ & $78.52$ & $73.72$ & $58.21$ & $45.73$ & $43.00$ & $\mathbf{64.00}$ \\
\midrule
Llama $65$B & $16/16/16$ & $82.32$ & $80.85$ & $80.71$ & $77.19$ & $58.71$ & $46.33$ & $44.60$ & $67.24$ \\
\midrule
SmoothQuant & $8/8/8$ & $78.78$ & $79.54$ & $79.11$ & $73.32$ & $56.23$ & $45.90$ & $43.80$ & $65.24$ \\
FlexRound & $8/8/8$ & $80.46$ & $79.38$ & $79.23$ & $74.98$ & $57.20$ & $46.42$ & $45.00$ & $66.10$ \\
LRQ (Ours) & $8/8/8$ & $82.35$ & $81.12$ & $79.96$ & $75.61$ & $58.96$ & $46.59$ & $45.40$ & $\mathbf{67.14}$ \\
\bottomrule
\end{tabular}
}
\end{center}
\vskip -0.15in
\end{table*}

\subsection{Effect of Low-rank Matrices on Generalization of Quantized LLMs}\label{subsec:analysis}

Considering that a full weight-scaling matrix is substituted with a low-rank matrix as seen in Eq. \ref{eq:LRQ} derived from Eq. \ref{eq:flexround}, one might wonder (i) whether the minimization of block reconstruction error on calibration samples is feasible despite the use of low-rank matrices, and (ii) how the utilization of low-rank matrices can result in improved generalization performance on unseen benchmarks as Figure \ref{fig:smoothquant_sniper} demonstrates. To address these concerns, we conduct a comparative analysis of accumulated root mean square error (RMSE) between $\mW\mX$ and $\widehat{\mW}\widetilde{\mX}$ for RTN, FlexRound, and LRQ. 

For a calibration sample that is selected from the C4 dataset, even if both FlexRound and LRQ initially start their learning process from the same RTN baseline, LRQ achieves an almost identical accumulated RMSE to FlexRound, as illustrated in Figure \ref{fig2:a}. This observation underscores that the use of low-rank weight-scaling matrices does not pose any noticeable obstacle to the minimization of block reconstruction error on calibration data. For common sense reasoning and MMLU benchmarks that are unseen during the reconstruction stage, however, accumulated RMSE for LRQ is much smaller than that for FlexRound as well as RTN as described in Figure \ref{fig2:b}. This compelling result implies that harnessing the parameter-efficiency of low-rank matrices can facilitate superior generalization on unseen benchmarks. In light of these findings, the incorporation of low-rank matrices into block-wise reconstruction is indeed a pivotal step in enhancing the generalization capability of quantized LLMs. For visual representation across various samples, three figures are incorporated in Appendix \ref{appendix:rmse_three_samples}, each depicting the accumulated RMSE for three distinct samples. In addition, we illustrate the sensitivity of the accumulated RMSE to the number of calibration samples in Appendix \ref{appendix:sensitivity}.

\begin{table*}[t]
\vskip -0.2in
\caption{Zero-shot performance of Llama $2$ on common sense reasoning tasks (BoolQ, PIQA, HellaSwag, WinoGrande, ARC easy and challenge, and OpenBookQA) with per-channel asymmetric weight quantization, per-tensor asymmetric static activation quantization, and per-token asymmetric KV cache quantization. The accuracy ($\%$) is reported for all tasks. The number of bits used for weights, activations, and KV cache is $8$-bit.} 
\vskip -0.15in
\label{tab:llama2_csr_per_tensor}
\begin{center}
\small
\resizebox{\linewidth}{!}{
\begin{tabular}{lccccccccc}
\toprule
Method & \makecell{\# Bits (W/A/KV)} & BoolQ & PIQA & HellaSwag &  WinoGrande & ARC-e & ARC-c & OBQA & Average\\
\midrule
Llama $2$ $7$B & $16/16/16$ & $71.07$ & $76.99$ & $72.96$ & $67.25$ & $53.58$ & $40.53$ & $40.80$ & $60.45$ \\
\midrule
SmoothQuant & $8/8/8$ & $67.65$ & $73.29$ & $67.52$ & $62.90$ & $51.35$ & $37.80$ & $37.60$ & $56.87$ \\
FlexRound & $8/8/8$ & $72.05$ & $77.26$ & $71.30$ & $65.98$ & $54.88$ & $39.16$ & $39.20$ & $\mathbf{59.98}$ \\
LRQ (Ours) & $8/8/8$ & $67.86$ & $76.99$ & $71.97$ & $67.01$ & $54.71$ & $40.19$ & $40.00$ & $59.82$ \\
\midrule
Llama $2$ $13$B & $16/16/16$ & $69.02$ & $79.05$ & $76.62$ & $69.61$ & $57.95$ & $44.28$ & $42.00$ & $62.65$ \\
\midrule
SmoothQuant & $8/8/8$ & $63.55$ & $75.95$ & $70.99$ & $66.30$ & $53.96$ & $40.10$ & $40.60$ & $58.78$ \\
FlexRound & $8/8/8$ & $66.94$ & $79.00$ & $75.32$ & $69.38$ & $58.54$ & $42.92$ & $40.40$ & $61.79$ \\
LRQ (Ours) & $8/8/8$ & $68.59$ & $78.67$ & $75.83$ & $70.64$ & $58.16$ & $43.34$ & $39.80$ & $\mathbf{62.15}$ \\
\midrule
Llama $2$ $70$B & $16/16/16$ & $76.70$ & $80.85$ & $80.85$ & $76.95$ & $59.72$ & $47.95$ & $44.40$ & $66.77$ \\
\midrule
SmoothQuant & $8/8/8$ & $76.21$ & $76.55$ & $79.30$ & $74.11$ & $55.85$ & $46.25$ & $45.60$ & $64.84$ \\
FlexRound & $8/8/8$ & $76.18$ & $80.36$ & $79.09$ & $75.06$ & $60.10$ & $46.42$ & $43.80$ & $65.86$ \\
LRQ (Ours) & $8/8/8$ & $77.95$ & $81.23$ & $79.78$ & $74.82$ & $57.83$ & $46.33$ & $43.60$ & $\mathbf{65.93}$ \\
\bottomrule
\end{tabular}
}
\end{center}
\vskip -0.15in
\end{table*}

\begin{table}[t]
\vskip -0.05in
\caption{Five-shot accuracy of Llama on Massive Multitask Language Understanding with per-channel asymmetric weight quantization, per-tensor asymmetric static activation quantization, and per-token asymmetric KV cache quantization. The accuracy ($\%$) is reported for four disciplines. The number of bits used for weights, activations, and KV cache is $8$-bit, the same as Table \ref{tab:llama_csr_per_tensor}.} 
\vskip -0.15in
\label{tab:llama_mmlu_per_tensor}
\begin{center}
\small
\resizebox{\linewidth}{!}{
\begin{tabular}{lccccc}
\toprule
Method & STEM & Humanities & \makecell{Social \\ Science} & Other & Average\\
\midrule
Llama $7$B & $30.58$ & $33.88$ & $38.19$ & $38.25$ & $35.12$ \\
\midrule
SmoothQuant & $28.40$ & $28.69$ & $32.79$ & $30.48$ & $29.94$ \\
FlexRound & $27.60$ & $28.71$ & $29.61$ & $31.99$ & $29.43$ \\
LRQ (Ours) & $29.72$ & $32.79$ & $37.44$ & $38.16$ & $\mathbf{34.39}$\\
\midrule
Llama $13$B & $36.35$ & $44.97$ & $54.14$ & $53.15$ & $47.02$ \\
\midrule
SmoothQuant & $27.24$ & $30.12$ & $30.58$ & $31.31$ & $29.87$ \\
FlexRound & $33.63$ & $42.81$ & $48.65$ & $49.26$ & $43.60$ \\
LRQ (Ours) & $35.16$ & $44.55$ & $51.74$ & $52.04$ & $\mathbf{45.83}$\\
\midrule
Llama $33$B & $46.69$ & $56.39$ & $67.40$ & $63.60$ & $58.38$ \\
\midrule
SmoothQuant & $37.94$ & $41.64$ & $50.57$ & $51.48$ & $45.07$ \\
FlexRound & $43.47$ & $52.20$ & $61.94$ & $59.90$ & $54.24$ \\
LRQ (Ours) & $45.26$ & $52.58$ & $63.99$ & $61.26$ & $\mathbf{55.51}$\\
\midrule
Llama $65$B & $51.95$ & $61.87$ & $73.32$ & $67.58$ & $63.57$ \\
\midrule
SmoothQuant & $44.83$ & $50.82$ & $63.34$ & $57.09$ & $53.72$ \\
FlexRound & $46.32$ & $54.60$ & $65.06$ & $62.49$ & $56.94$ \\
LRQ (Ours) & $50.96$ & $61.28$ & $71.99$ & $66.66$ & $\mathbf{62.65}$ \\
\bottomrule
\end{tabular}
}
\end{center}
\vskip -0.2in
\end{table}

\begin{table}[t]
\vskip -0.05in
\caption{Five-shot accuracy of Llama $2$ on Massive Multitask Language Understanding with per-channel asymmetric weight quantization, per-tensor asymmetric static activation quantization, and per-token asymmetric KV cache quantization. The accuracy ($\%$) is reported for four disciplines. The number of bits used for weights, activations, and KV cache is $8$-bit, the same as Table \ref{tab:llama2_csr_per_tensor}.} 
\vskip -0.15in
\label{tab:llama2_mmlu_per_tensor}
\begin{center}
\small
\resizebox{\linewidth}{!}{
\begin{tabular}{lccccc}
\toprule
Method & STEM & Humanities & \makecell{Social \\ Science} & Other & Average\\
\midrule
Llama $2$ $7$B & $37.04$ & $43.38$ & $51.84$ & $52.44$ & $45.96$ \\
\midrule
SmoothQuant & $30.42$ & $27.95$ & $34.29$ & $34.27$ & $31.33$ \\
FlexRound & $33.40$ & $36.96$ & $43.13$ & $46.30$ & $39.70$ \\
LRQ (Ours) & $34.82$ & $39.91$ & $46.47$ & $47.62$ & $\mathbf{42.04}$\\
\midrule
Llama $2$ $13$B & $44.27$ & $54.43$ & $63.41$ & $60.76$ & $55.68$ \\
\midrule
SmoothQuant & $30.98$ & $29.29$ & $35.36$ & $35.29$ & $32.37$ \\
FlexRound & $41.09$ & $51.58$ & $61.39$ & $59.41$ & $53.28$ \\
LRQ (Ours) & $42.88$ & $51.97$ & $62.14$ & $59.93$ & $\mathbf{54.08}$\\
\midrule
Llama $2$ $70$B & $57.79$ & $65.16$ & $80.44$ & $74.61$ & $69.11$ \\
\midrule
SmoothQuant & $47.51$ & $53.84$ & $68.35$ & $63.94$ & $57.99$ \\
FlexRound & $54.27$ & $61.11$ & $77.45$ & $71.31$ & $65.57$ \\
LRQ (Ours) & $54.44$ & $62.61$ & $76.99$ & $71.78$ & $\mathbf{66.12}$ \\
\bottomrule
\end{tabular}
}
\end{center}
\vskip -0.2in
\end{table}

\begin{table*}[t]
\vskip -0.2in
\caption{Zero-shot performance of Llama $2$ on common sense reasoning tasks (BoolQ, PIQA, HellaSwag, WinoGrande, ARC easy and challenge, and OpenBookQA) with per-channel asymmetric weight quantization, per-token asymmetric activation quantization, and per-token asymmetric KV cache quantization. The accuracy ($\%$) is reported for all tasks. Here, weights are quantized to $4$-bits, while activations and KV cache utilize $8$-bit quantization. } 
\vskip -0.15in
\label{tab:llama2_csr_per_token}
\begin{center}
\small
\resizebox{\linewidth}{!}{
\begin{tabular}{lccccccccc}
\toprule
Method & \makecell{\# Bits (W/A/KV)} & BoolQ & PIQA & HellaSwag &  WinoGrande & ARC-e & ARC-c & OBQA & Average\\
\midrule
Llama $2$ $7$B & $16/16/16$ & $71.07$ & $76.99$ & $72.96$ & $67.25$ & $53.58$ & $40.53$ & $40.80$ & $60.45$ \\
\midrule
SmoothQuant & $4/8/8$ & $43.03$ & $63.71$ & $41.08$ & $54.30$ & $35.69$ & $27.99$ & $32.60$ & $42.63$  \\
FlexRound & $4/8/8$ & $71.71$ & $76.77$ & $72.24$ & $66.14$ & $53.49$ & $40.02$ & $40.40$ & $60.11$ \\
LRQ (Ours) & $4/8/8$ & $73.00$ & $76.99$ & $71.90$ & $65.98$ & $54.38$ & $39.68$ & $41.20$ & $\mathbf{60.45}$ \\
\midrule
Llama $2$ $13$B & $16/16/16$ & $69.02$ & $79.05$ & $76.62$ & $69.61$ & $57.95$ & $44.28$ & $42.00$ & $62.65$ \\
\midrule
SmoothQuant & $4/8/8$ & $61.62$ & $56.53$ & $37.31$ & $51.38$ & $31.57$ & $24.74$ & $30.60$ & $41.96$  \\
FlexRound & $4/8/8$ & $69.05$ & $78.51$ & $75.51$ & $69.53$ & $58.75$ & $43.60$ & $41.20$ & $62.31$ \\
LRQ (Ours) & $4/8/8$ & $71.13$ & $78.29$ & $75.79$ & $68.90$ & $57.83$ & $43.34$ & $41.20$ & $\mathbf{62.35}$ \\
\midrule
Llama $2$ $70$B & $16/16/16$ & $76.70$ & $80.85$ & $80.85$ & $76.95$ & $59.72$ & $47.95$ & $44.40$ & $66.77$ \\
\midrule
SmoothQuant & $4/8/8$ & $50.46$ & $71.60$ & $48.35$ & $55.09$ & $44.87$ & $32.17$ & $37.40$ & $48.56$  \\
FlexRound & $4/8/8$ & $77.31$ & $80.96$ & $79.89$ & $75.30$ & $60.19$ & $48.21$ & $43.40$ & $66.47$ \\
LRQ (Ours) & $4/8/8$ & $77.92$ & $81.28$ & $80.42$ & $75.06$ & $60.94$ & $48.04$ & $42.60$ & $\mathbf{66.61}$ \\
\bottomrule
\end{tabular}
}
\end{center}
\vskip -0.15in
\end{table*}

\section{Experiments}\label{sec:experiments}

In this section, we first explore the influence of the rank $r$ in Eq. \ref{eq:LRQ} and the quantity of calibration samples on the performance of LRQ. Next, to verify the effectiveness of LRQ, we compare LRQ with existing state-of-the-art post-training quantization (PTQ) methods for large language models (LLMs).

\begin{figure}[t]
    \centering
    \subfigure[Rank Study]
    {
    \includegraphics[width=1.0\linewidth]{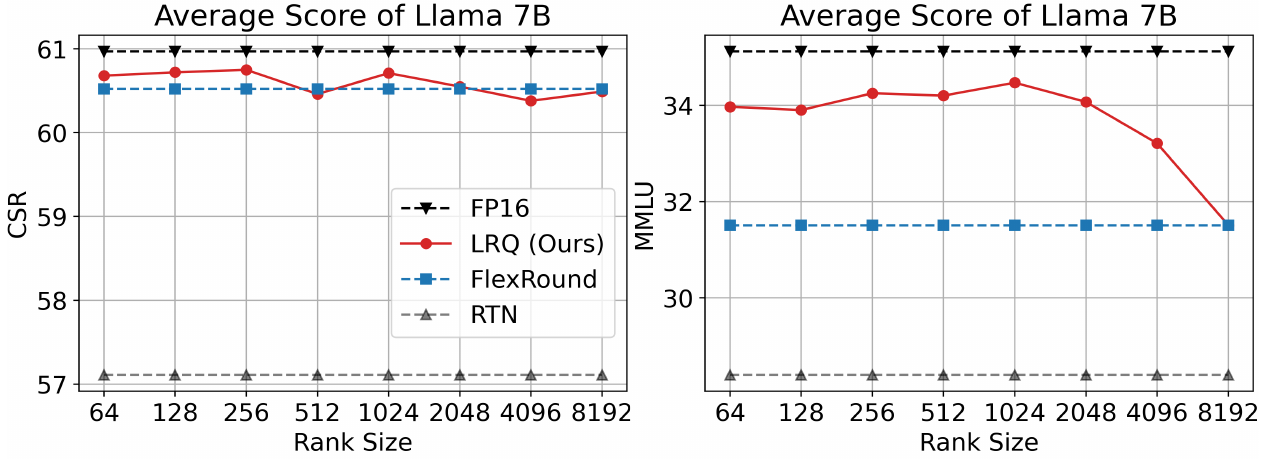}
    \label{fig4:a}
    }
    \vskip -0.05in
    \subfigure[Calibration Sample Size Study] 
    {
    \includegraphics[width=1.0\linewidth]{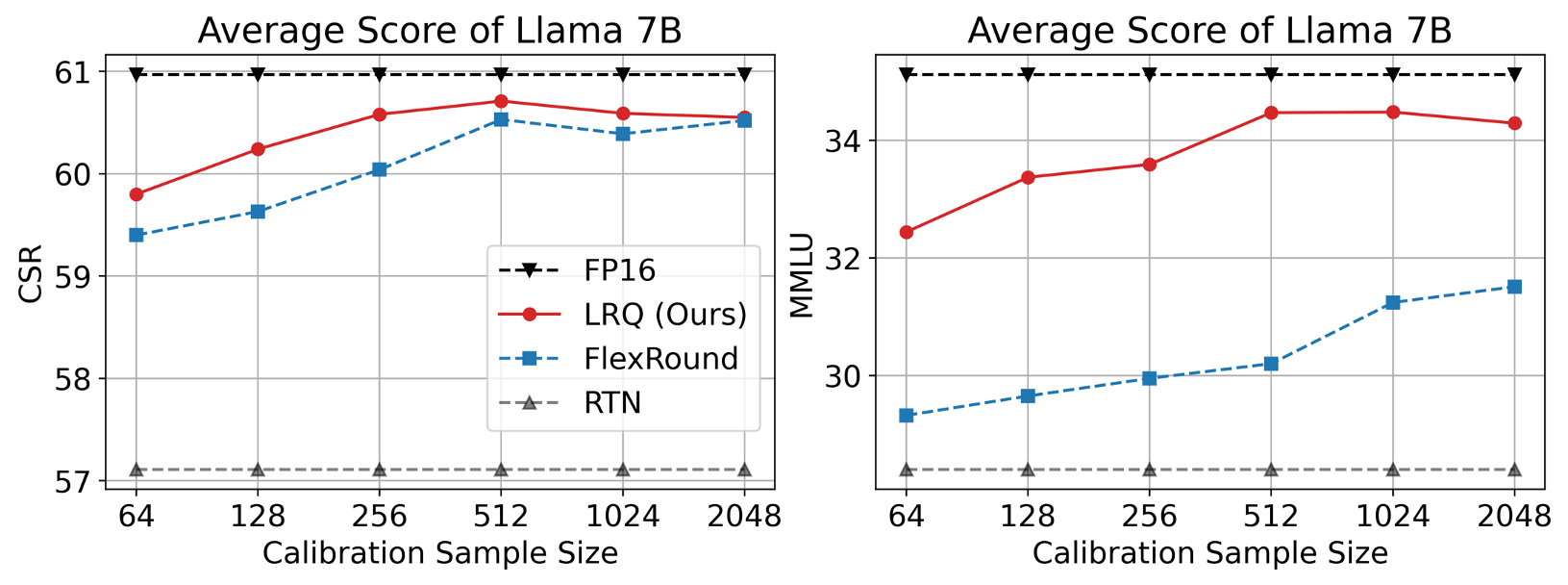}
        \label{fig4:b}
    }
    \vskip -0.1in
    \caption{Zero-shot and five-shot performances of Llama $7$B on common sense reasoning (CSR) tasks and MMLU, where weights and activations are quantized to $8$-bit while the KV cache is kept in FP16.} 
    \label{fig:ablations}
    \vskip -0.2in
\end{figure}

\begin{table}[t]
\caption{Five-shot accuracy of Llama $2$ on Massive Multitask Language Understanding with per-channel asymmetric weight quantization, per-token asymmetric activation quantization, and per-token asymmetric KV cache quantization. The accuracy ($\%$) is reported for four disciplines. Here, weights are quantized to $4$-bits, while activations and KV cache utilize $8$-bit quantization, which is the same as Table \ref{tab:llama2_csr_per_token}.} 
\vskip -0.15in
\label{tab:llama2_mmlu_per_token}
\begin{center}
\small
\resizebox{\linewidth}{!}{
\begin{tabular}{lccccc}
\toprule
Method & STEM & Humanities & \makecell{Social \\ Science} & Other & Average\\
\midrule
Llama $2$ $7$B & $37.04$ & $43.38$ & $51.84$ & $52.44$ & $45.96$ \\
\midrule
SmoothQuant & $26.77$ & $24.87$ & $22.81$ & $25.85$ & $25.05$ \\
FlexRound & $37.81$ & $42.55$ & $50.47$ & $50.65$ & $45.14$ \\
LRQ (Ours) & $36.88$ & $42.53$ & $50.80$ & $52.22$ & $\mathbf{45.36}$ \\
\midrule
Llama $2$ $13$B & $44.27$ & $54.43$ & $63.41$ & $60.76$ & $55.68$ \\
\midrule
SmoothQuant & $27.07$ & $24.25$ & $25.22$ & $26.43$ & $25.57$ \\
FlexRound & $42.88$ & $50.71$ & $61.94$ & $59.93$ & $53.77$\\
LRQ (Ours) & $43.90$ & $52.56$ & $62.07$ & $59.96$ & $\mathbf{54.49}$  \\
\midrule
Llama $2$ $70$B & $57.79$ & $65.16$ & $80.44$ & $74.61$ & $69.11$ \\
\midrule
SmoothQuant & $27.37$ & $24.59$ & $27.59$ & $25.94$ & $26.16$ \\
FlexRound & $56.26$ & $62.89$ & $78.78$ & $72.92$ & $67.26$ \\
LRQ (Ours) & $55.57$ & $64.65$ & $78.97$ & $72.52$ & $\mathbf{67.65}$ \\
\bottomrule
\end{tabular}
}
\end{center}
\vskip -0.2in
\end{table}

\begin{table*}[t]
\vskip -0.2in
\caption{Zero-shot performance of Llama $2$ on common sense reasoning tasks (BoolQ, PIQA, HellaSwag, WinoGrande, ARC easy and challenge, and OpenBookQA) and the causal language modeling task on WikiText2 with per-channel asymmetric weight-only quantization. The accuracy ($\%$) and the perplexity (PPL) are reported for common sense reasoning tasks and the causal language modeling task, respectively. The lower PPL, the better.} 
\vskip -0.15in
\label{tab:llama2_weight_only}
\begin{center}
\small
\resizebox{\linewidth}{!}{
\begin{tabular}{lcccccccccc}
\toprule
Method & \makecell{\# Bits (W/A/KV)} & BoolQ & PIQA & HellaSwag &  WinoGrande & ARC-e & ARC-c & OBQA & Average & WikiText2\\
\midrule
Llama $2$ $7$B & $16/16/16$ & $71.07$ & $76.99$ & $72.96$ & $67.25$ & $53.58$ & $40.53$ & $40.80$ & $60.45$ & $5.47$\\
\midrule
OmniQuant & $3/16/16$ & $65.72$ & $73.99$ & $67.65$ & $63.61$ & $49.71$ & $36.26$ & $39.80$ & $56.58$ & $6.57$ \\
FlexRound & $3/16/16$ & $70.15$ & $75.73$ & $69.92$ & $66.46$ & $51.43$ & $38.31$ & $39.20$ & $58.74$ & $\mathbf{6.34}$ \\
LRQ (Ours) & $3/16/16$ & $71.31$ & $76.44$ & $70.35$ & $64.88$ & $52.23$ & $39.08$ & $39.20$ & $\mathbf{59.07}$ & $6.48$  \\
\hdashline
OmniQuant & $4/16/16$ & $68.99$ & $77.48$ & $71.26$ & $67.01$ & $53.66$ & $39.08$ & $40.00$ & $59.64$ & $\mathbf{5.74}$ \\
FlexRound & $4/16/16$ & $73.24$ & $76.55$ & $72.09$ & $67.09$ & $52.82$ & $39.51$ & $40.80$ & $60.30$ & $5.83$ \\
LRQ (Ours) & $4/16/16$ & $71.80$ & $76.88$ & $72.40$ & $67.88$ & $53.24$ & $40.27$ & $40.20$ & $\mathbf{60.38}$ & $5.75$ \\
\midrule
Llama $2$ $13$B & $16/16/16$ & $69.02$ & $79.05$ & $76.62$ & $69.61$ & $57.95$ & $44.28$ & $42.00$ & $62.65$ & $4.88$ \\
\midrule
OmniQuant & $3/16/16$ & $69.02$ & $77.69$ & $72.77$ & $65.90$ & $54.00$ & $42.75$ & $38.80$ & $60.13$ & $5.58$ \\
FlexRound & $3/16/16$ & $66.02$ & $78.29$ & $74.41$ & $67.32$ & $57.15$ & $42.15$ & $40.60$ & $60.85$ & $5.59$ \\
LRQ (Ours) & $3/16/16$ & $67.49$ & $78.45$ & $74.25$ & $69.30$ & $56.23$ & $42.58$ & $41.60$ & $\mathbf{61.41}$ & $\mathbf{5.57}$ \\
\hdashline
OmniQuant & $4/16/16$ & $65.17$ & $78.94$ & $75.39$ & $67.80$ & $56.44$ & $42.75$ & $41.60$ & $61.16$ & $5.02$ \\
FlexRound & $4/16/16$ & $69.94$ & $79.00$ & $75.93$ & $69.06$ & $58.96$ & $43.26$ & $40.40$ & $62.36$ & $\mathbf{5.01}$ \\
LRQ (Ours) & $4/16/16$ & $70.49$ & $78.78$ & $76.13$ & $69.93$ & $59.85$ & $43.52$ & $41.20$ & $\mathbf{62.84}$ & $5.02$ \\
\midrule
Llama $2$ $70$B & $16/16/16$ & $76.70$ & $80.85$ & $80.85$ & $76.95$ & $59.72$ & $47.95$ & $44.40$ & $66.77$ & $3.31$ \\
\midrule
OmniQuant & $3/16/16$ & $66.54$ & $80.74$ & $78.13$ & $73.48$ & $57.20$ & $45.90$ & $42.60$ & $63.51$ & $3.93$ \\
FlexRound & $3/16/16$ & $75.02$ & $80.25$ & $79.02$ & $70.24$ & $58.63$ & $46.93$ & $42.20$ & $64.61$ & $3.92$ \\
LRQ (Ours) & $3/16/16$ & $78.13$ & $80.47$ & $79.54$ & $75.37$ & $59.55$ & $46.84$ & $43.60$ & $\mathbf{66.21}$ & $\mathbf{3.89}$ \\
\hdashline
OmniQuant & $4/16/16$ & $77.13$ & $80.96$ & $80.53$ & $75.85$ & $59.76$ & $46.76$ & $42.40$ & $66.20$ & $3.47$ \\
FlexRound & $4/16/16$ & $78.38$ & $80.58$ & $80.49$ & $75.85$ & $59.51$ & $47.87$ & $43.60$ & $66.61$ & $\mathbf{3.45}$ \\
LRQ (Ours) & $4/16/16$ & $79.72$ & $80.36$ & $80.58$ & $77.03$ & $59.72$ & $47.44$ & $43.80$ & $\mathbf{66.95}$ & $3.47$ \\
\bottomrule
\end{tabular}
}
\end{center}
\vskip -0.2in
\end{table*}

\begin{table}[t]
\vskip -0.01in
\caption{Average zero-shot accuracy and perplexity of Llama $3$ $8$B on common sense reasoning tasks (PIQA, HellaSwag, WinoGrande, ARC easy and challenge) and WikiText2 respectively, using $4$-bit per-channel asymmetric weight-only quantization. The lower PPL, the better. More details are given in Table~\ref{tab:llama3_8b_appendix}. The results of GPTQ, AWQ, and QuIP come from \citet{huang2024empiricalstudyllama3quantization}.} 
\vskip -0.15in
\label{tab:llama3_8b}
\begin{center}
\small
\resizebox{\linewidth}{!}{
\begin{tabular}{lccc} 
\toprule
Method & \makecell{\# Bits (W/A/KV)} & Average & WikiText2 \\
\midrule
Llama $3$ $8$B & $16/16/16$ & $68.6$ & $6.1$ \\
\midrule
GPTQ & $4/16/16$ & $64.8$ & $7.0$ \\
AWQ & $4/16/16$ & $67.0$ & $7.1$ \\
QuIP & $4/16/16$ & $67.1$ & $\mathbf{6.5}$ \\
FlexRound & $4/16/16$ & $67.8$ & $6.9$ \\
LRQ (Ours) & $4/16/16$ & $\mathbf{68.0}$ & $6.9$ \\
\bottomrule
\end{tabular}
}
\end{center}
\vskip -0.21in
\end{table}

We use just a single NVIDIA A100-80GB GPU to quantize LLMs via LRQ. We randomly choose $512$ calibration samples with a token length of $1024$ from the training set of C4 \citep{raffel2020c4}. Unless otherwise mentioned, LRQ is applied to all linear layers in both attention and feed-forward modules, and the rank $r$ in Eq. \ref{eq:LRQ} is set to $2048$ for large language models beyond $30$B parameters or to $1024$ for smaller models so as to reduce the number of learnable parameters by approximately half compared to FlexRound. The exact ratio of the number of learnable parameters in LRQ to the number of pre-trained weights for an intermediate Transformer block of each Llama model is given in Appendix \ref{appendix:ratio}.
Quantized models are evaluated on MMLU \citep{hendrycks2021measuring} in the five-shot setting or common sense reasoning benchmarks: BoolQ \citep{clark2019boolq}, PIQA \citep{bisk2020piqa}, HellaSwag \citep{zellers2019hellaswag}, WinoGrande \citep{sakaguchi2021winogrande}, ARC easy and challenge \citep{clark2018arc}, and OpenBookQA \citep{mihaylov2018obqa} in the zero-shot setting. More details are deferred to Appendix \ref{appendix:detail}.

\subsection{Ablation Study}

\paragraph{Rank Study} To examine the impact of the rank $r$ in Eq. \ref{eq:LRQ} on the generalization on unseen benchmarks, we compare LRQ with different $r$ (spanning from $64$ to $8192$) to FlexRound for Llama $7$B as shown in Figure \ref{fig4:a}. 
The performance of LRQ (depicted by the red curve) either remains relatively stable (the left side of Figure \ref{fig4:a}) or increases gradually from $33.97\%$ to $34.47\%$ (the right side of Figure \ref{fig4:a}) with the rise in the rank $r$ from $64$ to $1024$. As the rank $r$ continuously increases from $2048$ to $8192$, however, the performance of LRQ eventually declines to match that of FlexRound (indicated by the blue curve) on both common sense reasoning tasks and MMLU, which leads us to set the rank $r$ to $1024$ for Llama 7B. Hence, selecting an appropriate low rank $r$ becomes crucial to enable quantized LLMs via LRQ to achieve well-rounded performance across various tasks encompassing common sense reasoning and MMLU benchmarks. 

\paragraph{Calibration Sample Size Study} To identify whether the performance of LRQ improves with an increase in the number of calibration samples, we also conduct experiments on LRQ for Llama $7$B with various calibration sample size while fixing the rank $r$ to $1024$. The accuracy of LRQ rises with a larger calibration sample size, but it reaches a saturation point when exceeding $1024$ calibration samples as depicted in Figure \ref{fig4:b}. Nevertheless, LRQ can surpass FlexRound irrespective of the calibration sample size not only on common sense reasoning tasks but also on the MMLU benchmark, which sheds light on the effect of low-rank matrices on enhancing the generalization ability of quantized LLMs as we elaborate on in Section \ref{subsec:analysis}.  


\subsection{Per-tensor Static Activation Quantization}\label{subsec:per-tensor} 

As meticulously studied in \citet{xiao2022smoothquant}, per-tensor static activation quantization is hardware-efficient and can be implemented on off-the-shelf GPUs with FasterTransformer, the state-of-the-art Transformer inference framework provided from NVIDIA, to achieve up to $1.5\times$ inference speed-up and almost halving the memory footprint compared to FP16 baselines. Accordingly, we employ per-tensor asymmetric static activation quantization as well as per-channel asymmetric weight quantization. Moreover, we also quantize the KV cache to $8$-bit with a per-token asymmetric quantization scheme. It is worth noting that for large batch sizes, the KV cache can consume a much larger amount of memory than the model size, thus causing a bottleneck in high-throughput LLM inference. Fortunately, the performance discrepancy before and after per-token asymmetric KV cache quantization is almost insignificant no matter which quantization method is selected, as presented in Appendix \ref{appendix:exp}. For this reason, we also additionally utilize
per-token asymmetric KV cache quantization. 
Further experimental details are provided in Appendix \ref{appendix:detail}.

Table \ref{tab:llama_csr_per_tensor}, \ref{tab:llama2_csr_per_tensor}, \ref{tab:llama_mmlu_per_tensor}, and \ref{tab:llama2_mmlu_per_tensor} reveal the efficacy of LRQ on common sense reasoning tasks and MMLU. For common sense reasoning tasks, the zero-shot accuracy of LRQ is almost close to that of FP16 baselines, being superior to that of both SmoothQuant and FlexRound for most of the Llama and Llama $2$ models. Not only that, LRQ also considerably outperforms SmoothQuant and FlexRound on MMLU.

\subsection{Per-token Activation Quantization}\label{subsec:per-token} 

Although LRQ shows better performance than SmoothQuant and FlexRound on both common sense reasoning tasks and MMLU when employing per-tensor asymmetric static activation quantization, there is still the five-shot accuracy gap on MMLU between LRQ and FP16 baselines for Llama $2$ as in Table \ref{tab:llama2_mmlu_per_tensor}. Thus, we also conduct experiments on Llama $2$ with a per-token asymmetric activation quantization scheme. More details about experimental settings are given in Appendix \ref{appendix:detail}.

In Table \ref{tab:llama2_csr_per_token} and \ref{tab:llama2_mmlu_per_token}, 
when quantizing weights to $4$-bit and both activations and KV cache to $8$-bit, LRQ can attain similar zero-shot performance to FP16 baselines on common sense reasoning benchmarks and narrow the five-shot performance difference between FP16 baselines and quantized models to less than $1.5$ percent on the MMLU benchmark. 

\begin{figure}[t]
    \vskip -0.15in
    \centering
    \includegraphics[width=0.99\linewidth]{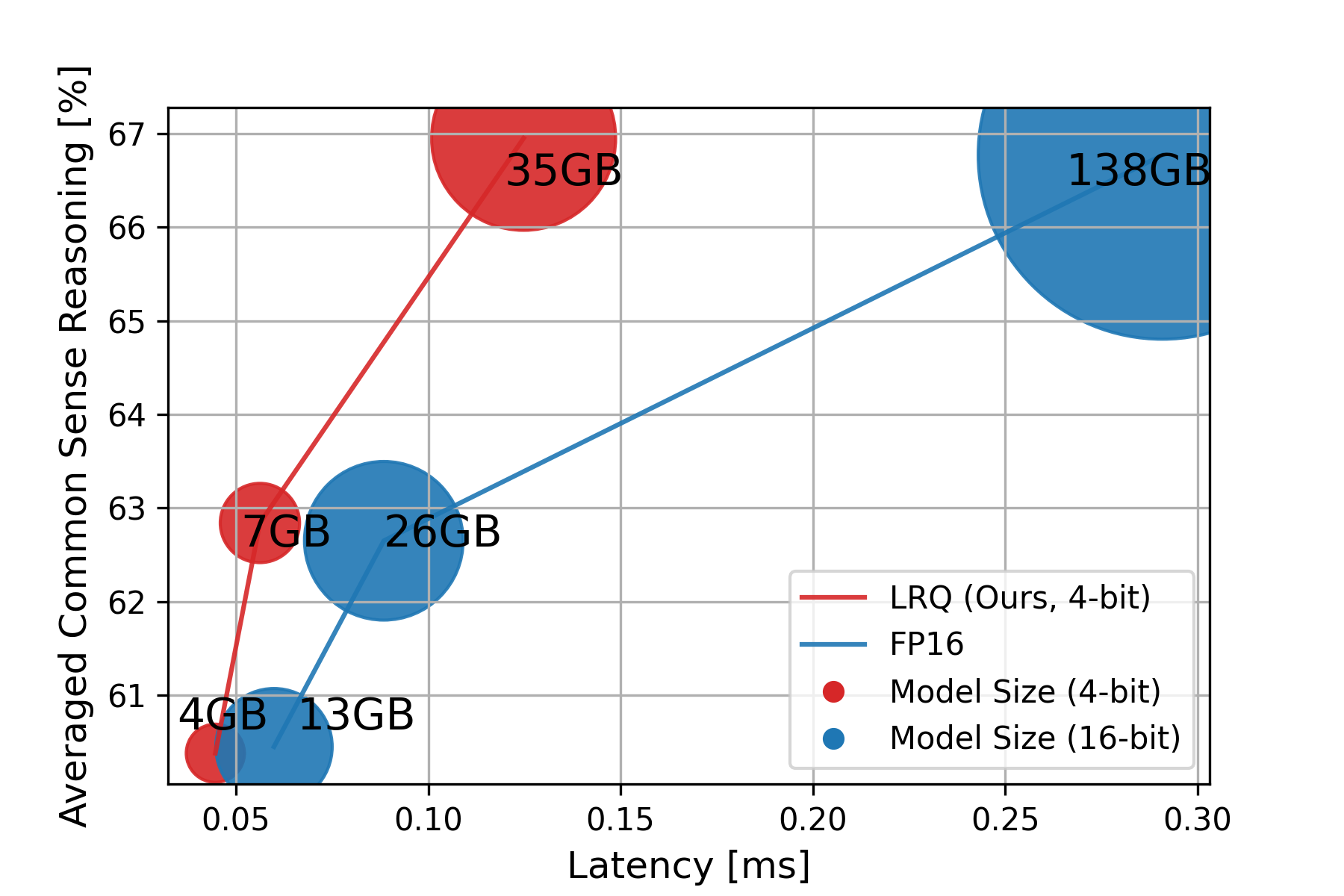}
    \vskip -0.15in
    \caption{Average zero-shot accuracy over latency for Llama $2$ $7$B, $13$B, and $70$B, respectively. The blue expresses FP16 baselines while the red represents 4-bit quantized models via LRQ. The size of a circle indicates the model size. More details are given in Appendix~\ref{appendix:test}.} 
    \label{fig:latency}
    \vskip -0.2in
\end{figure}

\subsection{Per-channel Weight-only Quantization}

As LRQ is designed as a post-training weight quantization technique for LLMs, we also run experiments on weight-only quantization for Llama $2$ and Llama $3$ $8$B \citep{dubey2024llama3herdmodels} on common sense reasoning tasks and WikiText2 \citep{merity2016pointer}. In Table \ref{tab:llama2_weight_only} and \ref{tab:llama3_8b}, we use per-channel weight-only quantization instead of group-wise weight-only quantization. Table \ref{tab:llama2_weight_only} and \ref{tab:llama3_8b} show that the average zero-shot accuracy of LRQ is consistently higher than that of all PTQ methods on common sense reasoning tasks. Specifically, in Table \ref{tab:llama2_weight_only}, LRQ can perform closely to FP16 baselines on common sense reasoning tasks even with 3-bit per-channel weight-only quantization. Despite the fact that Llama $3$ is hard to quantize, Table \ref{tab:llama3_8b} exhibits that LRQ attains the smallest accuracy drop (less than $0.7$ percent compared to the FP16 baseline) on common sense reasoning tasks. 
In addition, Figure \ref{fig:latency} displays compression and acceleration after quantizing Llama 2 to 4-bit via LRQ, showing that LRQ can perform comparably to FP16 baselines while reducing both latency and model size noticeably.

\section{Conclusion}
We propose a simple yet effective post-training weight quantization approach for LLMs, LRQ that learns low-rank weight-scaling matrices for block-by-block reconstructing the outputs of an intermediate Transformer block. LRQ can decrease the number of learnable parameters effectively while allowing for scaling weights individually, thereby enhancing the generalization of quantized LLMs. 



\clearpage
\section*{Limitations}

To push the limit of post-training weight-activation quantization, two research directions emerge: (i) $4$-bit weight and $8$-bit activation quantization, and (ii) INT4 weight-activation quantization. As explained in Appendix \ref{appendix:related}, \citet{lee2023aqas} attempted to quantize LLMs with $4$-bit weight and $8$-bit activation quantization, whereas \citet{wei2023outlier} and \citet{shao2024omniquant} strived to quantize LLMs with INT6 and even INT4 weight-activation quantization. In this paper, we only deal with the former quantization scheme, $4$-bit weight and $8$-bit activation quantization. 

Like \citet{wei2023outlier} and \citet{shao2024omniquant}, we could also focus on INT4 weight-activation quantization rather than $4$-bit weight and $8$-bit activation quantization in Section \ref{subsec:per-token}. However, \citet{liu2023llmqat}, an earlier LLM quantization work than \citet{wei2023outlier} and \citet{shao2024omniquant}, already exhibited the non-marginal accuracy degradation of $4$-bit weight and $8$-bit activation quantization despite the fact that \citet{liu2023llmqat} exploited quantization-aware training, not post-training quantization. Furthermore, in terms of serving throughput, \citet{lin2024qservew4a8kv4quantizationcodesign} shows the superiority of $4$-bit weight and $8$-bit activation quantization over INT4 weight-activation quantization as well as INT8 weight-activation quantization. For these reasons, we prioritize $4$-bit weight and $8$-bit activation quantization over INT4 weight-activation quantization in this paper. 



\bibliography{custom}

\appendix
\onecolumn

\section{Related Work}\label{appendix:related}

Quantization works can be generally categorized into quantization-aware training (QAT) and post-training quantization (PTQ). As QAT can maintain the performance of FP32/FP16 baselines, QAT has been applied to computer vision models \citep{jung2019learning, esser2020learned, lee2021cluster}. Notwithstanding, there exist many challenges associated with applying QAT to large language models (LLMs) due to the sheer scale of pre-training data and a huge amount of computational resources required for training on the whole pre-training dataset. Although \citet{liu2023llmqat} presented the possibility of applying QAT to LLMs, unfortunately, they did not perform experiments on Llama $65$B, the largest and best performing model among the Llama models, in spite of using a single $8$-GPU node. \citet{kim2023peqa} introduced Parameter-Efficient and Quantization-aware Adaptation (PEQA) that fine-tunes only quantization step sizes. While PEQA consumes less computational resources and can perform better than QAT, it still requires a single $8$-GPU node to quantize Llama $65$B. 

As \citet{frantar2023optq} demonstrated the application of PTQ to LLMs only with a single GPU, many researchers have recently paid attention to PTQ for LLMs. LLM PTQ can be classified into two categories: LLM weight-only quantization \citep{frantar2023optq, lin2023awq, chee2023quip} and LLM weight-activation quantization \citep{dettmers2022llm, yao2022zeroquant, xiao2022smoothquant, lee2023flexround, wei2023outlier, shao2024omniquant}. For the former quantization, \citet{frantar2023optq, chee2023quip} quantized the weights of LLMs into low-bit integers based on layer-wise reconstruction, whereas \citet{lin2023awq} did by not counting on reconstruction but per-channel scaling in consideration of both weight and activation magnitudes. Despite the fact that all these studies exhibited decent quantization performance, the main benefit of weight-only quantization does not align with serving LLMs with high throughput as delineated in Section \ref{sec:intro}. In this light, we concentrates on weight-activation quantization.

When it comes to weight-activation quantization, \citet{yao2022zeroquant} presented ZeroQuant with a $8$-bit group-wise weight quantization scheme and a $8$-bit per-token activation quantization scheme based on layer-wise knowledge distillation, and \citet{dettmers2022llm} proposed LLM.int8() with a $8$-bit per-channel weight quantization scheme and a $8$-bit per-token activation quantization scheme while keeping activation outliers in FP16. As discussed in \citet{xiao2022smoothquant}, however, ZeroQuant incurs severe accuracy degradation for an open-source LLM, and the inference latency of LLM.int8() can be higher than that of the FP16 baseline. To deal with both issues, \citet{xiao2022smoothquant} devised SmoothQuant that can preserve the accuracy of OPT \citep{zhang2022opt} by easing the difficulty of activation quantization and accelerate LLM inference by up to $1.5$ times. Yet, SmoothQuant suffers from non-negligible performance degradation for other open-source models such as Llama and Llama $2$ with a $8$-bit per-tensor static activation quantization scheme as illustrated in Figure \ref{fig:smoothquant_sniper}. FlexRound that \citet{lee2023flexround} created showed the experimental results of Llama up to $33$B with a $8$-bit per-channel weight quantization scheme and a $8$-bit per-tensor static activation quantization scheme, but FlexRound incurs considerable performance degradation on the massive multitask language understanding (MMLU) benchmark as described in Figure \ref{fig1:b}. Beyond INT8 weight-activation quantization, \citet{lee2023aqas} attempted to quantize LLMs with $4$-bit weight and $8$-bit activation quantization, whereas \citet{wei2023outlier} and \citet{shao2024omniquant} strived to quantize LLMs with INT6 and even INT4 weight-activation quantization. However, in terms of serving throughput, \citet{lin2024qservew4a8kv4quantizationcodesign} showed the superiority of $4$-bit weight and $8$-bit activation quantization over INT4 weight-activation quantization as well as INT8 weight-activation quantization.

\clearpage
\section{Effect of \texorpdfstring{$\vr_2$}{} and \texorpdfstring{$\vc_2$}{} in LRQ}\label{appendix:r2_and_c2}

To show the effect of $\vr_2$ and $\vc_2$ in Eq. \ref{eq:LRQ}, we compare FlexRound, FlexRound with $\mS_2 = \mL_2 \mU_2$, and LRQ for Llama $7$B and $13$B.

\begin{table}[h]
\caption{Zero-shot performance of FlexRound, FlexRound with $\mS_2 = \mL_2 \mU_2$, and LRQ on common sense reasoning tasks (BoolQ, PIQA, HellaSwag, WinoGrande, ARC easy and challenge, and OpenBookQA) with per-channel asymmetric weight quantization, per-tensor asymmetric static activation quantization, and per-token asymmetric KV cache quantization (if applied). The number of bits used for weights, activations, and KV cache is expressed as W/A/KV.} 
\label{tab:ablation_L2U2_csr}
\begin{center}
\small
\begin{tabular}{lccc}
\toprule
Method & \makecell{\# Bits (W/A/KV)} & Llama $7$B & Llama $13$B\\
\midrule
FlexRound & $8/8/16$ & $60.53$ & $62.40$ \\
FlexRound with $\mS_2 = \mL_2 \mU_2$ & $8/8/16$ & $60.69$ & $62.62$ \\
LRQ (Ours) & $8/8/16$ & $\mathbf{60.71}$ & $\mathbf{62.92}$ \\
\hdashline
FlexRound & $8/8/8$ & $60.35$ & $62.37$ \\
FlexRound with $\mS_2 = \mL_2 \mU_2$ & $8/8/8$ & $60.49$ & $62.62$ \\
LRQ (Ours) & $8/8/8$ & $\mathbf{60.63}$ & $\mathbf{62.76}$ \\
\bottomrule
\end{tabular}
\end{center}
\end{table}

\begin{table}[h]
\caption{Five-shot performance of FlexRound, FlexRound with $\mS_2 = \mL_2 \mU_2$, and LRQ on Massive Multitask Language Understanding with per-channel asymmetric weight quantization, per-tensor asymmetric static activation quantization, and per-token asymmetric KV cache quantization (if applied). The number of bits used for weights, activations, and KV cache is expressed as W/A/KV.} 
\label{tab:ablation_L2U2_mmlu}
\begin{center}
\small
\begin{tabular}{lccc}
\toprule
Method & \makecell{\# Bits (W/A/KV)} & Llama $7$B & Llama $13$B\\
\midrule
FlexRound & $8/8/16$ & $30.20$ & $43.82$ \\
FlexRound with $\mS_2 = \mL_2 \mU_2$ & $8/8/16$ & $33.86$ & $45.48$ \\
LRQ (Ours) & $8/8/16$ & $\mathbf{34.47}$ & $\mathbf{45.83}$ \\
\hdashline
FlexRound & $8/8/8$ & $29.43$ & $43.60$ \\
FlexRound with $\mS_2 = \mL_2 \mU_2$ & $8/8/8$ & $33.96$ & $45.21$ \\
LRQ (Ours) & $8/8/8$ & $\mathbf{34.39}$ & $\mathbf{45.83}$ \\
\bottomrule
\end{tabular}
\end{center}
\end{table}

As evident from the tables above, FlexRound with $\mS_2 = \mL_2 \mU_2$ surpasses the performance of FlexRound but falls short of LRQ, which implies that the effect of $\vr_2$ and $\vc_2$ cannot be ignored. It is noteworthy that the five-shot accuracy on MMLU can witness an increase ranging from $1.5\%$ to $4\%$ by simply substituting $\mS_2$ with $\mL_2 \mU_2$, which corroborates the significance of leveraging the parameter-efficiency inherent in low-rank weight-scaling matrices.

\clearpage
\section{Figures of Accumulated RMSE on Assorted Samples}\label{appendix:rmse_three_samples}

\begin{figure}[h]
    \vskip -0.135in
    \centering
    \subfigure[Calibration sample] 
    {
    \includegraphics[width=0.24\linewidth]{Figures/Fig2a.png}
    
    }
    \subfigure[Unseen sample]
    {
    \includegraphics[width=0.455\linewidth]{Figures/Fig2b.png}

    }
    \vskip -0.1in
    \centering
    \subfigure[Calibration sample] 
    {
    \includegraphics[width=0.24\linewidth]{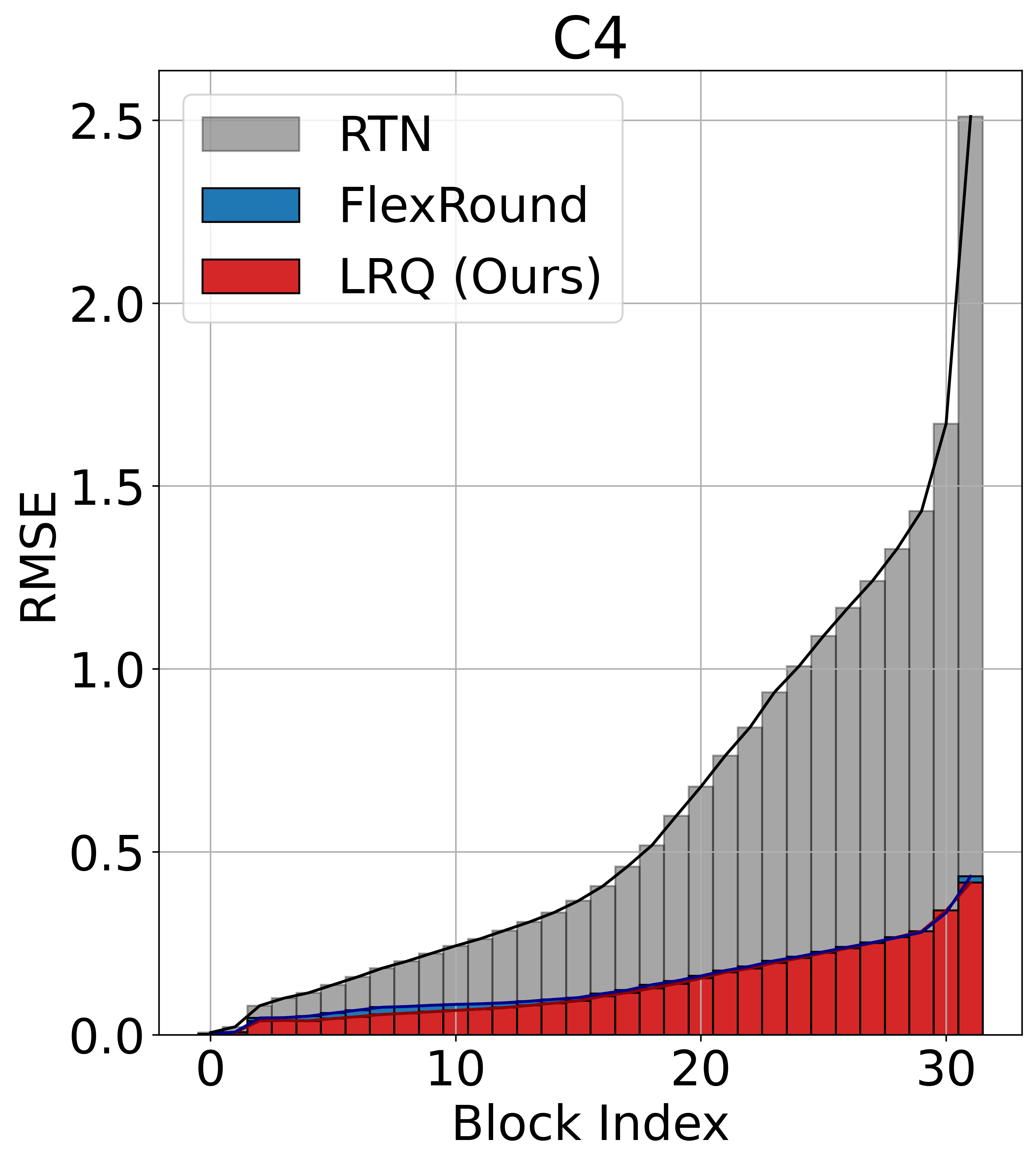}

    }
    \subfigure[Unseen sample]
    {
    \includegraphics[width=0.48\linewidth]{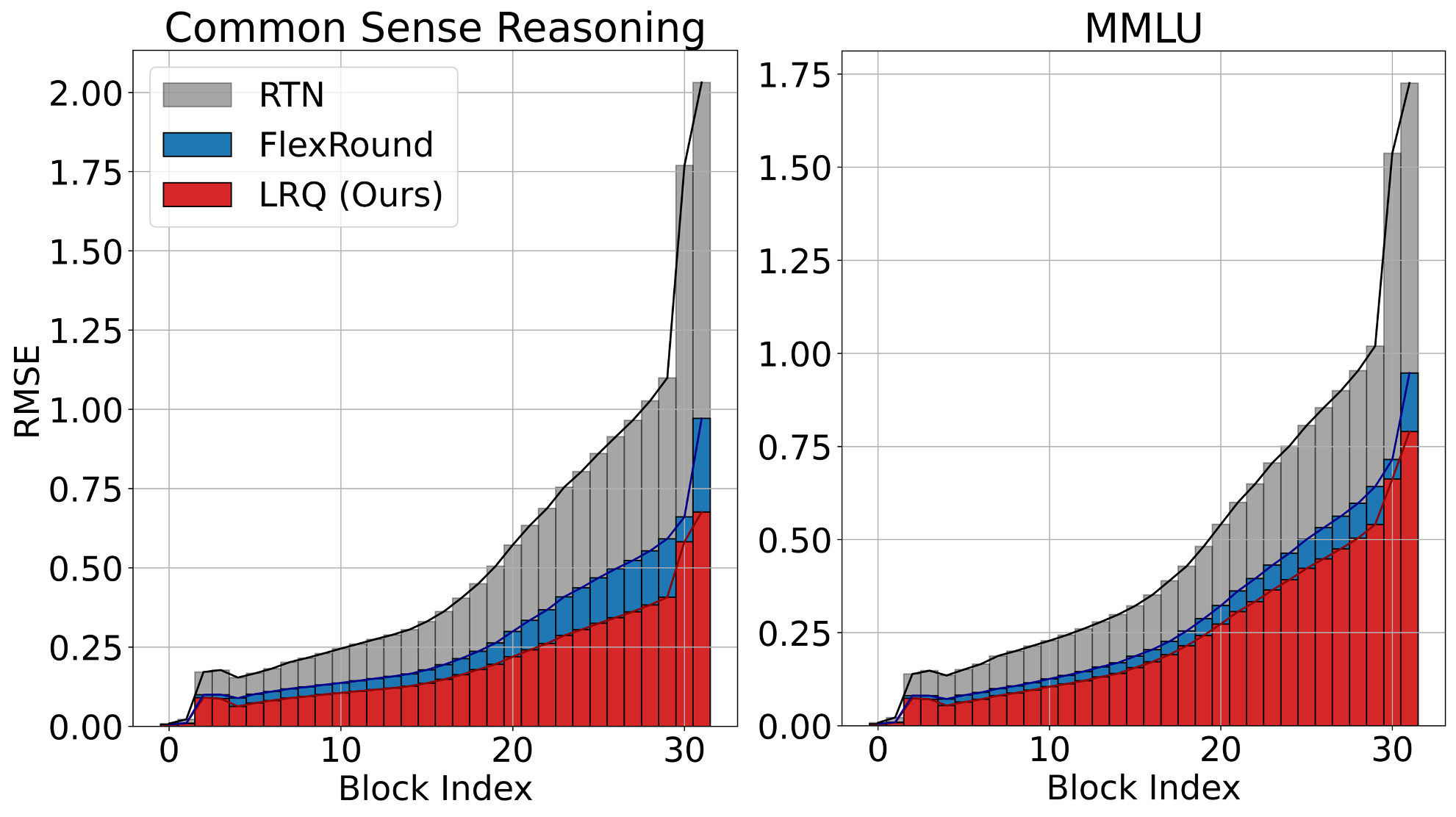}

    }
    \vskip -0.1in
    \centering
    \subfigure[Calibration sample] 
    {
    \includegraphics[width=0.235\linewidth]{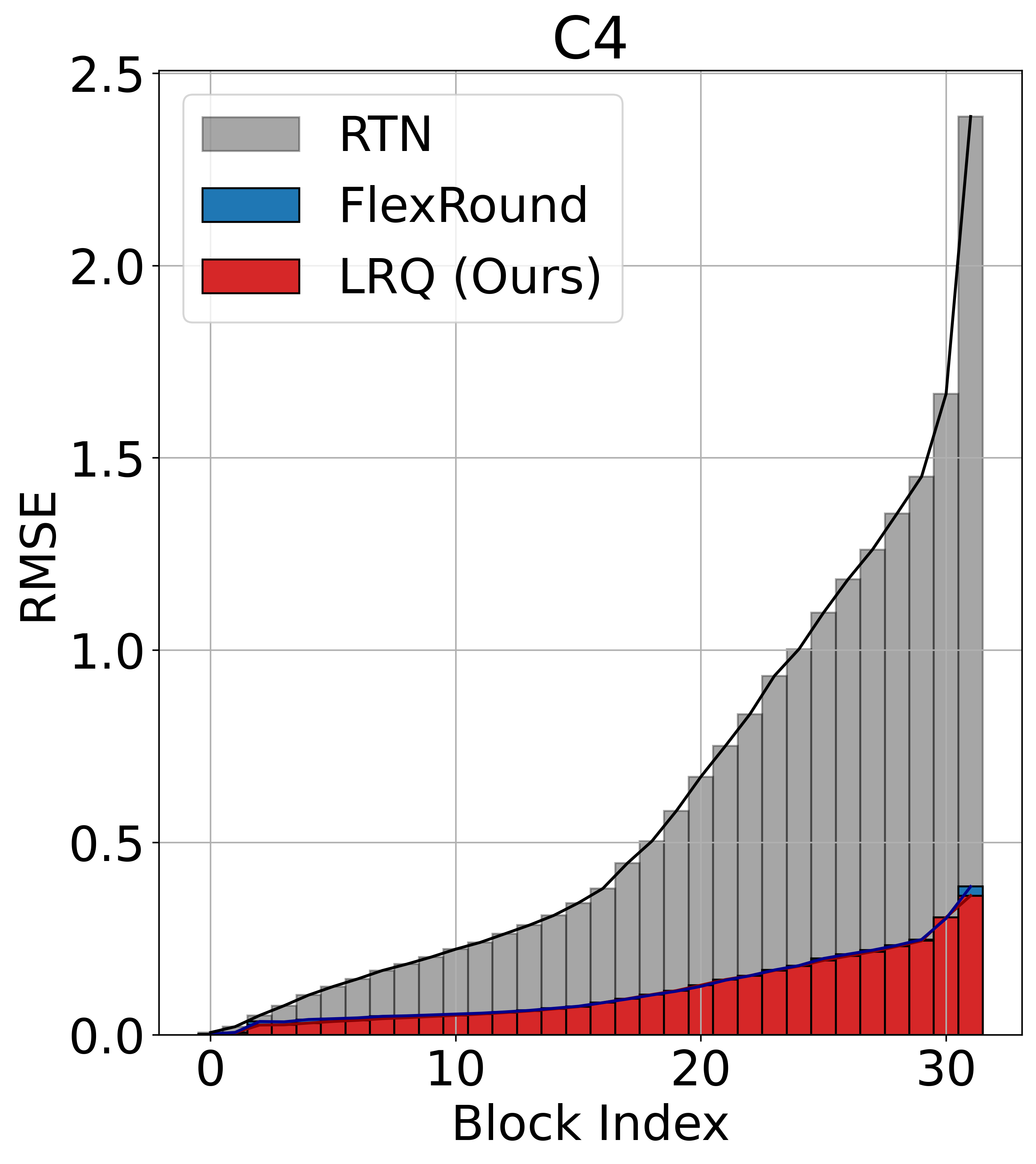}
    
    }
    \subfigure[Unseen sample]
    {
    \includegraphics[width=0.47\linewidth]{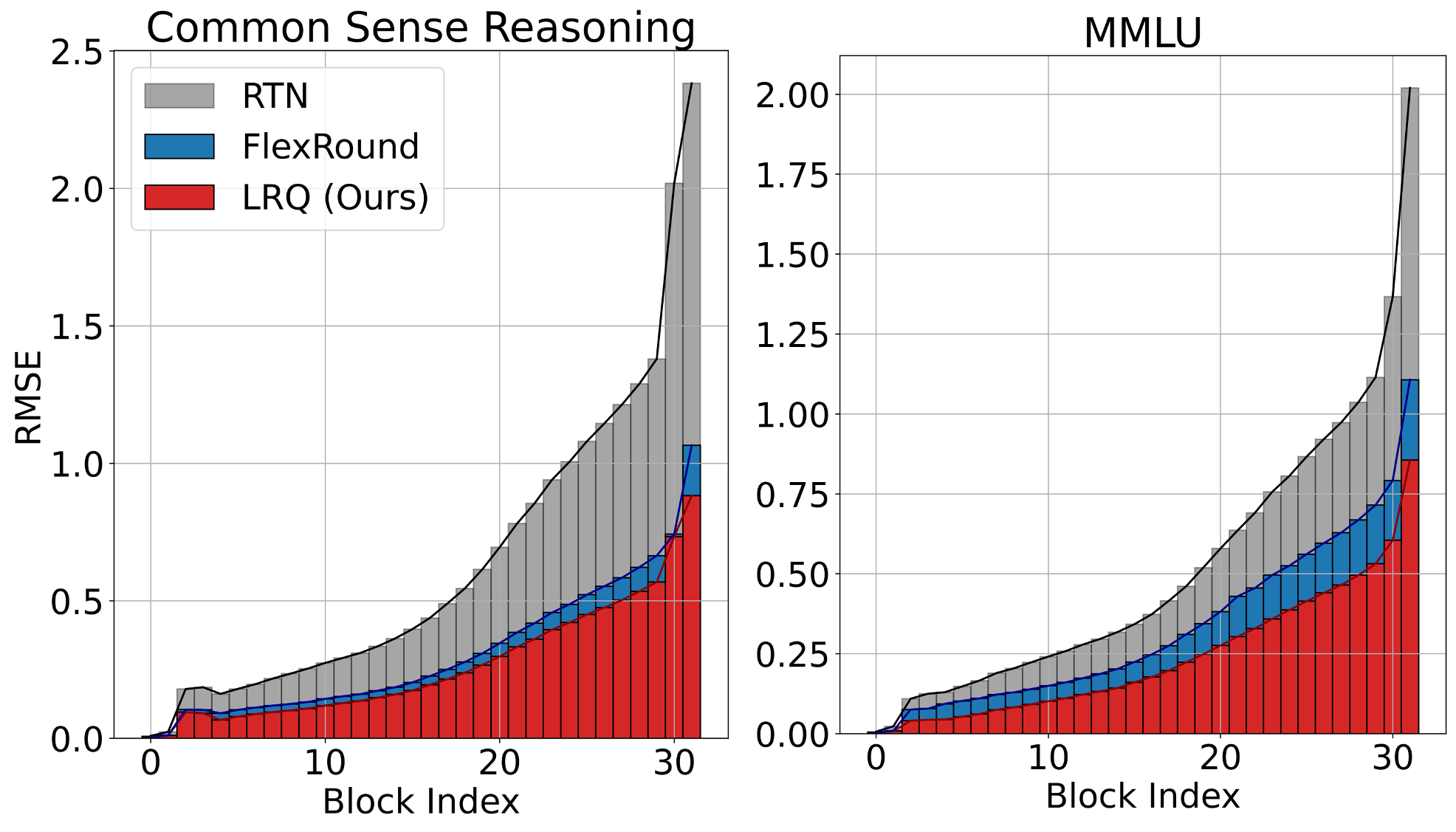}

    }
    \vskip -0.1in
    \caption{Accumulated root mean square error (RMSE) between $\mW\mX$ and $\widehat{\mW}\widetilde{\mX}$ for RTN, FlexRound, and LRQ on (a), (c), (e) three different calibration samples from the C4 dataset and (b), (d), (f) three different unseen samples from common sense reasoning and MMLU benchmarks, ranging from the first Transformer block to the last Transformer block of Llama $7$B. Here, weights and activations are quantized to $8$-bit with per-channel asymmetric quantization and per-tensor asymmetric static quantization respectively, while the KV cache remains in FP16. Note that RMSE tends to rise in line with the block index due to the presence of $\widetilde{\mX}$ that accumulates quantization error resulting from previous quantized Transformer blocks.}
    \label{fig:recon_others}
\end{figure}

\clearpage
\section{Sensitivity of Accumulated RMSE to the Number of Calibration Samples}\label{appendix:sensitivity}

\begin{figure*}[h]
    \centering
    \subfigure[Calibration sample] 
    {
    \includegraphics[width=0.310\linewidth]{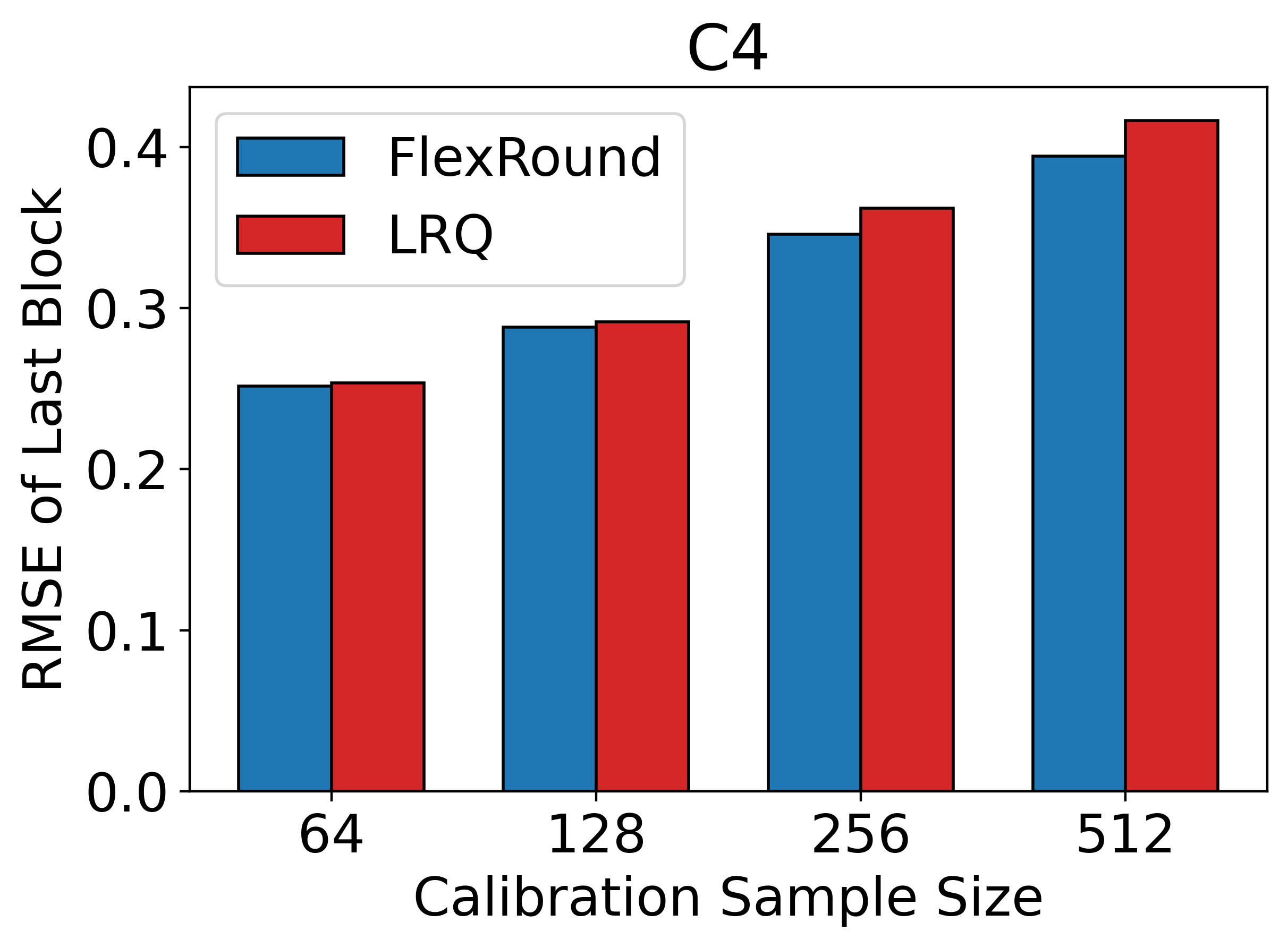}
    
    }
    \subfigure[Unseen sample]
    {
    \includegraphics[width=0.615\linewidth]{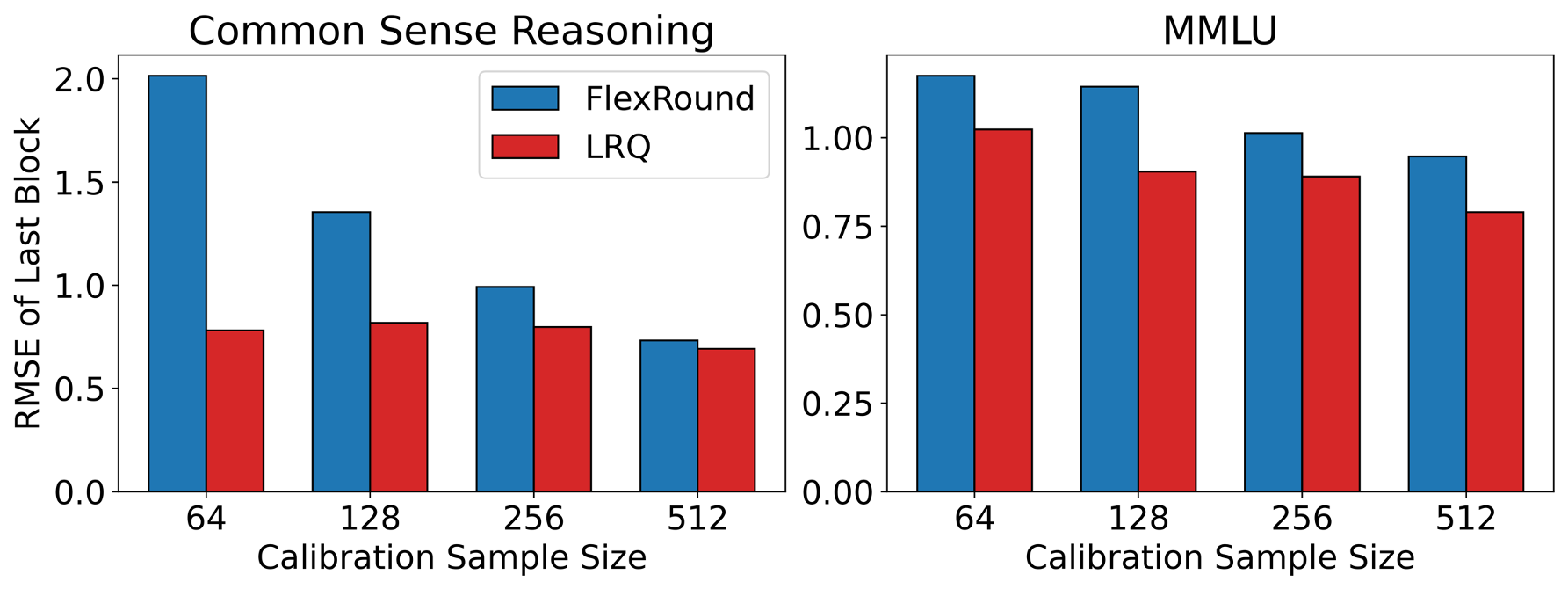}

    }
    \caption{Accumulated root mean square error (RMSE) between $\mW\mX$ and $\widehat{\mW}\widetilde{\mX}$ for FlexRound and LRQ on (a) a calibration sample from the C4 dataset and (b) an unseen sample from common sense reasoning and MMLU benchmarks at the last Transformer block of Llama $7$B. Here, weights and activations are quantized to $8$-bit with per-channel asymmetric quantization and per-tensor asymmetric static quantization, respectively.}
    \label{fig:sensitivity}
\end{figure*}

To figure out the sensitivity of accumulated root mean square error (RMSE) to the number of calibration samples used for the block-wise reconstruction, we compare accumulated RMSE between $\mW\mX$ and $\widehat{\mW}\widetilde{\mX}$ for FlexRound and LRQ at the last Transformer block of Llama $7$B with the number of calibration samples varying from $64$ to $512$. As depicted in Figure 8(a), the accumulated RMSE of the last Transformer block on a calibration sample diminishes with a reduction in the number of calibration samples. This phenomenon is because FlexRound and LRQ are more likely to be fitted to calibration samples as the number of calibration samples becomes smaller. Conversely, Figure 8(b) reveals that the accumulated RMSE of the last Transformer block on each unseen sample from common sense reasoning and MMLU decreases with a larger number of calibration samples.

Notably, the pattern elucidated in Section \ref{subsec:analysis} persists consistently across varying calibration sample sizes from $64$ to $512$. In other words, for every calibration sample size spanning from $64$ to $512$, LRQ consistently attains nearly identical accumulated RMSE to FlexRound for a calibration sample from the C4 dataset. Concurrently, the accumulated RMSE of LRQ remains markedly smaller than that of FlexRound for an unseen sample from common sense reasoning and MMLU. This observation provides additional support for the insight presented in Figure \ref{fig:recon}, as discussed in Section \ref{subsec:analysis}.

\clearpage
\section{Detailed Experimental Results for Llama 3 8B and Mistral 7B}

\begin{table*}[h]
\caption{Zero-shot performance of Llama $3$ $8$B on common sense reasoning tasks (PIQA, HellaSwag, WinoGrande, ARC easy and challenge) and the causal language modeling task on WikiText2 with $4$-bit per-channel asymmetric weight-only quantization. The accuracy ($\%$) and the perplexity (PPL) are reported for common sense reasoning tasks and the causal language modeling task, respectively. The lower PPL, the better. The experimental results of GPTQ, AWQ, and QuIP originate from \citet{huang2024empiricalstudyllama3quantization}.} 
\label{tab:llama3_8b_appendix}
\begin{center}
\small
\begin{tabular}{lcccccccc}
\toprule
Method & \makecell{\# Bits (W/A/KV)} & PIQA & HellaSwag & WinoGrande & ARC-e & ARC-c & Average & WikiText2 \\
\midrule
Llama $3$ $8$B & $16/16/16$ & $79.9$ & $60.2$ & $72.8$ & $80.1$ & $50.4$ & $68.6$ & $6.1$ \\
\midrule
GPTQ & $4/16/16$ & $76.8$ & $57.4$ & $72.8$ & $74.3$ & $42.4$ & $64.8$ & $7.0$ \\
AWQ & $4/16/16$ & $78.3$ & $58.6$ & $72.5$ & $77.6$ & $48.3$ & $67.0$ & $7.1$ \\
QuIP & $4/16/16$ & $78.2$ & $58.6$ & $73.2$ & $78.2$ & $47.4$ & $67.1$ & $\mathbf{6.5}$ \\
FlexRound & $4/16/16$ & $79.3$ & $59.2$ & $73.4$ & $79.3$ & $47.7$ & $67.8$ & $6.9$ \\
LRQ (Ours) & $4/16/16$ & $79.2$ & $59.2$ & $74.4$ & $79.3$ & $47.8$ & $\mathbf{68.0}$ & $6.9$ \\
\bottomrule
\end{tabular}
\end{center}
\end{table*}

\begin{table*}[h]
\caption{Zero-shot performance of Mistral $7$B on common sense reasoning tasks (BoolQ, PIQA, HellaSwag, WinoGrande, ARC easy and challenge, and OpenBookQA) and the causal language modeling task on WikiText2 with per-channel asymmetric weight-only quantization. The accuracy ($\%$) and the perplexity (PPL) are reported for common sense reasoning tasks and the causal language modeling task, respectively. The lower PPL, the better.} 
\label{tab:mistral_7b_appendix}
\begin{center}
\small
\resizebox{\linewidth}{!}{
\begin{tabular}{lcccccccccc}
\toprule
Method & \makecell{\# Bits (W/A/KV)} & BoolQ & PIQA & HellaSwag &  WinoGrande & ARC-e & ARC-c & OBQA & Average & WikiText2\\
\midrule
Mistral $7$B & $16/16/16$ & $83.82$ & $82.15$ & $81.07$ & $73.95$ & $79.50$ & $54.01$ & $43.80$ & $71.19$ & $5.25$\\
\midrule
FlexRound & $3/16/16$ & $79.72$ & $80.58$ & $77.27$ & $66.61$ & $75.59$ & $46.84$ & $39.40$ & $66.57$ & $6.60$ \\
LRQ (Ours) & $3/16/16$ & $80.95$ & $80.74$ & $77.50$ & $68.27$ & $75.76$ & $48.81$ & $40.20$ & $\mathbf{67.46}$ & $\mathbf{6.13}$  \\
\hdashline
FlexRound & $4/16/16$ & $82.91$ & $81.34$ & $79.78$ & $72.45$ & $78.66$ & $51.11$ & $44.40$ & $70.09$ & $5.48$ \\
LRQ (Ours) & $4/16/16$ & $84.01$ & $81.66$ & $79.79$ & $73.09$ & $78.58$ & $51.71$ & $45.00$ & $\mathbf{70.55}$ & $\mathbf{5.45}$ \\
\bottomrule
\end{tabular}
}
\end{center}
\end{table*}

To further justify the effectiveness of LRQ for other model families than Llama, we conduct additional experiments for Mistral 7B v0.1 \citep{jiang2023mistral7b} on common sense reasoning tasks and WikiText2 in a low-bit per-channel weight-only quantization scheme. In Table \ref{tab:mistral_7b_appendix}, LRQ outperforms FlexRound in both $3$-bit and $4$-bit per-channel asymmetric weight-only quantization schemes. In particular, in a $3$-bit per-channel asymmetric weight-only quantization scheme, LRQ surpasses FlexRound by about $0.5$ PPL on WikiText2 and by almost $0.9$ percent on common sense reasoning tasks.

\clearpage
\section{Comparison of Computation Cost to Complete the Quantization Process}

For a comparative analysis of SmoothQuant, FlexRound, and LRQ in terms of computational cost to complete the quantization process, as delineated in Table \ref{tab:computation_cost}, we measure the execution time and peak GPU memory usage while quantizing Llama $7$B with $8$-bit per-channel asymmetric weight quantization and $8$-bit per-tensor asymmetric static activation quantization using $512$ calibration samples and a batch size of $2$. Since both FlexRound and LRQ involve gradient-based optimization in a block-wise manner while SmoothQuant is a learning-free quantization method, FlexRound and LRQ naturally spend more time and GPU memory quantizing LLMs than SmoothQuant. In Table \ref{tab:computation_cost}, LRQ's extended processing time compared to FlexRound is attributed to the multiplication involving $\mL_2$ and $\mU_2$ in Eq. \ref{eq:LRQ}. Despite the slightly longer runtime, LRQ demonstrates an advantage in peak GPU memory usage, utilizing $23.5$ GB compared to FlexRound's $25.4$ GB. This efficiency is attributed to LRQ's fewer learnable parameters in comparison to FlexRound. 

\begin{table}[h]
\caption{Execution time and peak GPU memory usage while quantizing Llama $7$B with $8$-bit per-channel asymmetric weight quantization and $8$-bit per-tensor asymmetric static activation quantization using $512$ calibration samples and a batch size of $2$} 
\label{tab:computation_cost}
\begin{center}
\small
\begin{tabular}{lcc}
\toprule
Method & Execution time (A) & Peak GPU memory usage \\
\midrule
SmoothQuant & $10$ minutes & $13.5$ GB \\
FlexRound & $5$ hours $7$ minutes & $25.4$ GB \\
LRQ (Ours) & $5$ hours $22$ minutes & $23.5$ GB \\
\bottomrule
\end{tabular}
\end{center}
\end{table}

Additionally, we also measure the execution time and peak GPU memory usage while quantizing Llama $2$ $7$B with $4$-bit per-channel asymmetric weight-only quantization using $512$ calibration samples and a batch size of $2$. Similar to Table \ref{tab:computation_cost}, Table \ref{tab:computation_cost_weight_only} shows that LRQ consumes more processing time but less peak GPU memory usage than FlexRound.

\begin{table}[h]
\caption{Execution time and peak GPU memory usage while quantizing Llama $2$ $7$B with $4$-bit per-channel asymmetric weight-only quantization using $512$ calibration samples and a batch size of $2$} 
\label{tab:computation_cost_weight_only}
\begin{center}
\small
\begin{tabular}{lcc}
\toprule
Method & Execution time (A) & Peak GPU memory usage \\
\midrule
FlexRound & $2$ hours $50$ minutes & $23.6$ GB \\
LRQ (Ours) & $3$ hours $3$ minutes & $21.3$ GB \\
\bottomrule
\end{tabular}
\end{center}
\end{table}

\clearpage
\section{Compression and Acceleration Effects after LRQ at Inference Time}\label{appendix:test}

Since only a quantization step size ($\vs_1$) and an integer matrix ($\widetilde{\mW}$) are required during inference, once we obtain an integer matrix by setting $\widetilde{\mW}$ to $\Big\lfloor \frac{\mW}{{\vs_1 \odot \text{exp}(\mL_2 \mU_2 + \vr_2 + \vc_2)}} \Big\rceil$ after $\mL_2$, $\mU_2$, $\vr_2$, and $\vc_2$ are learned, there is no need to recompute the multiplication involving $\mL_2$ and $\mU_2$ at test time. In other words, for inference, like other uniform quantization methods including GPTQ, SmoothQuant, and AWQ, LRQ also requires only a quantization step size $\vs_1$ and an integer matrix $\widetilde{\mW}$ without the presence of $\mL_2$, $\mU_2$, $\vr_2$, and $ \vc_2$. Therefore, packing/unpacking techniques and acceleration kernels \citep{frantar2023optq, lin2023awq, park2024lutgemmquantizedmatrixmultiplication} can be applied to LRQ without additional effort.

Figure \ref{fig:latency} and Table \ref{tab:latency} exhibit compression and acceleration effects after LRQ with $3$-bit and $4$-bit per-channel weight-only uniform quantization. Regarding the compression effect, $3$-bit uniform quantization shows x$4.55$ compression ratio, and $4$-bit uniform quantization shows x$3.58$ compression ratio on the model size for Llama $2$ $7$B. To verify the acceleration effect, we measure the matrix multiplication latency of Llama $2$ FFN layers across $7$B to $70$B on a single token generation. We utilize cuBLAS for FP$16$ baselines, while employing the LUT-GEMM kernel \cite{park2024lutgemmquantizedmatrixmultiplication} for per-channel weight-only uniformly quantized models. As shown in Table \ref{tab:latency}, for Llama 2 $70$B, the $4$-bit per-channel weight-only quantized model shows x$2.33$ faster latency, and the $3$-bit per-channel weight-only quantized model shows x$2.77$ faster latency than the FP$16$ cuBLAS baseline.

\begin{table}[h]
\caption{Average zero-shot performance on common sense reasoning (CSR) tasks, model size, and latency when quantizing Llama 2 $7$B, $13$B, and $70$B with $3$-bit and $4$-bit per-channel weight-only uniform quantization. Matrix multiplication latency of each model's FFN layers is measured using the LUT-GEMM kernel \cite{park2024lutgemmquantizedmatrixmultiplication}.} 
\label{tab:latency}
\begin{center}
\small
\begin{tabular}{llccc}
\toprule
Model & Method & CSR Avg. & Model Size [GB] & Latency [ms] \\
\midrule
& FP16 & $60.45$ &$13.48$ & $0.05987$ \\
Llama 2 7B & LRQ ($3$-bit) &$59.07$ &$2.95$ & $0.03750$  \\
& LRQ ($4$-bit) &$60.38$ &$3.76$ & $0.04462$  \\
\midrule
& FP16 & $62.65$& $26.03$ & $0.08843$ \\
Llama 2 13B & LRQ ($3$-bit) & $61.41$ &$5.41$ & $0.04789$ \\
& LRQ ($4$-bit) & $62.84$ &$7.00$ & $0.05621$ \\
\midrule
& FP16 & $66.77$ & $137.95$ & $0.29088$ \\
Llama 2 70B & LRQ ($3$-bit) & $66.21$ & $26.72$ &$0.10482$ \\
& LRQ ($4$-bit) & $66.95$ & $35.28$ &$0.12479$ \\
\bottomrule
\end{tabular}
\end{center}
\end{table}

\clearpage
\section{Comparison of Experimental Results before and after Per-token Asymmetric KV Cache Quantization}\label{appendix:exp}

Table \ref{tab:llama_csr_per_tensor_appendix}, \ref{tab:llama_mmlu_per_tensor_appendix}, \ref{tab:llama2_csr_per_tensor_appendix}, \ref{tab:llama2_mmlu_per_tensor_appendix}, \ref{tab:llama2_csr_per_token_appendix}, and \ref{tab:llama2_mmlu_per_token_appendix} show the comparison of experimental results before and after per-token asymmetric KV cache quantization. It can be easily seen that the performance difference before and after per-token asymmetric KV cache quantization is nearly inconsiderable no matter which quantization technique is chosen, as mentioned in Section \ref{subsec:per-tensor}. Furthermore, even without per-token asymmetric KV cache quantization, LRQ still outperforms prior state-of-the-art LLM post-training weight-activation quantization methods in most cases.

\begin{table}[h]
\caption{Zero-shot performance of Llama on common sense reasoning tasks (BoolQ, PIQA, HellaSwag, WinoGrande, ARC easy and challenge, and OpenBookQA) with per-channel asymmetric weight quantization, per-tensor asymmetric static activation quantization, and per-token asymmetric KV cache quantization (if applied). Please refer to Figure \ref{fig:diagram}. The accuracy ($\%$) is reported for common sense reasoning tasks. The number of bits used for weights, activations, and KV cache is expressed as W/A/KV.} 
\label{tab:llama_csr_per_tensor_appendix}
\begin{center}
\small
\resizebox{\linewidth}{!}{
\begin{tabular}{lccccccccc}
\toprule
Method & \makecell{\# Bits (W/A/KV)} & BoolQ & PIQA & HellaSwag &  WinoGrande & ARC-e & ARC-c & OBQA & Average\\
\midrule
Llama $7$B & $16/16/16$ & $73.15$ & $77.31$ & $72.96$ & $67.09$ & $52.48$ & $41.38$ & $42.40$ & $60.97$ \\
\midrule
RTN & $8/8/16$ & $71.56$ & $73.72$ & $65.86$ & $63.93$ & $49.49$ & $36.43$ & $38.80$ & $57.11$ \\
SmoothQuant & $8/8/16$ & $69.63$ & $73.12$ & $68.88$ & $65.43$ & $48.70$ & $38.57$ & $38.00$ & $57.48$ \\
FlexRound & $8/8/16$ & $73.76$ & $76.66$ & $71.75$ & $67.01$ & $52.31$ & $40.02$ & $42.20$ & $60.53$ \\
LRQ (Ours) & $8/8/16$ & $73.03$ & $77.64$ & $72.10$ & $66.77$ & $52.95$ & $40.87$ & $41.60$ & $\mathbf{60.71}$ \\
\hdashline
RTN & $8/8/8$ & $69.76$ & $73.72$ & $65.95$ & $62.75$ & $48.91$ & $37.12$ & $37.60$ & $56.54$ \\
SmoothQuant & $8/8/8$ & $69.42$ & $72.63$ & $69.07$ & $64.72$ & $48.61$ & $37.12$ & $39.20$ & $57.25$ \\
FlexRound & $8/8/8$ & $72.54$ & $76.50$ & $71.88$ & $66.77$ & $53.03$ & $39.76$ & $42.00$ & $60.35$ \\
LRQ (Ours) & $8/8/8$ & $72.84$ & $77.37$ & $72.04$ & $67.01$ & $53.03$ & $40.53$ & $41.60$ & $\mathbf{60.63}$ \\
\midrule
Llama $13$B & $16/16/16$ & $68.53$ & $79.11$ & $76.23$ & $70.01$ & $59.89$ & $44.54$ & $42.20$ & $62.93$ \\
\midrule
RTN & $8/8/16$ & $66.06$ & $71.82$ & $65.70$ & $62.98$ & $50.97$ & $35.58$ & $36.60$ & $55.67$ \\
SmoothQuant & $8/8/16$ & $68.29$ & $75.30$ & $71.82$ & $68.03$ & $55.18$ & $40.19$ & $41.20$ & $60.00$ \\
FlexRound & $8/8/16$ & $68.59$ & $78.67$ & $75.21$ & $70.64$ & $58.88$ & $43.60$ & $41.20$ & $62.40$ \\
LRQ (Ours) & $8/8/16$ & $68.99$ & $79.22$ & $75.61$ & $71.19$ & $58.92$ & $43.52$ & $43.00$ & $\mathbf{62.92}$ \\
\hdashline
RTN & $8/8/8$ & $65.87$ & $72.25$ & $62.52$ & $62.19$ & $51.81$ & $35.41$ & $38.40$ & $55.49$ \\
SmoothQuant & $8/8/8$ & $67.34$ & $75.19$ & $71.78$ & $69.06$ & $54.92$ & $40.44$ & $38.80$ & $59.65$ \\
FlexRound & $8/8/8$ & $68.78$ & $78.51$ & $75.23$ & $70.56$ & $58.46$ & $44.03$ & $41.00$ & $62.37$ \\
LRQ (Ours) & $8/8/8$ & $68.84$ & $78.78$ & $75.56$ & $70.80$ & $59.13$ & $44.62$ & $41.60$ & $\mathbf{62.76}$ \\
\midrule
Llama $33$B & $16/16/16$ & $68.38$ & $80.09$ & $79.21$ & $72.93$ & $58.92$ & $45.48$ & $42.00$ & $63.86$ \\
\midrule
RTN & $8/8/16$ & $69.02$ & $76.01$ & $69.11$ & $66.54$ & $57.07$ & $41.64$ & $41.40$ & $60.11$ \\
SmoothQuant & $8/8/16$ & $71.04$ & $75.24$ & $71.01$ & $69.38$ & $54.38$ & $43.34$ & $40.60$ & $60.71$ \\
FlexRound & $8/8/16$ & $69.08$ & $79.16$ & $77.43$ & $72.53$ & $56.61$ & $44.97$ & $44.00$ & $63.40$ \\
LRQ (Ours) & $8/8/16$ & $68.44$ & $80.03$ & $78.37$ & $74.19$ & $58.16$ & $46.33$ & $42.20$ & $\mathbf{63.96}$ \\
\hdashline
RTN & $8/8/8$ & $68.81$ & $76.55$ & $68.76$ & $66.06$ & $56.48$ & $42.49$ & $42.40$ & $60.22$ \\
SmoothQuant & $8/8/8$ & $71.31$ & $75.30$ & $71.29$ & $68.98$ & $53.66$ & $43.26$ & $41.00$ & $60.69$ \\
FlexRound & $8/8/8$ & $69.05$ & $79.49$ & $77.49$ & $70.88$ & $56.86$ & $43.60$ & $42.00$ & $62.77$ \\
LRQ (Ours) & $8/8/8$ & $68.84$ & $79.98$ & $78.52$ & $73.72$ & $58.21$ & $45.73$ & $43.00$ & $\mathbf{64.00}$ \\
\midrule
Llama $65$B & $16/16/16$ & $82.32$ & $80.85$ & $80.71$ & $77.19$ & $58.71$ & $46.33$ & $44.60$ & $67.24$ \\
\midrule
RTN & $8/8/16$ & $79.48$ & $77.04$ & $74.15$ & $71.19$ & $52.48$ & $43.52$ & $43.80$ & $63.09$ \\
SmoothQuant & $8/8/16$ & $78.72$ & $78.84$ & $79.12$ & $74.03$ & $56.23$ & $45.22$ & $43.20$ & $65.05$ \\
FlexRound & $8/8/16$ & $81.31$ & $79.33$ & $79.16$ & $73.56$ & $57.83$ & $46.08$ & $44.60$ & $65.98$ \\
LRQ (Ours) & $8/8/16$ & $82.45$ & $80.69$ & $79.92$ & $76.64$ & $58.92$ & $46.67$ & $45.60$ & $\mathbf{67.27}$ \\
\hdashline
RTN & $8/8/8$ & $79.51$ & $75.79$ & $74.13$ & $71.35$ & $51.85$ & $44.03$ & $43.60$ & $62.89$ \\
SmoothQuant & $8/8/8$ & $78.78$ & $79.54$ & $79.11$ & $73.32$ & $56.23$ & $45.90$ & $43.80$ & $65.24$ \\
FlexRound & $8/8/8$ & $80.46$ & $79.38$ & $79.23$ & $74.98$ & $57.20$ & $46.42$ & $45.00$ & $66.10$ \\
LRQ (Ours) & $8/8/8$ & $82.35$ & $81.12$ & $79.96$ & $75.61$ & $58.96$ & $46.59$ & $45.40$ & $\mathbf{67.14}$ \\
\bottomrule
\end{tabular}
}
\end{center}
\end{table}

\begin{table}[h]
\caption{Five-shot performance of Llama on Massive Multitask Language Understanding with per-channel asymmetric weight quantization, per-tensor asymmetric static activation quantization, and per-token asymmetric KV cache quantization (if applied). Please refer to Figure \ref{fig:diagram}. The accuracy ($\%$) is reported for four groups of disciplines (STEM, Humanities, Social Science, and Other). The number of bits used for weights, activations, and KV cache is expressed as W/A/KV.} 
\label{tab:llama_mmlu_per_tensor_appendix}
\begin{center}
\small
\begin{tabular}{lcccccc}
\toprule
Method & \makecell{\# Bits (W/A/KV)} & STEM & Humanities & Social Science & Other & Average\\
\midrule
Llama $7$B & $16/16/16$ & $30.58$ & $33.88$ & $38.19$ & $38.25$ & $35.12$ \\
\midrule
RTN & $8/8/16$ & $27.40$ & $27.16$ & $29.18$ & $30.38$ & $28.40$ \\
SmoothQuant & $8/8/16$ & $28.36$ & $27.89$ & $32.63$ & $30.41$ & $29.61$ \\
FlexRound & $8/8/16$ & $28.30$ & $29.20$ & $30.13$ & $33.47$ & $30.20$ \\
LRQ (Ours) & $8/8/16$ & $29.69$ & $32.48$ & $37.63$ & $38.80$ & $\mathbf{34.47}$ \\
\hdashline
RTN & $8/8/8$ & $27.04$ & $27.23$ & $29.28$ & $30.38$ & $28.36$ \\
SmoothQuant & $8/8/8$ & $28.40$ & $28.69$ & $32.79$ & $30.48$ & $29.94$ \\
FlexRound & $8/8/8$ & $27.60$ & $28.71$ & $29.61$ & $31.99$ & $29.43$ \\
LRQ (Ours) & $8/8/8$ & $29.72$ & $32.79$ & $37.44$ & $38.16$ & $\mathbf{34.39}$\\
\midrule
Llama $13$B & $16/16/16$ & $36.35$ & $44.97$ & $54.14$ & $53.15$ & $47.02$ \\
\midrule
RTN & $8/8/16$ & $26.61$ & $25.53$ & $27.40$ & $24.52$ & $25.94$ \\
SmoothQuant & $8/8/16$ & $27.80$ & $29.31$ & $31.04$ & $30.88$ & $29.73$ \\
FlexRound & $8/8/16$ & $35.06$ & $41.68$ & $49.37$ & $49.81$ & $43.82$ \\
LRQ (Ours) & $8/8/16$ & $34.72$ & $44.65$ & $51.71$ & $52.28$ & $\mathbf{45.83}$ \\
\hdashline
RTN & $8/8/8$ & $26.38$ & $25.33$ & $27.95$ & $24.83$ & $26.01$ \\
SmoothQuant & $8/8/8$ & $27.24$ & $30.12$ & $30.58$ & $31.31$ & $29.87$ \\
FlexRound & $8/8/8$ & $33.63$ & $42.81$ & $48.65$ & $49.26$ & $43.60$ \\
LRQ (Ours) & $8/8/8$ & $35.16$ & $44.55$ & $51.74$ & $52.04$ & $\mathbf{45.83}$\\
\midrule
Llama $33$B & $16/16/16$ & $46.69$ & $56.39$ & $67.40$ & $63.60$ & $58.38$ \\
\midrule
RTN & $8/8/16$ & $32.14$ & $32.22$ & $37.11$ & $38.25$ & $34.67$ \\
SmoothQuant & $8/8/16$ & $38.17$ & $41.45$ & $50.37$ & $51.08$ & $44.92$ \\
FlexRound & $8/8/16$ & $43.94$ & $52.31$ & $62.14$ & $60.21$ & $54.49$ \\
LRQ (Ours) & $8/8/16$ & $45.13$ & $52.99$ & $64.12$ & $61.88$ & $\mathbf{55.79}$ \\
\hdashline
RTN & $8/8/8$ & $32.47$ & $32.37$ & $38.35$ & $40.59$ & $35.60$ \\
SmoothQuant & $8/8/8$ & $37.94$ & $41.64$ & $50.57$ & $51.48$ & $45.07$ \\
FlexRound & $8/8/8$ & $43.47$ & $52.20$ & $61.94$ & $59.90$ & $54.24$ \\
LRQ (Ours) & $8/8/8$ & $45.26$ & $52.58$ & $63.99$ & $61.26$ & $\mathbf{55.51}$\\
\midrule
Llama $65$B & $16/16/16$ & $51.95$ & $61.87$ & $73.32$ & $67.58$ & $63.57$ \\
\midrule
RTN & $8/8/16$ & $42.25$ & $46.74$ & $61.13$ & $54.57$ & $50.73$ \\
SmoothQuant & $8/8/16$ & $44.70$ & $50.54$ & $63.99$ & $57.28$ & $53.79$ \\
FlexRound & $8/8/16$ & $46.52$ & $54.30$ & $66.36$ & $60.83$ & $56.78$ \\
LRQ (Ours) & $8/8/16$ & $50.89$ & $61.15$ & $72.64$ & $66.04$ & $\mathbf{62.59}$ \\
\hdashline
RTN & $8/8/8$ & $41.22$ & $47.23$ & $61.39$ & $54.69$ & $50.76$ \\
SmoothQuant & $8/8/8$ & $44.83$ & $50.82$ & $63.34$ & $57.09$ & $53.72$ \\
FlexRound & $8/8/8$ & $46.32$ & $54.60$ & $65.06$ & $62.49$ & $56.94$ \\
LRQ (Ours) & $8/8/8$ & $50.96$ & $61.28$ & $71.99$ & $66.66$ & $\mathbf{62.65}$ \\
\bottomrule
\end{tabular}
\end{center}
\end{table}

\begin{table}[h]
\caption{Zero-shot performance of Llama $2$ on common sense reasoning tasks (BoolQ, PIQA, HellaSwag, WinoGrande, ARC easy and challenge, and OpenBookQA) with per-channel asymmetric weight quantization, per-tensor asymmetric static activation quantization, and per-token asymmetric KV cache quantization (if applied). Please refer to Figure \ref{fig:diagram}. The accuracy ($\%$) is reported for common sense reasoning tasks. The number of bits used for weights, activations, and KV cache is expressed as W/A/KV.}
\label{tab:llama2_csr_per_tensor_appendix}
\begin{center}
\small
\resizebox{\linewidth}{!}{
\begin{tabular}{lccccccccc}
\toprule
Method & \makecell{\# Bits (W/A/KV)} & BoolQ & PIQA & HellaSwag &  WinoGrande & ARC-e & ARC-c & OBQA & Average\\
\midrule
Llama $2$ $7$B & $16/16/16$ & $71.07$ & $76.99$ & $72.96$ & $67.25$ & $53.58$ & $40.53$ & $40.80$ & $60.45$ \\
\midrule
RTN & $8/8/16$ & $60.86$ & $67.19$ & $57.53$ & $59.43$ & $45.50$ & $32.00$ & $34.20$ & $50.96$ \\
SmoothQuant & $8/8/16$ & $67.09$ & $72.03$ & $67.34$ & $65.43$ & $50.88$ & $37.12$ & $38.20$ & $56.87$ \\
FlexRound & $8/8/16$ & $71.99$ & $77.04$ & $71.23$ & $65.11$ & $54.42$ & $40.44$ & $38.80$ & $59.86$ \\
LRQ (Ours) & $8/8/16$ & $67.49$ & $77.58$ & $72.19$ & $67.96$ & $54.76$ & $39.59$ & $40.40$ & $\mathbf{60.00}$ \\
\hdashline
RTN & $8/8/8$ & $60.58$ & $67.08$ & $57.66$ & $60.54$ & $45.83$ & $31.57$ & $34.40$ & $51.09$ \\
SmoothQuant & $8/8/8$ & $67.65$ & $73.29$ & $67.52$ & $62.90$ & $51.35$ & $37.80$ & $37.60$ & $56.87$ \\
FlexRound & $8/8/8$ & $72.05$ & $77.26$ & $71.30$ & $65.98$ & $54.88$ & $39.16$ & $39.20$ & $\mathbf{59.98}$ \\
LRQ (Ours) & $8/8/8$ & $67.86$ & $76.99$ & $71.97$ & $67.01$ & $54.71$ & $40.19$ & $40.00$ & $59.82$ \\
\midrule
Llama $2$ $13$B & $16/16/16$ & $69.02$ & $79.05$ & $76.62$ & $69.61$ & $57.95$ & $44.28$ & $42.00$ & $62.65$ \\
\midrule
RTN & $8/8/16$ & $63.12$ & $73.99$ & $62.60$ & $58.80$ & $52.15$ & $36.26$ & $36.40$ & $54.76$ \\
SmoothQuant & $8/8/16$ & $64.19$ & $76.28$ & $70.75$ & $66.06$ & $54.42$ & $40.78$ & $39.60$ & $58.87$ \\
FlexRound & $8/8/16$ & $66.70$ & $78.56$ & $75.63$ & $69.06$ & $58.33$ & $43.26$ & $40.00$ & $61.65$ \\
LRQ (Ours) & $8/8/16$ & $68.65$ & $78.45$ & $75.79$ & $71.74$ & $59.34$ & $43.94$ & $41.40$ & $\mathbf{62.76}$ \\
\hdashline
RTN & $8/8/8$ & $62.97$ & $73.72$ & $62.60$ & $57.77$ & $52.86$ & $36.77$ & $37.00$ & $54.81$ \\
SmoothQuant & $8/8/8$ & $63.55$ & $75.95$ & $70.99$ & $66.30$ & $53.96$ & $40.10$ & $40.60$ & $58.78$ \\
FlexRound & $8/8/8$ & $66.94$ & $79.00$ & $75.32$ & $69.38$ & $58.54$ & $42.92$ & $40.40$ & $61.79$ \\
LRQ (Ours) & $8/8/8$ & $68.59$ & $78.67$ & $75.83$ & $70.64$ & $58.16$ & $43.34$ & $39.80$ & $\mathbf{62.15}$ \\
\midrule
Llama $2$ $70$B & $16/16/16$ & $76.70$ & $80.85$ & $80.85$ & $76.95$ & $59.72$ & $47.95$ & $44.40$ & $66.77$ \\
\midrule
RTN & $8/8/16$ & $73.27$ & $78.18$ & $76.89$ & $69.69$ & $57.91$ & $45.90$ & $41.60$ & $63.35$ \\
SmoothQuant & $8/8/16$ & $76.82$ & $76.82$ & $79.35$ & $72.77$ & $56.06$ & $45.39$ & $43.20$ & $64.34$ \\
FlexRound & $8/8/16$ & $75.81$ & $80.25$ & $79.03$ & $74.59$ & $59.43$ & $46.42$ & $43.40$ & $65.56$ \\
LRQ (Ours) & $8/8/16$ & $77.71$ & $80.69$ & $79.83$ & $74.11$ & $57.91$ & $45.99$ & $43.60$ & $\mathbf{65.69}$ \\
\hdashline
RTN & $8/8/8$ & $72.39$ & $78.51$ & $76.49$ & $69.61$ & $57.74$ & $44.62$ & $40.40$ & $62.82$ \\
SmoothQuant & $8/8/8$ & $76.21$ & $76.55$ & $79.30$ & $74.11$ & $55.85$ & $46.25$ & $45.60$ & $64.84$ \\
FlexRound & $8/8/8$ & $76.18$ & $80.36$ & $79.09$ & $75.06$ & $60.10$ & $46.42$ & $43.80$ & $65.86$ \\
LRQ (Ours) & $8/8/8$ & $77.95$ & $81.23$ & $79.78$ & $74.82$ & $57.83$ & $46.33$ & $43.60$ & $\mathbf{65.93}$ \\
\bottomrule
\end{tabular}
}
\end{center}
\end{table}

\begin{table}[h]
\caption{Zero-shot performance of Llama $2$ on common sense reasoning tasks (BoolQ, PIQA, HellaSwag, WinoGrande, ARC easy and challenge, and OpenBookQA) with per-channel asymmetric weight quantization, per-tensor asymmetric static activation quantization, and per-token asymmetric KV cache quantization (if applied). Please refer to Figure \ref{fig:diagram}. The accuracy ($\%$) is reported for common sense reasoning tasks. The number of bits used for weights, activations, and KV cache is expressed as W/A/KV.}
\begin{center}
\small
\resizebox{\linewidth}{!}{
\begin{tabular}{lccccccccc}
\toprule
Method & \makecell{\# Bits (W/A/KV)} & BoolQ & PIQA & HellaSwag &  WinoGrande & ARC-e & ARC-c & OBQA & Average\\
\midrule
Llama $2$ $7$B & $16/16/16$ & $71.07$ & $76.99$ & $72.96$ & $67.25$ & $53.58$ & $40.53$ & $40.80$ & $60.45$ \\
\midrule
RTN & $8/8/8$ & $60.58$ & $67.08$ & $57.66$ & $60.54$ & $45.83$ & $31.57$ & $34.40$ & $51.09$ \\
SmoothQuant & $8/8/8$ & $67.65$ & $73.29$ & $67.52$ & $62.90$ & $51.35$ & $37.80$ & $37.60$ & $56.87$ \\
FlexRound & $8/8/8$ & $72.05$ & $77.26$ & $71.30$ & $65.98$ & $54.88$ & $39.16$ & $39.20$ & $\mathbf{59.98}$ \\
LRQ (Ours) & $8/8/8$ & $67.86$ & $76.99$ & $71.97$ & $67.01$ & $54.71$ & $40.19$ & $40.00$ & $59.82$ \\
\midrule
Llama $2$ $13$B & $16/16/16$ & $69.02$ & $79.05$ & $76.62$ & $69.61$ & $57.95$ & $44.28$ & $42.00$ & $62.65$ \\
\midrule
RTN & $8/8/8$ & $62.97$ & $73.72$ & $62.60$ & $57.77$ & $52.86$ & $36.77$ & $37.00$ & $54.81$ \\
SmoothQuant & $8/8/8$ & $63.55$ & $75.95$ & $70.99$ & $66.30$ & $53.96$ & $40.10$ & $40.60$ & $58.78$ \\
FlexRound & $8/8/8$ & $66.94$ & $79.00$ & $75.32$ & $69.38$ & $58.54$ & $42.92$ & $40.40$ & $61.79$ \\
LRQ (Ours) & $8/8/8$ & $68.59$ & $78.67$ & $75.83$ & $70.64$ & $58.16$ & $43.34$ & $39.80$ & $\mathbf{62.15}$ \\
\midrule
Llama $2$ $70$B & $16/16/16$ & $76.70$ & $80.85$ & $80.85$ & $76.95$ & $59.72$ & $47.95$ & $44.40$ & $66.77$ \\
\midrule
RTN & $8/8/8$ & $72.39$ & $78.51$ & $76.49$ & $69.61$ & $57.74$ & $44.62$ & $40.40$ & $62.82$ \\
SmoothQuant & $8/8/8$ & $76.21$ & $76.55$ & $79.30$ & $74.11$ & $55.85$ & $46.25$ & $45.60$ & $64.84$ \\
FlexRound & $8/8/8$ & $76.18$ & $80.36$ & $79.09$ & $75.06$ & $60.10$ & $46.42$ & $43.80$ & $65.86$ \\
LRQ (Ours) & $8/8/8$ & $77.95$ & $81.23$ & $79.78$ & $74.82$ & $57.83$ & $46.33$ & $43.60$ & $\mathbf{65.93}$ \\
\bottomrule
\end{tabular}
}
\end{center}
\end{table}

\begin{table}[h]
\caption{Five-shot performance of Llama $2$ on Massive Multitask Language Understanding with per-channel asymmetric weight quantization, per-tensor asymmetric static activation quantization, and per-token asymmetric KV cache quantization (if applied). Please refer to Figure \ref{fig:diagram}. The accuracy ($\%$) is reported for four groups of disciplines (STEM, Humanities, Social Science, and Other). The number of bits used for weights, activations, and KV cache is expressed as W/A/KV.}
\label{tab:llama2_mmlu_per_tensor_appendix}
\begin{center}
\small
\begin{tabular}{lcccccc}
\toprule
Method & \makecell{\# Bits (W/A/KV)} & STEM & Humanities & Social Science & Other & Average\\
\midrule
Llama $2$ $7$B & $16/16/16$ & $37.04$ & $43.38$ & $51.84$ & $52.44$ & $45.96$ \\
\midrule
RTN & $8/8/16$ & $28.26$ & $24.65$ & $31.39$ & $24.68$ & $26.91$ \\
SmoothQuant & $8/8/16$ & $28.99$ & $29.14$ & $35.33$ & $34.98$ & $31.81$ \\
FlexRound & $8/8/16$ & $32.70$ & $38.38$ & $43.58$ & $45.77$ & $40.01$ \\
LRQ (Ours) & $8/8/16$ & $34.36$ & $40.02$ & $46.64$ & $47.32$ & $\mathbf{41.94}$ \\
\hdashline
RTN & $8/8/8$ & $29.66$ & $24.06$ & $30.45$ & $24.49$ & $26.76$ \\
SmoothQuant & $8/8/8$ & $30.42$ & $27.95$ & $34.29$ & $34.27$ & $31.33$ \\
FlexRound & $8/8/8$ & $33.40$ & $36.96$ & $43.13$ & $46.30$ & $39.70$ \\
LRQ (Ours) & $8/8/8$ & $34.82$ & $39.91$ & $46.47$ & $47.62$ & $\mathbf{42.04}$\\
\midrule
Llama $2$ $13$B & $16/16/16$ & $44.27$ & $54.43$ & $63.41$ & $60.76$ & $55.68$ \\
\midrule
RTN & $8/8/16$ & $29.16$ & $24.38$ & $30.52$ & $29.49$ & $27.93$ \\
SmoothQuant & $8/8/16$ & $28.76$ & $29.46$ & $34.68$ & $35.44$ & $31.83$ \\
FlexRound & $8/8/16$ & $41.95$ & $51.20$ & $60.90$ & $59.65$ & $53.29$ \\
LRQ (Ours) & $8/8/16$ & $42.78$ & $52.65$ & $61.85$ & $59.25$ & $\mathbf{54.07}$ \\
\hdashline
RTN & $8/8/8$ & $29.06$ & $24.23$ & $29.93$ & $29.03$ & $27.62$ \\
SmoothQuant & $8/8/8$ & $30.98$ & $29.29$ & $35.36$ & $35.29$ & $32.37$ \\
FlexRound & $8/8/8$ & $41.09$ & $51.58$ & $61.39$ & $59.41$ & $53.28$ \\
LRQ (Ours) & $8/8/8$ & $42.88$ & $51.97$ & $62.14$ & $59.93$ & $\mathbf{54.08}$\\
\midrule
Llama $2$ $70$B & $16/16/16$ & $57.79$ & $65.16$ & $80.44$ & $74.61$ & $69.11$ \\
\midrule
RTN & $8/8/16$ & $45.99$ & $52.69$ & $65.52$ & $59.16$ & $55.58$ \\
SmoothQuant & $8/8/16$ & $48.11$ & $54.05$ & $68.12$ & $63.23$ & $57.98$ \\
FlexRound & $8/8/16$ & $53.64$ & $61.36$ & $77.35$ & $71.90$ & $65.64$ \\
LRQ (Ours) & $8/8/16$ & $54.41$ & $62.78$ & $77.48$ & $71.56$ & $\mathbf{66.23}$ \\
\hdashline
RTN & $8/8/8$ & $46.82$ & $53.37$ & $66.23$ & $58.51$ & $55.97$ \\
SmoothQuant & $8/8/8$ & $47.51$ & $53.84$ & $68.35$ & $63.94$ & $57.99$ \\
FlexRound & $8/8/8$ & $54.27$ & $61.11$ & $77.45$ & $71.31$ & $65.57$ \\
LRQ (Ours) & $8/8/8$ & $54.44$ & $62.61$ & $76.99$ & $71.78$ & $\mathbf{66.12}$ \\
\bottomrule
\end{tabular}
\end{center}
\end{table}

\begin{table}[h]
\caption{Five-shot performance of Llama $2$ on Massive Multitask Language Understanding with per-channel asymmetric weight quantization, per-tensor asymmetric static activation quantization, and per-token asymmetric KV cache quantization (if applied). Please refer to Figure \ref{fig:diagram}. The accuracy ($\%$) is reported for four groups of disciplines (STEM, Humanities, Social Science, and Other). The number of bits used for weights, activations, and KV cache is expressed as W/A/KV.}
\begin{center}
\small
\begin{tabular}{lcccccc}
\toprule
Method & \makecell{\# Bits (W/A/KV)} & STEM & Humanities & Social Science & Other & Average\\
\midrule
Llama $2$ $7$B & $16/16/16$ & $37.04$ & $43.38$ & $51.84$ & $52.44$ & $45.96$ \\
\midrule
RTN & $8/8/8$ & $29.66$ & $24.06$ & $30.45$ & $24.49$ & $26.76$ \\
SmoothQuant & $8/8/8$ & $30.42$ & $27.95$ & $34.29$ & $34.27$ & $31.33$ \\
FlexRound & $8/8/8$ & $33.40$ & $36.96$ & $43.13$ & $46.30$ & $39.70$ \\
LRQ (Ours) & $8/8/8$ & $34.82$ & $39.91$ & $46.47$ & $47.62$ & $\mathbf{42.04}$\\
\midrule
Llama $2$ $13$B & $16/16/16$ & $44.27$ & $54.43$ & $63.41$ & $60.76$ & $55.68$ \\
\midrule
RTN & $8/8/8$ & $29.06$ & $24.23$ & $29.93$ & $29.03$ & $27.62$ \\
SmoothQuant & $8/8/8$ & $30.98$ & $29.29$ & $35.36$ & $35.29$ & $32.37$ \\
FlexRound & $8/8/8$ & $41.09$ & $51.58$ & $61.39$ & $59.41$ & $53.28$ \\
LRQ (Ours) & $8/8/8$ & $42.88$ & $51.97$ & $62.14$ & $59.93$ & $\mathbf{54.08}$\\
\midrule
Llama $2$ $70$B & $16/16/16$ & $57.79$ & $65.16$ & $80.44$ & $74.61$ & $69.11$ \\
\midrule
RTN & $8/8/8$ & $46.82$ & $53.37$ & $66.23$ & $58.51$ & $55.97$ \\
SmoothQuant & $8/8/8$ & $47.51$ & $53.84$ & $68.35$ & $63.94$ & $57.99$ \\
FlexRound & $8/8/8$ & $54.27$ & $61.11$ & $77.45$ & $71.31$ & $65.57$ \\
LRQ (Ours) & $8/8/8$ & $54.44$ & $62.61$ & $76.99$ & $71.78$ & $\mathbf{66.12}$ \\
\bottomrule
\end{tabular}
\end{center}
\end{table}

\begin{table}[h]
\caption{Zero-shot performance of Llama $2$ on common sense reasoning tasks (BoolQ, PIQA, HellaSwag, WinoGrande, ARC easy and challenge, and OpenBookQA) with per-channel asymmetric weight quantization, per-token asymmetric activation quantization, and per-token asymmetric KV cache quantization (if applied). Please refer to Figure \ref{fig:diagram_pertoken}. The accuracy ($\%$) is reported for common sense reasoning tasks. The number of bits used for weights, activations, and KV cache is expressed as W/A/KV.} 
\label{tab:llama2_csr_per_token_appendix}
\begin{center}
\small
\resizebox{\linewidth}{!}{
\begin{tabular}{lccccccccc}
\toprule
Method & \makecell{\# Bits (W/A/KV)} & BoolQ & PIQA & HellaSwag &  WinoGrande & ARC-e & ARC-c & OBQA & Average\\
\midrule
Llama $2$ $7$B & $16/16/16$ & $71.07$ & $76.99$ & $72.96$ & $67.25$ & $53.58$ & $40.53$ & $40.80$ & $60.45$ \\
\midrule
RTN & $8/8/16$ & $69.54$ & $76.93$ & $72.21$ & $67.17$ & $53.24$ & $41.04$ & $40.60$ & $60.10$ \\
SmoothQuant & $8/8/16$ & $70.73$ & $77.04$ & $72.67$ & $66.77$ & $53.37$ & $40.78$ & $41.60$ & $60.42$ \\
FlexRound & $8/8/16$ & $72.26$ & $76.88$ & $72.57$ & $66.93$ & $53.70$ & $40.36$ & $40.40$ & $60.44$ \\
LRQ (Ours) & $8/8/16$ & $72.54$ & $77.15$ & $72.58$ & $67.09$ & $53.70$ & $41.04$ & $40.40$ & $\mathbf{60.64}$ \\
\hdashline
RTN & $8/8/8$ & $69.60$ & $77.20$ & $72.26$ & $67.09$ & $53.62$ & $39.85$ & $41.00$ & $60.09$  \\
SmoothQuant & $8/8/8$ & $70.61$ & $77.42$ & $72.62$ & $66.54$ & $53.37$ & $40.27$ & $40.80$ & $60.23$ \\
FlexRound & $8/8/8$ & $72.02$ & $77.09$ & $72.50$ & $67.40$ & $54.17$ & $40.19$ & $40.80$ & $60.60$ \\
LRQ (Ours) & $8/8/8$ & $72.45$ & $77.04$ & $72.70$ & $67.09$ & $53.66$ & $40.61$ & $41.60$ & $\mathbf{60.74}$ \\
\hdashline
RTN & $4/8/16$ & $67.95$ & $74.32$ & $65.84$ & $62.12$ & $46.68$ & $37.20$ & $35.80$ & $55.70$ \\
SmoothQuant & $4/8/16$ & $42.54$ & $64.15$ & $41.15$ & $54.06$ & $35.61$ & $27.99$ & $32.00$ & $42.50$ \\
FlexRound & $4/8/16$ & $71.96$ & $77.04$ & $72.17$ & $65.59$ & $53.58$ & $39.85$ & $40.20$ & $60.06$ \\
LRQ (Ours) & $4/8/16$ & $72.94$ & $76.88$ & $71.85$ & $65.27$ & $53.96$ & $39.85$ & $40.80$ & $\mathbf{60.22}$ \\
\hdashline
RTN & $4/8/8$ & $68.13$ & $75.14$ & $65.89$ & $62.67$ & $46.42$ & $36.52$ & $36.20$ & $55.85$ \\
SmoothQuant & $4/8/8$ & $43.03$ & $63.71$ & $41.08$ & $54.30$ & $35.69$ & $27.99$ & $32.60$ & $42.63$  \\
FlexRound & $4/8/8$ & $71.71$ & $76.77$ & $72.24$ & $66.14$ & $53.49$ & $40.02$ & $40.40$ & $60.11$ \\
LRQ (Ours) & $4/8/8$ & $73.00$ & $76.99$ & $71.90$ & $65.98$ & $54.38$ & $39.68$ & $41.20$ & $\mathbf{60.45}$ \\
\midrule
Llama $2$ $13$B & $16/16/16$ & $69.02$ & $79.05$ & $76.62$ & $69.61$ & $57.95$ & $44.28$ & $42.00$ & $62.65$ \\
\midrule
RTN & $8/8/16$ & $67.80$ & $78.89$ & $75.61$ & $68.90$ & $58.08$ & $43.69$ & $41.60$ & $62.08$ \\
SmoothQuant & $8/8/16$ & $68.99$ & $79.33$ & $76.47$ & $70.64$ & $57.53$ & $44.71$ & $41.80$ & $\mathbf{62.78}$ \\
FlexRound & $8/8/16$ & $69.27$ & $78.73$ & $76.62$ & $69.69$ & $57.62$ & $44.71$ & $42.20$ & $62.69$ \\
LRQ (Ours) & $8/8/16$ & $69.24$ & $78.67$ & $76.48$ & $69.30$ & $57.79$ & $44.03$ & $42.40$ & $62.56$ \\
\hdashline
RTN & $8/8/8$ & $67.46$ & $78.73$ & $75.57$ & $68.51$ & $58.12$ & $44.28$ & $41.40$ & $62.01$ \\
SmoothQuant & $8/8/8$ & $68.81$ & $78.78$ & $76.39$ & $70.72$ & $57.11$ & $44.20$ & $41.60$ & $62.52$ \\
FlexRound & $8/8/8$ & $69.36$ & $79.16$ & $76.67$ & $69.53$ & $57.83$ & $44.37$ & $42.80$ & $\mathbf{62.82}$ \\
LRQ (Ours) & $8/8/8$ & $69.02$ & $78.78$ & $76.48$ & $69.93$ & $57.83$ & $43.86$ & $42.00$ & $62.56$ \\
\hdashline
RTN & $4/8/16$ & $65.20$ & $73.61$ & $60.00$ & $58.80$ & $49.12$ & $36.18$ & $34.80$ & $53.96$ \\
SmoothQuant & $4/8/16$ & $61.74$ & $57.07$ & $37.17$ & $52.09$ & $31.61$ & $24.40$ & $31.40$ & $42.21$ \\
FlexRound & $4/8/16$ & $69.14$ & $78.67$ & $75.67$ & $68.98$ & $58.92$ & $44.20$ & $41.00$ & $62.37$ \\
LRQ (Ours) & $4/8/16$ & $71.10$ & $78.29$ & $75.75$ & $69.30$ & $57.74$ & $43.69$ & $41.00$ & $\mathbf{62.41}$ \\
\hdashline
RTN & $4/8/8$ & $65.23$ & $74.05$ & $60.04$ & $58.64$ & $49.07$ & $35.92$ & $35.20$ & $54.02$ \\
SmoothQuant & $4/8/8$ & $61.62$ & $56.53$ & $37.31$ & $51.38$ & $31.57$ & $24.74$ & $30.60$ & $41.96$  \\
FlexRound & $4/8/8$ & $69.05$ & $78.51$ & $75.51$ & $69.53$ & $58.75$ & $43.60$ & $41.20$ & $62.31$ \\
LRQ (Ours) & $4/8/8$ & $71.13$ & $78.29$ & $75.79$ & $68.90$ & $57.83$ & $43.34$ & $41.20$ & $\mathbf{62.35}$ \\
\midrule
Llama $2$ $70$B & $16/16/16$ & $76.70$ & $80.85$ & $80.85$ & $76.95$ & $59.72$ & $47.95$ & $44.40$ & $66.77$ \\
\midrule
RTN & $8/8/16$ & $76.02$ & $81.07$ & $80.37$ & $76.01$ & $60.14$ & $48.04$ & $44.40$ & $66.58$ \\
SmoothQuant & $8/8/16$ & $75.75$ & $80.96$ & $80.60$ & $77.90$ & $59.34$ & $47.70$ & $45.20$ & $66.78$ \\
FlexRound & $8/8/16$ & $75.72$ & $81.56$ & $80.60$ & $75.77$ & $60.19$ & $48.89$ & $44.80$ & $\mathbf{66.79}$ \\
LRQ (Ours) & $8/8/16$ & $75.84$ & $81.66$ & $80.64$ & $75.93$ & $60.40$ & $48.38$ & $44.00$ & $66.69$ \\
\hdashline
RTN & $8/8/8$ & $76.02$ & $81.07$ & $80.45$ & $75.61$ & $60.31$ & $47.87$ & $43.80$ & $66.45$ \\
SmoothQuant & $8/8/8$ & $76.06$ & $81.07$ & $80.63$ & $76.32$ & $59.51$ & $47.61$ & $45.20$ & $66.63$ \\
FlexRound & $8/8/8$ & $75.93$ & $81.45$ & $80.48$ & $75.85$ & $60.06$ & $48.55$ & $44.80$ & $66.73$ \\
LRQ (Ours) & $8/8/8$ & $75.99$ & $81.50$ & $80.61$ & $75.77$ & $59.97$ & $49.49$ & $45.20$ & $\mathbf{66.93}$ \\
\hdashline
RTN & $4/8/16$ & $75.63$ & $78.73$ & $71.28$ & $69.61$ & $53.24$ & $43.34$ & $40.20$ & $61.72$ \\
SmoothQuant & $4/8/16$ & $49.79$ & $70.95$ & $48.43$ & $54.70$ & $44.74$ & $32.51$ & $37.80$ & $48.42$ \\
FlexRound & $4/8/16$ & $77.80$ & $80.90$ & $80.06$ & $74.66$ & $60.31$ & $47.61$ & $43.60$ & $66.42$ \\
LRQ (Ours) & $4/8/16$ & $77.92$ & $80.74$ & $80.38$ & $75.14$ & $60.35$ & $47.95$ & $42.80$ & $\mathbf{66.47}$ \\
\hdashline
RTN & $4/8/8$ & $75.90$ & $79.22$ & $71.39$ & $70.56$ & $53.11$ & $43.60$ & $40.40$ & $62.03$ \\
SmoothQuant & $4/8/8$ & $50.46$ & $71.60$ & $48.35$ & $55.09$ & $44.87$ & $32.17$ & $37.40$ & $48.56$  \\
FlexRound & $4/8/8$ & $77.31$ & $80.96$ & $79.89$ & $75.30$ & $60.19$ & $48.21$ & $43.40$ & $66.47$ \\
LRQ (Ours) & $4/8/8$ & $77.92$ & $81.28$ & $80.42$ & $75.06$ & $60.94$ & $48.04$ & $42.60$ & $\mathbf{66.61}$ \\
\bottomrule
\end{tabular}
}
\end{center}
\end{table}

\begin{table}[h]
\caption{Zero-shot performance of Llama $2$ on common sense reasoning tasks (BoolQ, PIQA, HellaSwag, WinoGrande, ARC easy and challenge, and OpenBookQA) with per-channel asymmetric weight quantization, per-token asymmetric activation quantization, and per-token asymmetric KV cache quantization (if applied). Please refer to Figure \ref{fig:diagram_pertoken}. The accuracy ($\%$) is reported for common sense reasoning tasks. The number of bits used for weights, activations, and KV cache is expressed as W/A/KV.} 
\begin{center}
\small
\resizebox{\linewidth}{!}{
\begin{tabular}{lccccccccc}
\toprule
Method & \makecell{\# Bits (W/A/KV)} & BoolQ & PIQA & HellaSwag &  WinoGrande & ARC-e & ARC-c & OBQA & Average\\
\midrule
Llama $2$ $7$B & $16/16/16$ & $71.07$ & $76.99$ & $72.96$ & $67.25$ & $53.58$ & $40.53$ & $40.80$ & $60.45$ \\
\midrule
RTN & $4/8/8$ & $68.13$ & $75.14$ & $65.89$ & $62.67$ & $46.42$ & $36.52$ & $36.20$ & $55.85$ \\
SmoothQuant & $4/8/8$ & $43.03$ & $63.71$ & $41.08$ & $54.30$ & $35.69$ & $27.99$ & $32.60$ & $42.63$  \\
FlexRound & $4/8/8$ & $71.71$ & $76.77$ & $72.24$ & $66.14$ & $53.49$ & $40.02$ & $40.40$ & $60.11$ \\
LRQ (Ours) & $4/8/8$ & $73.00$ & $76.99$ & $71.90$ & $65.98$ & $54.38$ & $39.68$ & $41.20$ & $\mathbf{60.45}$ \\
\midrule
Llama $2$ $13$B & $16/16/16$ & $69.02$ & $79.05$ & $76.62$ & $69.61$ & $57.95$ & $44.28$ & $42.00$ & $62.65$ \\
\midrule
RTN & $4/8/8$ & $65.23$ & $74.05$ & $60.04$ & $58.64$ & $49.07$ & $35.92$ & $35.20$ & $54.02$ \\
SmoothQuant & $4/8/8$ & $61.62$ & $56.53$ & $37.31$ & $51.38$ & $31.57$ & $24.74$ & $30.60$ & $41.96$  \\
FlexRound & $4/8/8$ & $69.05$ & $78.51$ & $75.51$ & $69.53$ & $58.75$ & $43.60$ & $41.20$ & $62.31$ \\
LRQ (Ours) & $4/8/8$ & $71.13$ & $78.29$ & $75.79$ & $68.90$ & $57.83$ & $43.34$ & $41.20$ & $\mathbf{62.35}$ \\
\midrule
Llama $2$ $70$B & $16/16/16$ & $76.70$ & $80.85$ & $80.85$ & $76.95$ & $59.72$ & $47.95$ & $44.40$ & $66.77$ \\
\midrule
RTN & $4/8/8$ & $75.90$ & $79.22$ & $71.39$ & $70.56$ & $53.11$ & $43.60$ & $40.40$ & $62.03$ \\
SmoothQuant & $4/8/8$ & $50.46$ & $71.60$ & $48.35$ & $55.09$ & $44.87$ & $32.17$ & $37.40$ & $48.56$  \\
FlexRound & $4/8/8$ & $77.31$ & $80.96$ & $79.89$ & $75.30$ & $60.19$ & $48.21$ & $43.40$ & $66.47$ \\
LRQ (Ours) & $4/8/8$ & $77.92$ & $81.28$ & $80.42$ & $75.06$ & $60.94$ & $48.04$ & $42.60$ & $\mathbf{66.61}$ \\
\bottomrule
\end{tabular}
}
\end{center}
\end{table}

\begin{table}[h]
\caption{Five-shot performance of Llama $2$ on Massive Multitask Language Understanding with per-channel asymmetric weight quantization, per-token asymmetric activation quantization, and per-token asymmetric KV cache quantization (if applied). Please refer to Figure \ref{fig:diagram_pertoken}. The accuracy ($\%$) is reported for four groups of disciplines (STEM, Humanities, Social Science, and Other). The number of bits used for weights, activations, and KV cache is expressed as W/A/KV.} 
\label{tab:llama2_mmlu_per_token_appendix}
\begin{center}
\small
\begin{tabular}{lcccccc}
\toprule
Method & \makecell{\# Bits (W/A/KV)} & STEM & Humanities & Social Science & Other & Average\\
\midrule
Llama $2$ $7$B & $16/16/16$ & $37.04$ & $43.38$ & $51.84$ & $52.44$ & $45.96$ \\
\midrule
RTN & $8/8/16$ & $36.41$ & $42.49$ & $50.31$ & $52.47$ & $45.20$ \\
SmoothQuant & $8/8/16$ & $37.01$ & $43.04$ & $51.64$ & $52.13$ & $45.73$ \\
FlexRound & $8/8/16$ & $36.38$ & $42.91$ & $51.80$ & $52.87$ & $45.76$ \\
LRQ (Ours) & $8/8/16$ & $36.91$ & $43.27$ & $52.19$ & $52.78$ & $\mathbf{46.05}$ \\
\hdashline
RTN & $8/8/8$ & $36.15$ & $42.85$ & $50.34$ & $52.31$ & $45.24$ \\
SmoothQuant & $8/8/8$ & $36.22$ & $43.19$ & $51.25$ & $52.22$ & $45.54$ \\
FlexRound & $8/8/8$ & $36.98$ & $42.91$ & $51.87$ & $52.28$ & $45.76$ \\
LRQ (Ours) & $8/8/8$ & $36.88$ & $43.12$ & $51.67$ & $52.53$ & $\mathbf{45.83}$ \\
\hdashline
RTN & $4/8/16$ & $27.63$ & $25.87$ & $27.82$ & $28.32$ & $27.24$ \\
SmoothQuant & $4/8/16$ & $27.07$ & $24.72$ & $22.49$ & $25.88$ & $25.00$ \\
FlexRound & $4/8/16$ & $37.01$ & $42.40$ & $50.80$ & $50.34$ & $44.92$ \\
LRQ (Ours) & $4/8/16$ & $36.78$ & $42.66$ & $51.19$ & $51.73$ & $\mathbf{45.36}$ \\
\hdashline
RTN & $4/8/8$ & $28.00$ & $25.80$ & $27.53$ & $28.01$ & $27.16$ \\
SmoothQuant & $4/8/8$ & $26.77$ & $24.87$ & $22.81$ & $25.85$ & $25.05$ \\
FlexRound & $4/8/8$ & $37.81$ & $42.55$ & $50.47$ & $50.65$ & $45.14$ \\
LRQ (Ours) & $4/8/8$ & $36.88$ & $42.53$ & $50.80$ & $52.22$ & $\mathbf{45.36}$ \\
\midrule
Llama $2$ $13$B & $16/16/16$ & $44.27$ & $54.43$ & $63.41$ & $60.76$ & $55.68$ \\
\midrule
RTN & $8/8/16$ & $43.57$ & $52.88$ & $61.88$ & $61.17$ & $54.76$ \\
SmoothQuant & $8/8/16$ & $44.04$ & $53.69$ & $63.21$ & $61.35$ & $55.47$ \\
FlexRound & $8/8/16$ & $43.84$ & $53.65$ & $63.37$ & $61.10$ & $55.39$ \\
LRQ (Ours) & $8/8/16$ & $44.80$ & $53.75$ & $63.47$ & $60.73$ & $\mathbf{55.57}$ \\
\hdashline
RTN & $8/8/8$ & $43.87$ & $52.88$ & $62.33$ & $60.67$ & $54.81$ \\
SmoothQuant & $8/8/8$ & $44.10$ & $53.58$ & $63.11$ & $60.95$ & $55.33$ \\
FlexRound & $8/8/8$ & $44.17$ & $52.88$ & $63.76$ & $61.29$ & $55.33$ \\
LRQ (Ours) & $8/8/8$ & $44.50$ & $53.07$ & $63.24$ & $61.26$ & $\mathbf{55.35}$ \\
\hdashline
RTN & $4/8/16$ & $30.55$ & $26.08$ & $33.51$ & $35.07$ & $30.74$ \\
SmoothQuant & $4/8/16$ & $27.04$ & $24.23$ & $25.48$ & $26.16$ & $25.55$ \\
FlexRound & $4/8/16$ & $42.91$ & $50.80$ & $62.11$ & $60.27$ & $53.77$ \\
LRQ (Ours) & $4/8/16$ & $43.24$ & $52.41$ & $61.78$ & $60.24$ & $\mathbf{54.30}$ \\
\hdashline
RTN & $4/8/8$ & $30.95$ & $26.31$ & $32.92$ & $34.58$ & $30.67$ \\
SmoothQuant & $4/8/8$ & $27.07$ & $24.25$ & $25.22$ & $26.43$ & $25.57$ \\
FlexRound & $4/8/8$ & $42.88$ & $50.71$ & $61.94$ & $59.93$ & $53.77$\\
LRQ (Ours) & $4/8/8$ & $43.90$ & $52.56$ & $62.07$ & $59.96$ & $\mathbf{54.49}$  \\
\midrule
Llama $2$ $70$B & $16/16/16$ & $57.79$ & $65.16$ & $80.44$ & $74.61$ & $69.11$ \\
\midrule
RTN & $8/8/16$ & $56.06$ & $63.00$ & $78.32$ & $73.10$ & $67.20$ \\
SmoothQuant & $8/8/16$ & $58.02$ & $64.53$ & $80.21$ & $74.21$ & $\mathbf{68.80}$ \\
FlexRound & $8/8/16$ & $57.69$ & $63.80$ & $79.98$ & $73.63$ & $68.30$ \\
LRQ (Ours) & $8/8/16$ & $57.95$ & $64.48$ & $80.21$ & $73.90$ & $68.70$ \\
\hdashline
RTN & $8/8/8$ & $56.23$ & $63.55$ & $78.39$ & $73.01$ & $67.41$ \\
SmoothQuant & $8/8/8$ & $57.49$ & $64.68$ & $80.37$ & $74.43$ & $\mathbf{68.82}$ \\
FlexRound & $8/8/8$ & $57.22$ & $63.97$ & $79.62$ & $73.81$ & $68.22$ \\
LRQ (Ours) & $8/8/8$ & $57.95$ & $63.85$ & $80.34$ & $73.94$ & $68.52$ \\
\hdashline
RTN & $4/8/16$ & $41.12$ & $45.72$ & $56.78$ & $53.49$ & $48.95$ \\
SmoothQuant & $4/8/16$ & $26.84$ & $24.08$ & $26.55$ & $25.88$ & $25.63$ \\
FlexRound & $4/8/16$ & $59.96$ & $62.98$ & $79.04$ & $73.23$ & $67.56$ \\
LRQ (Ours) & $4/8/16$ & $56.46$ & $64.59$ & $79.07$ & $72.83$ & $\mathbf{67.92}$ \\
\hdashline
RTN & $4/8/8$ & $41.19$ & $45.74$ & $57.52$ & $53.61$ & $49.16$ \\
SmoothQuant & $4/8/8$ & $27.37$ & $24.59$ & $27.59$ & $25.94$ & $26.16$ \\
FlexRound & $4/8/8$ & $56.26$ & $62.89$ & $78.78$ & $72.92$ & $67.26$ \\
LRQ (Ours) & $4/8/8$ & $55.57$ & $64.65$ & $78.97$ & $72.52$ & $\mathbf{67.65}$ \\
\bottomrule
\end{tabular}
\end{center}
\end{table}

\begin{table}[h]
\caption{Five-shot performance of Llama $2$ on Massive Multitask Language Understanding with per-channel asymmetric weight quantization, per-token asymmetric activation quantization, and per-token asymmetric KV cache quantization (if applied). Please refer to Figure \ref{fig:diagram_pertoken}. The accuracy ($\%$) is reported for four groups of disciplines (STEM, Humanities, Social Science, and Other). The number of bits used for weights, activations, and KV cache is expressed as W/A/KV.} 
\begin{center}
\small
\begin{tabular}{lcccccc}
\toprule
Method & \makecell{\# Bits (W/A/KV)} & STEM & Humanities & Social Science & Other & Average\\
\midrule
Llama $2$ $7$B & $16/16/16$ & $37.04$ & $43.38$ & $51.84$ & $52.44$ & $45.96$ \\
\midrule
RTN & $4/8/8$ & $28.00$ & $25.80$ & $27.53$ & $28.01$ & $27.16$ \\
SmoothQuant & $4/8/8$ & $26.77$ & $24.87$ & $22.81$ & $25.85$ & $25.05$ \\
FlexRound & $4/8/8$ & $37.81$ & $42.55$ & $50.47$ & $50.65$ & $45.14$ \\
LRQ (Ours) & $4/8/8$ & $36.88$ & $42.53$ & $50.80$ & $52.22$ & $\mathbf{45.36}$ \\
\midrule
Llama $2$ $13$B & $16/16/16$ & $44.27$ & $54.43$ & $63.41$ & $60.76$ & $55.68$ \\
\midrule
RTN & $4/8/8$ & $30.95$ & $26.31$ & $32.92$ & $34.58$ & $30.67$ \\
SmoothQuant & $4/8/8$ & $27.07$ & $24.25$ & $25.22$ & $26.43$ & $25.57$ \\
FlexRound & $4/8/8$ & $42.88$ & $50.71$ & $61.94$ & $59.93$ & $53.77$\\
LRQ (Ours) & $4/8/8$ & $43.90$ & $52.56$ & $62.07$ & $59.96$ & $\mathbf{54.49}$  \\
\midrule
Llama $2$ $70$B & $16/16/16$ & $57.79$ & $65.16$ & $80.44$ & $74.61$ & $69.11$ \\
\midrule
RTN & $4/8/8$ & $41.19$ & $45.74$ & $57.52$ & $53.61$ & $49.16$ \\
SmoothQuant & $4/8/8$ & $27.37$ & $24.59$ & $27.59$ & $25.94$ & $26.16$ \\
FlexRound & $4/8/8$ & $56.26$ & $62.89$ & $78.78$ & $72.92$ & $67.26$ \\
LRQ (Ours) & $4/8/8$ & $55.57$ & $64.65$ & $78.97$ & $72.52$ & $\mathbf{67.65}$ \\
\bottomrule
\end{tabular}
\end{center}
\end{table}

\clearpage

In Table \ref{tab:llama_csr_per_tensor_appendix} and \ref{tab:llama2_csr_per_tensor_appendix}, LRQ exhibits a slightly superior zero-shot performance on common sense reasoning tasks compared to FlexRound, which we believe is noteworthy since FlexRound already achieves the zero-shot performance on common sense reasoning tasks comparable to FP16 baselines. The close proximity in zero-shot performance between FlexRound and FP16 baselines on common sense reasoning tasks limits the potential for a substantial performance disparity between FlexRound and LRQ. Despite LRQ approaching the zero-shot performance of FP16 baselines more closely than FlexRound, the difference in zero-shot performance between FlexRound and LRQ cannot be anticipated to be large after all.

Nevertheless, as expounded in Section \ref{sec:intro}, it is crucial to emphasize that LRQ demonstrates competitive performance relative to FP16 baselines on both common sense reasoning tasks and Massive Multitask Language Understanding (MMLU), a feat not accomplished by FlexRound that excels solely on common sense reasoning tasks. Given the comprehensive evaluation of large language models (LLMs) across diverse benchmarks, the proficiency of LRQ in excelling across both common sense reasoning tasks and MMLU holds significant implications in the field of LLM quantization.

Regarding the $8$-bit weight quantization presented in Table \ref{tab:llama2_csr_per_token_appendix} and \ref{tab:llama2_mmlu_per_token_appendix}, the adoption of a per-token asymmetric activation quantization scheme results in even naive rounding-to-nearest (RTN) performing closely to the levels of FP16 baselines on both common sense reasoning tasks and MMLU. As a result, while LRQ exhibits slightly higher accuracy compared to SmoothQuant and FlexRound for most Llama $2$ models, it can be concluded that SmoothQuant, FlexRound, and LRQ are nearly evenly matched. 

In the context of $4$-bit weight quantization as presented in Table \ref{tab:llama2_csr_per_token_appendix}, FlexRound achieves zero-shot accuracy levels comparable to FP16 baselines on common sense reasoning tasks, resulting in a relatively small zero-shot performance gap between FlexRound and LRQ, like the scenario depicted in Table \ref{tab:llama_csr_per_tensor_appendix} and \ref{tab:llama2_csr_per_tensor_appendix}. However, in the case of $4$-bit weight quantization in Table \ref{tab:llama2_mmlu_per_token_appendix}, LRQ surpasses FlexRound by a margin ranging from $0.2$ to $0.7$ percent. Although these increments in five-shot accuracy on MMLU in Table \ref{tab:llama2_mmlu_per_token_appendix} may seem modest compared to those in Table \ref{tab:llama_mmlu_per_tensor_appendix} and \ref{tab:llama2_mmlu_per_tensor_appendix}, we believe that the rise in five-shot accuracy by $0.2$ to $0.7$ percent on MMLU is significant. This is particularly noteworthy as it brings the five-shot accuracy gap between LRQ and FP16 baselines to less than $1.5$ percent on MMLU, while the corresponding gap between FlexRound and FP16 baselines remains more or less at two percent for Llama $2$ $13$B and $70$B.

\clearpage
\section{Implementation Details}\label{appendix:detail}

\begin{figure*}[h]
    \centering
    \includegraphics[width=0.925\linewidth]{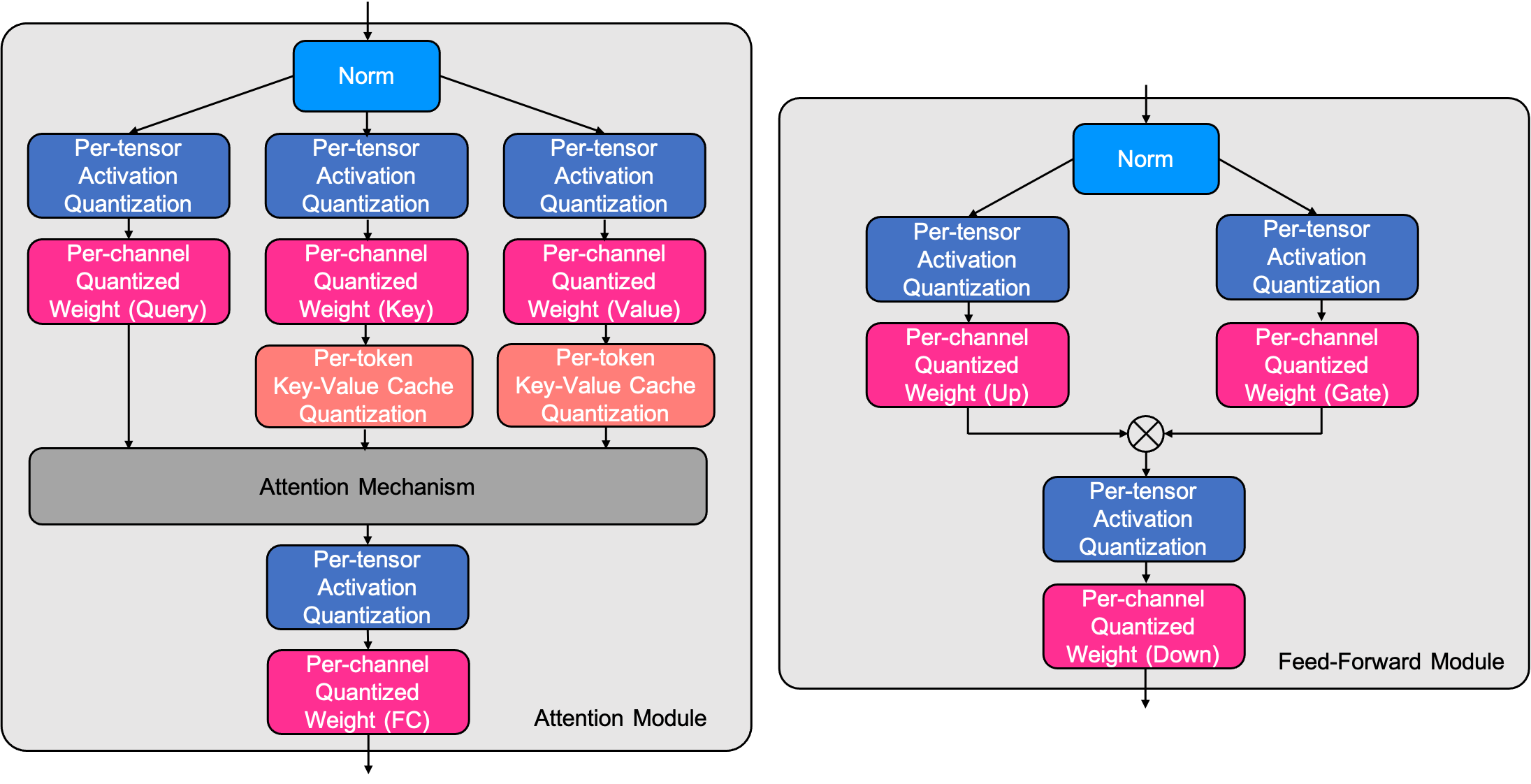}
    \caption{Illustration of a quantized Transformer block with per-channel asymmetric weight quantization, per-tensor asymmetric static activation quantization, and per-token asymmetric KV cache quantization. We remain the inputs of softmax and normalization layers in FP16.} 
    \label{fig:diagram}
\end{figure*}

\begin{table}[h]
\caption{Learning rate and batch size for FlexRound and LRQ when employing a per-tensor asymmetric static activation quantization scheme (see Figure \ref{fig:diagram}) in Table \ref{tab:llama_csr_per_tensor}, \ref{tab:llama2_csr_per_tensor}, \ref{tab:llama_mmlu_per_tensor}, \ref{tab:llama2_mmlu_per_tensor}, \ref{tab:llama_csr_per_tensor_appendix}, \ref{tab:llama_mmlu_per_tensor_appendix}, \ref{tab:llama2_csr_per_tensor_appendix}, and \ref{tab:llama2_mmlu_per_tensor_appendix}.}\label{tab:hyperparameter_per_tensor} 
\begin{center}
\small
\resizebox{\linewidth}{!}{
\begin{tabular}{llccccccc}
\toprule
Method & \makecell{Configuration} & Llama $7$B & Llama $13$B & Llama $33$B & Llama $65$B & Llama $2$ $7$B & Llama $2$ $13$B & Llama $2$ $70$B \\
\midrule
FlexRound & Learning rate & $3$e-$3$ & $3$e-$3$ & $1$e-$3$ & $2$e-$3$ & $3$e-$3$ & $3$e-$3$ & $1$e-$3$ \\
& Batch size & $4$ & $4$ & $2$ & $2$ & $2$ & $2$ & $2$ \\
\midrule
LRQ & Learning rate & $3$e-$3$ & $2$e-$3$ & $1.5$e-$3$ & $1$e-$3$ & $1$e-$3$ & $1.5$e-$3$ & $1$e-$3$ \\
& Batch size & $2$ & $2$ & $2$ & $2$ & $2$ & $2$ & $2$ \\
\bottomrule
\end{tabular}
}
\end{center}
\end{table}

For the quantization scheme depicted in Figure \ref{fig:diagram}, both FlexRound and LRQ are implemented in the experimental setting of QDrop \citep{wei2022qdrop} with the exception of the number of iterations for block-wise reconstruction, the batch size, and the learning rate. For all the Llama and Llama $2$ models, the number of iterations for block-wise reconstruction is set to $5000$ for both FlexRound and LRQ. The learning rate and the batch size for FlexRound and LRQ are described in \ref{tab:hyperparameter_per_tensor}. Notice that when applying LRQ to Llama $2$ $70$B, the key and value projection weights are quantized via not LRQ but FlexRound due to the presence of GQA \citep{ainslie2023gqa} in Llama $2$ $70$B. To obtain the experimental results in Table \ref{tab:llama_csr_per_tensor} and \ref{tab:llama_mmlu_per_tensor}, per-token asymmetric KV cache quantization is applied after completing block-wise reconstruction for all the Transformer blocks. For both activation quantization and KV cache quantization, we employ rounding-to-nearest.

\clearpage

\begin{figure*}[h]
    \centering
    \includegraphics[width=0.925\linewidth]{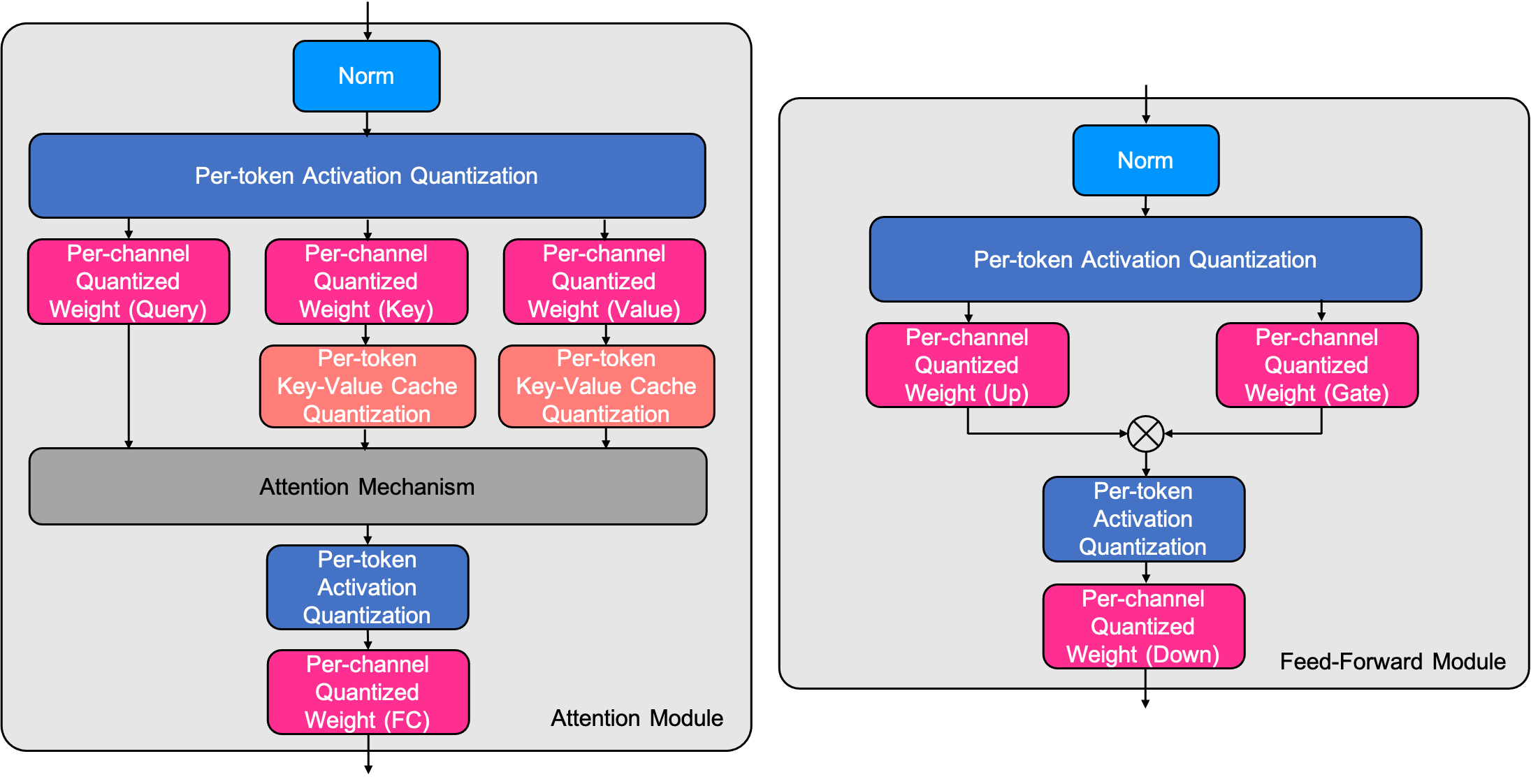}
    \caption{Illustration of a quantized Transformer block with per-channel asymmetric weight quantization, per-token asymmetric activation quantization, and per-token asymmetric KV cache quantization. We remain the inputs of softmax and normalization layers in FP16.}
    \label{fig:diagram_pertoken}
\end{figure*}

\begin{table}[h]
\caption{Learning rate for FlexRound and LRQ when adopting a per-token asymmetric activation quantization scheme (see Figure \ref{fig:diagram_pertoken}) in Table \ref{tab:llama2_csr_per_token}, \ref{tab:llama2_mmlu_per_token}, \ref{tab:llama2_csr_per_token_appendix}, and \ref{tab:llama2_mmlu_per_token_appendix}.}\label{tab:hyperparameter_per_token} 
\begin{center}
\small
\begin{tabular}{lcccc}
\toprule
Method & Weight & Llama $2$ $7$B & Llama $2$ $13$B & Llama $2$ $70$B \\
\midrule
FlexRound & $8$-bit & $1$e-$4$ & $4$e-$4$ & $3$e-$4$ \\
& $4$-bit & $5$e-$4$ & $4$e-$4$ & $5$e-$4$ \\
\midrule
LRQ & $8$-bit & $1$e-$4$ & $2$e-$4$ & $4$e-$4$ \\
& $4$-bit & $5$e-$4$ & $5$e-$4$ & $4$e-$4$  \\
\bottomrule
\end{tabular}
\end{center}
\end{table}

In the case of quantization scheme indicated in Figure \ref{fig:diagram_pertoken}, both FlexRound and LRQ are first implemented in the experimental setting of BRECQ \citep{li2021brecq} with the exception of the number of iterations for block-wise reconstruction, the batch size, and the learning rate. The number of iterations for block-wise reconstruction and the batch size are set to $5000$ and $2$ respectively, for every Llama $2$ model regardless of the number of bits used for weights. Table \ref{tab:hyperparameter_per_token} exhibits the learning rate for FlexRound and LRQ in the case of $8$-bit and $4$-bit weight quantization, respectively. As explained in the above paragraph, when LRQ is applied to Llama $2$ $70$B, weights in key and value projections are quantized via FlexRound. Here, when quantizing Llama $2$ $7$B into $4$-bit via LRQ, the attention module is quantized via LRQ, but the feed-forward module is quantized via FlexRound. In addition, when quantizing Llama $2$ $70$B into $4$-bit via LRQ, the feed-forward module is quantized via LRQ, but the attention module is quantized via FlexRound. 
To gain the experimental results in Table \ref{tab:llama2_csr_per_token} and \ref{tab:llama2_mmlu_per_token}, per-token asymmetric activation quantization and per-token asymmetric KV cache quantization are sequentially applied after finishing block-wise reconstruction for all the Transformer blocks. For both activation quantization and KV cache quantization, we employ rounding-to-nearest.

All experiments about SmoothQuant are conducted based on the code provided in the SmoothQuant github repository\footnote{\url{https://github.com/mit-han-lab/smoothquant}}. Following \citet{xiao2022smoothquant}, we select $\alpha$, the hyperparameter to determine how much difficulty of activation quantization to shift to weight quantization, to be $0.8$ for every Llama model, $0.85$ for Llama $2$ $7$B and $13$B, and $0.9$ for Llama $2$ $70$B.

\begin{table}[h]
\caption{Learning rate for LRQ when adopting per-channel weight-only quantization in Table \ref{tab:llama2_weight_only}.}\label{tab:hyperparameter_weight_only} 
\begin{center}
\small
\begin{tabular}{lcccc}
\toprule
Method & Weight & Llama $2$ $7$B & Llama $2$ $13$B & Llama $2$ $70$B \\
\midrule
FlexRound & $3$-bit & $2.5$e-$3$ & $5$e-$4$ & $5$e-$4$ \\
& $4$-bit & $3$e-$3$ & $7$e-$4$ & $5$e-$4$  \\
\hdashline
LRQ & $3$-bit & $6$e-$4$ & $5$e-$4$ & $5$e-$4$ \\
& $4$-bit & $9$e-$4$ & $6$e-$4$ & $5$e-$4$  \\
\bottomrule
\end{tabular}
\end{center}
\end{table}

Table \ref{tab:hyperparameter_weight_only} displays the learning rate for LRQ employed in Table \ref{tab:llama2_weight_only}. For OmniQuant in Table \ref{tab:llama2_weight_only}, we utilized pre-trained OmniQuant models provided in the OmniQuant github repository\footnote{\url{https://github.com/OpenGVLab/OmniQuant}}.

For evaluation, we use Eleuther AI's \textit{lm-evaluation-harness} \citep{eval-harness} for common sense reasoning tasks and follow the evalution code in the MMLU github repository\footnote{\url{https://github.com/hendrycks/test}} for the MMLU benchmark.

\clearpage
\section{Ratio of the number of learnable parameters in LRQ to the number of pre-trained weights}\label{appendix:ratio}

\begin{table}[h]
\caption{Ratio of the number of learnable parameters in LRQ to the number of pre-trained weights for an intermediate Transformer block of each Llama model when setting the rank $r$ to $2048$ for large language models beyond $30$B parameters or to $1024$ for smaller models.} 
\label{tab:ratio}
\begin{center}
\small
\begin{tabular}{lccc}
\toprule
Model & Number of pre-trained weights (A) & Number of learnable parameters in LRQ (B) & Ratio (B/A) \\
\midrule
Llama 7B & \makecell{$4096\times4096\times4$ \\ $+ 4096\times11008\times3$} & \makecell{$(4096\times1024+1024\times4096)\times4$ \\ $+ (4096\times1024 + 1024\times11008)\times3$} & $\mathbf{39.51\%}$ \\
Llama 13B & \makecell{$5120\times5120\times4$ \\ $+ 5120\times13824\times3$} & \makecell{$(5120\times1024+1024\times5120)\times4$ \\ $+ (5120\times1024 + 1024\times13824)\times3$} & $\mathbf{31.57\%}$ \\
Llama 33B & \makecell{$6656\times6656\times4$ \\ $+ 6656\times17920\times3$} & \makecell{$(6656\times2048+2048\times6656)\times4$ \\ $+ (6656\times2048 + 2048\times17920)\times3$} & $\mathbf{48.60\%}$ \\
Llama 65B & \makecell{$8192\times8192\times4$ \\ $+ 8192\times22016\times3$} & \makecell{$(8192\times2048+2048\times8192)\times4$ \\ $+ (8192\times2048 + 2048\times22016)\times3$} & $\mathbf{39.51\%}$ \\
\bottomrule
\end{tabular}
\end{center}
\end{table}

\clearpage
\section{Average and Standard Deviation of FlexRound and LRQ}\label{appendix:standard_deviation}

For common sense reasoning tasks in Table \ref{tab:llama_csr_per_tensor} and \ref{tab:llama2_csr_per_tensor}, LRQ slightly outperforms FlexRound in the case of Llama $2$ $70$B and significantly surpasses FlexRound in the case of Llama $33$B, but FlexRound is better than LRQ in the case of Llama $2$ $7$B. To investigate how meaningful the improvement of LRQ over FlexRound is, we carry out three random trials for Llama $2$ $7$B, Llama $33$B, and Llama $2$ $70$B, presenting the average and standard deviation of them.

\begin{table}[h]
\caption{Average and standard deviation of zero-shot performance of FlexRound and LRQ over three random trials on common sense reasoning tasks (BoolQ, PIQA, HellaSwag, WinoGrande, ARC easy and challenge, and OpenBookQA) with per-channel asymmetric weight quantization, per-tensor asymmetric static activation quantization, and per-token asymmetric KV cache quantization. The number of bits used for weights, activations, and KV cache is $8$-bit.} 
\label{tab:standard_deviation}
\begin{center}
\small
\begin{tabular}{lcccc}
\toprule
Method & \makecell{\# Bits (W/A/KV)} & Llama $2$ $7$B & Llama $33$B & Llama $2$ $70$B \\
\midrule
FlexRound & $8/8/8$ & $59.72 \pm 0.73$ & $62.83 \pm 0.36$ & $65.65 \pm 0.30$ \\
LRQ (Ours) & $8/8/8$ & $\mathbf{59.90 \pm 0.18}$ & $\mathbf{63.81 \pm 0.16}$ & $\mathbf{65.89 \pm 0.06}$ \\
\bottomrule
\end{tabular}
\end{center}
\end{table}

As seen in Table \ref{tab:standard_deviation}, not only does the average of LRQ surpass that of FlexRound, but the standard deviation of LRQ is also smaller than that of FlexRound, which strengthens our assertion that FlexRound might be prone to overfitting when applied to the quantization of LLMs.

\clearpage
\section{Combination of SmoothQuant with FlexRound and LRQ}

\begin{table}[h]
\caption{Zero-shot performance of Llama $7$B on common sense reasoning tasks (BoolQ, PIQA, HellaSwag, WinoGrande, ARC easy and challenge, and OpenBookQA) with per-channel asymmetric weight quantization and per-tensor asymmetric static activation quantization, while keeping the KV cache in FP16. Here, `SQ + FlexRound' and `SQ + LRQ' denote FlexRound and LRQ that initially begin their own learning process from the SmoothQuant baseline in lieu of the rounding-to-nearest baseline, respectively. The accuracy ($\%$) is reported for common sense reasoning tasks. The number of bits used for weights, activations, and KV cache is expressed as W/A/KV.}
\label{tab:sq_initialization_csr}
\begin{center}
\small
\resizebox{\linewidth}{!}{
\begin{tabular}{lccccccccc}
\toprule
Method & \makecell{\# Bits (W/A/KV)} & BoolQ & PIQA & HellaSwag &  WinoGrande & ARC-e & ARC-c & OBQA & Average\\
\midrule
Llama $7$B & $16/16/16$ & $73.15$ & $77.31$ & $72.96$ & $67.09$ & $52.48$ & $41.38$ & $42.40$ & $60.97$ \\
\midrule
FlexRound & $8/8/16$ & $73.76$ & $76.66$ & $71.75$ & $67.01$ & $52.31$ & $40.02$ & $42.20$ & $\mathbf{60.53}$ \\
SQ+FlexRound & $8/8/16$ & $73.85$ & $76.77$ & $71.54$ & $66.38$ & $51.43$ & $40.44$ & $41.60$ & $60.29$ \\
\hdashline
LRQ & $8/8/16$ & $73.03$ & $77.64$ & $72.10$ & $66.77$ & $52.95$ & $40.87$ & $41.60$ & $\mathbf{60.71}$ \\
SQ+LRQ & $8/8/16$ & $73.15$ & $76.88$ & $72.24$ & $66.38$ & $52.86$ & $40.61$ & $40.60$ & $60.39$ \\
\bottomrule
\end{tabular}
}
\end{center}
\end{table}

\begin{table}[h]
\caption{Five-shot performance of Llama $7$B on Massive Multitask Language Understanding with per-channel asymmetric weight quantization and per-tensor asymmetric static activation quantization, while keeping the KV cache in FP16. Here, `SQ + FlexRound' and `SQ + LRQ' denote FlexRound and LRQ that initially begin their own learning process from the SmoothQuant baseline in lieu of the rounding-to-nearest baseline, respectively. The accuracy ($\%$) is reported for four groups of disciplines (STEM, Humanities, Social Science, and Other). The number of bits used for weights, activations, and KV cache is expressed as W/A/KV.}
\label{tab:sq_initialization_mmlu}
\begin{center}
\small
\begin{tabular}{lcccccc}
\toprule
Method & \makecell{\# Bits (W/A/KV)} & STEM & Humanities & Social Science & Other & Average\\
\midrule
Llama $7$B & $16/16/16$ & $30.58$ & $33.88$ & $38.19$ & $38.25$ & $35.12$ \\
\midrule
FlexRound & $8/8/16$ & $28.30$ & $29.20$ & $30.13$ & $33.47$ & $30.20$ \\
SQ+FlexRound & $8/8/16$ & $30.98$ & $29.71$ & $33.80$ & $35.26$ & $\mathbf{32.16}$ \\
\hdashline
LRQ & $8/8/16$ & $29.69$ & $32.48$ & $37.63$ & $38.80$ & $\mathbf{34.47}$ \\
SQ+LRQ & $8/8/16$ & $30.35$ & $31.84$ & $37.44$ & $37.32$ & $34.01$\\
\bottomrule
\end{tabular}
\end{center}
\end{table}

As SmoothQuant is orthogonal to block-wise reconstruction, one might wonder how the performance of FlexRound and LRQ would change when FlexRound and LRQ start their own learning process from the SmoothQuant baseline in place of the RTN baseline. Table \ref{tab:sq_initialization_csr} and \ref{tab:sq_initialization_mmlu} reveal the performance of `SmoothQuant (SQ) + FlexRound' and `SmoothQuant (SQ) + LRQ' on common sense reasoning benchmarks and the MMLU benchmark, respectively. Unfortunately, in most cases, SmoothQuant does not display its efficacy when combined with FlexRound and LRQ. Although SmoothQuant enhances five-shot performance of FlexRound on MMLU by almost two percent, `SQ + FlexRound' still underperforms LRQ as well as `SQ + LRQ' on MMLU, which implies that employing low-rank weight-scaling matrices would be a better choice than using full weight-scaling matrices with additional pre-processing like an uniform per-channel scaling transformation in SmoothQuant.

\clearpage
\section{Process of \texorpdfstring{$\mL_2 \mU_2 + \vr_2 + \vc_2$}{} in Eq. \ref{eq:LRQ}}

Similar to the broadcasting process in Python Numpy, we add $\mL_2 \mU_2$, $\vr_2$, and $\vc_2$. \\
To be more specific, let $\mL_2 \mU_2$ be 
\[
\begin{bmatrix} 
LU_{(1, 1)} & LU_{(1, 2)} & \cdots & LU_{(1, C_{in})}  \\ 
\vdots & \vdots & \vdots & \vdots \\
LU_{(C_{out}, 1)} & LU_{(C_{out}, 2)} & \cdots & LU_{(C_{out}, C_{in})},
\end{bmatrix}
\]
$\vr_2$ be 
\[
\begin{bmatrix} 
r_{1} \\ 
r_{2} \\
\vdots \\
r_{C_{out}}
\end{bmatrix},
\]
and $\mathbf{c}_2$ be 
\[
\begin{bmatrix} c_{1} & c_{2} & \cdots & c_{C_{in}} \end{bmatrix}.
\]
Then, by the broadcasting process, $\mL_2 \mU_2 + \vr_2 + \vc_2$ can be expressed as
\[
\begin{bmatrix} LU_{(1, 1)} + r_{1} + c_{1} & LU_{(1, 2)} + r_{1} + c_{2} & \cdots & LU_{(1, C_{in})} + r_{1} + c_{C_{in}}  \\
\vdots & \vdots & \vdots & \vdots \\
LU_{(C_{out}, 1)} + r_{C_{out}} + c_{1} & LU_{(C_{out}, 2)} + r_{C_{out}} + c_{2} & \cdots & LU_{(C_{out}, C_{in})} + r_{C_{out}} + c_{C_{in}} 
\end{bmatrix}.
\]

\end{document}